\documentclass[a4paper, 12pt, twoside]{book}
\usepackage[round]{natbib}
\usepackage{bibentry}
\usepackage[utf8x]{inputenc}
\usepackage[T1]{fontenc}
\usepackage{setspace}
\usepackage{hyperref} 
\usepackage{amsmath,amssymb}
\usepackage{algorithmic} 
\usepackage[ruled,vlined]{algorithm2e} 

\usepackage{changepage} 
\usepackage{bm} 
\usepackage{titlepic}
\usepackage{mathrsfs}
\usepackage{nomencl} 
\usepackage{xfp} 
\usepackage[aboveskip=1pt,labelfont=bf,labelsep=period,singlelinecheck=off]{caption}
\makenomenclature
\usepackage{geometry}
\usepackage{subcaption} 
\usepackage{microtype} 
\usepackage{fancyhdr} 
\usepackage{ragged2e}
\usepackage{bmpsize} 
\usepackage{lastpage,fancyhdr,graphicx} 
\DisableLigatures[f]{encoding = *, family = * }
\usepackage[table]{xcolor}

\newcolumntype{+}{!{\vrule width 2pt}}
 \DeclareMathOperator*{\argmaxB}{argmax}   
 \DeclareMathOperator*{\argminB}{argmin}   

\newlength\savedwidth

\definecolor{tgcolor}{rgb}{0.8,0.2,0.2}

\makeatletter
\renewcommand{\@biblabel}[1]{\quad#1.}
\makeatother

\newgeometry{left=2.9cm, right=2.6cm, top=3.6cm, bottom=4.4cm}
\begin{document}

\onehalfspacing
\pdfbookmark[0]{Abstract}{abstract}
\title{}
\begin{titlepage}
    \begin{center}
        \vspace*{1cm}
            
        \Huge
        \textbf{Visual processing in context of \\
reinforcement learning}
            
        \vspace{0.5cm}
        \LARGE
        \vspace{1.5cm}
            
        \textbf{Hlynur Davíð Hlynsson}
            
        \vfill
            
        A thesis presented for the degree of\\
        Doctor of Engineering 
            
        \vspace{0.8cm}
            
        \includegraphics[width=0.4\textwidth]{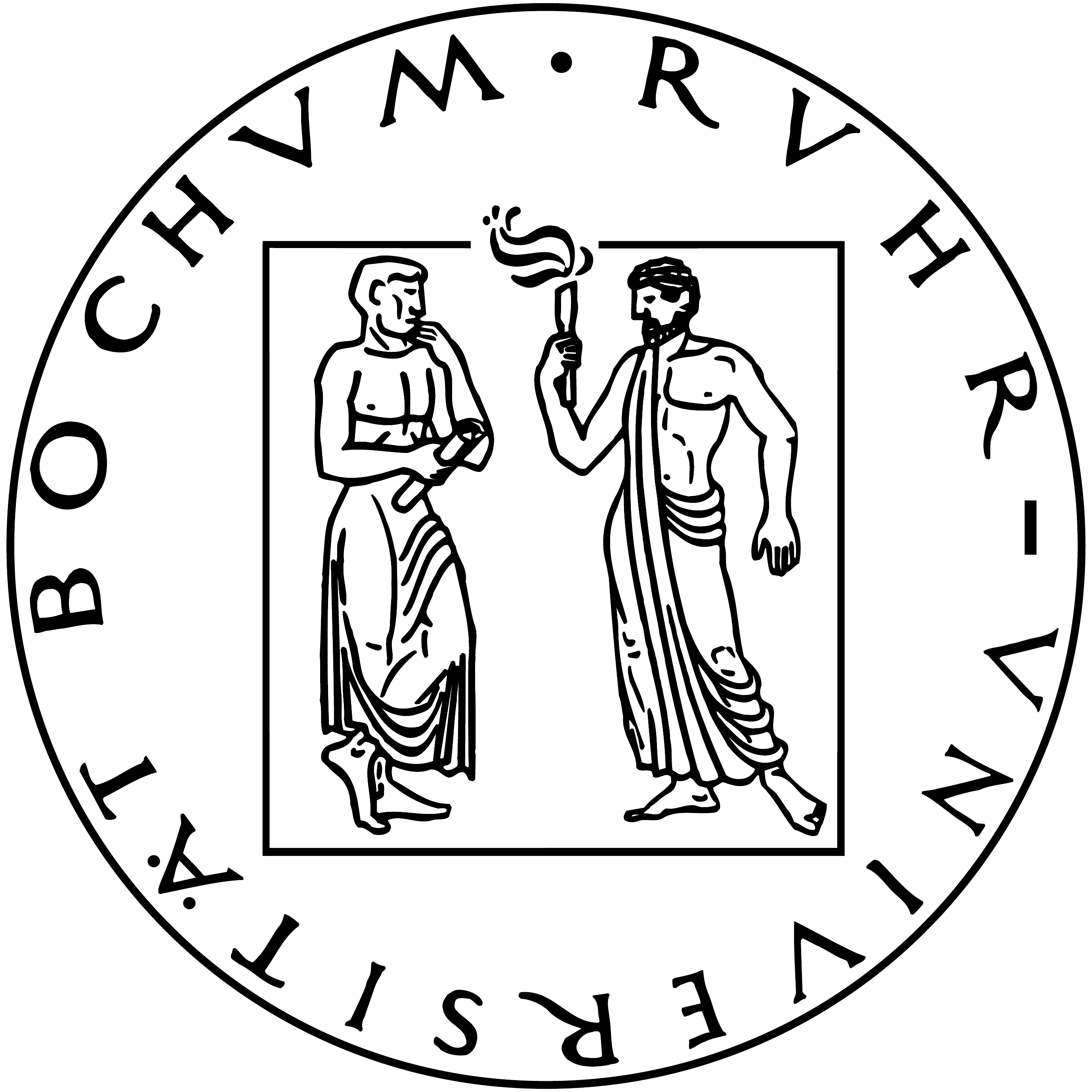}
            
        \Large
        Faculty of Electrical Engineering
and Information Technology\\
        Ruhr-University Bochum \\
        Germany\\
        July 2021
            
    \end{center}
\end{titlepage}

\section*{Visual processing in context of reinforcement learning}
\vspace{1cm}
\subsection*{Dissertation for the degree of Doctor of Engineering of the Faculty of Electrical Engineering and Information Technology at the Ruhr-Universität Bochum}

\begin{tabular}[b]{ll}
             &                                    \\
            &                                    \\
             &                                    \\

             &                                    \\
Name of the author: & Hlynur Davíð Hlynsson \\
             &                                    \\
Place of birth: & Reykjavík \\
             &                                    \\

First supervisor: & Prof. Dr. Laurenz Wiskott \\
             & Ruhr-Universität Bochum, Germany\\
             &                                    \\
Second supervisor: & Prof. Dr. Tobias Glasmachers \\
             & Ruhr-Universität Bochum, Germany \\
             &                                    \\

Year of thesis submission: &  2021 \\
             &                                    \\

Date of the oral examination: &  March 17th 2022 \\
             &                                    \\
\end{tabular}

\chapter*{Abstract}

Although deep reinforcement learning (RL) has recently enjoyed many successes, its methods are still data inefficient, which makes solving numerous problems prohibitively expensive in terms of data. We aim to remedy this by taking advantage of the rich supervisory signal in unlabeled data for learning state representations. This thesis introduces three different representation learning algorithms that have access to different subsets of the data sources that traditional RL algorithms use:
 
(i) GRICA is inspired by independent component analysis (ICA) and trains a deep neural network to output statistically independent features of the input. GrICA does so by minimizing the mutual information between each feature and the other features. Additionally, GrICA only requires an unsorted collection of environment states.

(ii) Latent Representation Prediction (LARP) requires more context: in addition to requiring a state as an input, it also needs the previous state and an action that connects them. This method learns state representations by predicting the representation of the environment's next state given a current state and action. The predictor is used with a graph search algorithm.

(iii) RewPred learns a state representation by training a deep neural network to learn a smoothed version of the reward function. The representation is used for preprocessing inputs to deep RL, while the reward predictor is used for reward shaping. This method needs only state-reward pairs from the environment for learning the representation.

We discover that every method has their strengths and weaknesses, and conclude from our experiments that including unsupervised representation learning in RL problem-solving pipelines can speed up learning.
\newpage

\pdfbookmark[0]{Kurzfassung der Dissertation}{kurzfassung}
\chapter*{Kurzfassung der Dissertation}
Obwohl tiefes Verstärkungslernen (VL) in den letzten Jahren große Erfolge erzielt hat, sind dessen Methoden immer noch datenineffizient, was die Lösung vieler Probleme unerschwinglich macht. Wir untersuchen die Möglichkeit, dies zu beheben, indem wir das informationsreiche Überwachungssignal in nicht gekennzeichnete Daten für die Darstellung von Lernzuständen nutzen. In dieser Arbeit werden drei verschiedene Repräsentationslernalgorithmen vorgestellt, die Zugriff auf verschiedene Teilmengen der Datenquellen haben, die herkömmliche VL-Algorithmen zum Lernen verwenden:

(i) GrICA ist von der unabhängigen Komponentenanalyse (ICA) inspiriert und trainiert ein tiefes neuronales Netzwerk, um statistisch unabhängige Komponenten der Eingabe auszugeben. GrICA minimiert die gemeinsamen Informationen von einzelnen Merkmalen mit den jeweils anderen Merkmalen. Zusätzlich erfordert GrICA lediglich eine unsortierte Sammlung von Umgebungszuständen.

(ii) Latent Representation Prediction (LARP) erfordert mehr Kontextdaten: Als Eingabe benötigt sie zusätzlich zu einem Zustand auch den entsprechenden vorherigen Zustand und eine Handlung, welche diese verbindet. Die Methode lernt Zustandsdarstellungen, indem sie die Darstellung des nächsten Zustands der Umgebung mithilfe eines aktuellen Zustands und einer aktuellen Aktion vorhersagt. Der Prädiktor wird zusammen mit einem Graphensuchalgorithmus verwendet.

(iii) RewPred lernt die Zustandsdarstellung, indem ein tiefes neuronales Netzwerk trainiert wird eine geglättete Version der Belohnungsfunktion zu lernen. Die Darstellung wird zur Vorverarbeitung von Eingaben im tiefen VL verwendet, während der Belohnungsprädiktor als Belohnungsformung dient. Diese Methode benötigt einzig Status-Belohnungs-Paare aus der Umgebung, um die Darstellung zu lernen.

Wir stellen fest, dass jede Methode ihre Stärken und Schwächen hat, und schließen aus unseren Experimenten, dass das Einbeziehen von unbeaufsichtigtem Repräsentationslernen in VL-Problemlösungspipelines das Lernen beschleunigen kann.

\pdfbookmark[0]{Dedication}{dedication}
\clearpage
\begin{center}
    \thispagestyle{empty}
    \vspace*{\fill}
    \vspace*{\fill}
\end{center}
\clearpage

\clearpage
\pdfbookmark[0]{Acknowledgements}{acknowledgements}
\chapter*{Acknowledgements}
I want to first thank my supervisor Prof. Laurenz Wiskott for giving me the opportunity to research the niches of machine learning that I find interesting. His insightful feedback and outstanding enthusiasm and intuition for the field proved invaluable to me and others in this fast growing area of research. The advice of my second supervisor, Tobias Glasmachers, also proved extremely helpful, especially on the topic of reinforcement learning.

I'm grateful for the endless love and support from my partner Lisa Schmitz, who made the time of my PhD studies the best in my life -- so far. My special thanks go also to her parents, Rosemarie Schmitz and Georg Schmitz, for their support during this time.

I'm thankful for the helpful and pleasant environment created by the other PhD students of the Institut für Neuroinformatik (INI): Merlin Schüler, Robin Schiewer, Zahra Fayyaz, Eddie Seabrook, Mortiz Lange, Frederick Baucks, Jan Bollenbacher and Jan Tekülve. I would also like to thank two alumni of the INI, Alberto Escalante and Fabian Schönfeld, for the support they have given me. I also want to thank the INI staff outside the group for their help over the years: Arno Berg, Angelika Wille and Kathleen Schmidt.

Last but not least, I want to thank my loving family for creating the circumstances that gave me room to train the skills that I needed to pursue a PhD to begin with: Ólöf Ingibjörg Einarsdóttir, Hlynur Höskuldsson, Ólafur Hlynsson and Höskuldur Hlynsson.

\clearpage

\tableofcontents
\cleardoublepage
\listoffigures

\clearpage
\listoftables

 \pagenumbering{arabic}
\fancyhead[RE,LO]{\nouppercase{\rightmark}}

\chapter{Introduction}


Mankind has been interested in the concept of infusing inanimate objects with its intellect for thousands of years, with stories of artificially intelligent beings reaching back thousands of years. This interest manifests itself for example in the  story told by ancient Greeks of the great automata Talos, who was crafted out of bronze by the smithing god Hephaestus to protect the mythological queen Europa \citep{rhodios2008argonautika}. Automata have been built by craftsmen from different cultures throughout the ages, but they have been simple mechanical beings \citep{mccorduck1979machines}. 

The possibility of satisfying the human desire to craft intelligent beings has only arisen in the middle of the 19th century with the founding of artificial intelligence (AI) as an academic discipline. The vast scope of AI has given rise to different fields,
each with its own application domains, methodologies and philosophies. One such example is machine learning (ML), an area of AI that is concerned with algorithms that leverage data for decision-making. This field has experienced great success in both academia and industry in the last decade with increased access to powerful computers and large databases in addition to a deluge of advanced computational techniques and flexible software solutions \citep{clark20152015}.     

The field is not without its drawbacks, however. Machine learning algorithms need data corresponding to months or years of human experience to get competence in tasks that a person can master in minutes. People have the advantage that they come equipped with a stronger understanding of the world. In this dissertation, we aim to level the playing field for ML models that perform sequential decision-making by exploring different ways for them to "understand" the world through representations.

 In Section \ref{Deepreinforcementlearning}, we discuss one of the most promising fields of artificial intelligence, deep reinforcement learning, and mention its victories. The current situation is evaluated in Section \ref{openproblems} where we describe the challenges and disadvantages of the field. The value of the research direction we put forward is underlined in Section \ref{researchaim} as well as our research objective and hypothesis. The chapter concludes with Section \ref{thesisoutline} where we outline the order of content in the dissertation.


\section{Deep reinforcement learning}
\label{Deepreinforcementlearning}
A watershed moment for artificial intelligence happened when \cite{krizhevsky2012imagenet} combined several techniques from the literature and constructed a deep\footnote{Neural networks that process the input hierarchically using at least more than two layers of computational layers are called \textit{deep}} neural network to outperform the competition by a significant margin in an image classification contest. This was the catalyst of the so-called \textit{deep learning revolution} \citep{sejnowski2018deep} which has impacted fields such as natural language processing \citep{wolf2020transformers}, bioinformatics \citep{li2019deep}, computer vision \citep{khan2018guide}, fraud detection and many others \citep{alom2019state}. 

The area of machine learning concerned with the training of deep neural networks is called deep learning. Deep learning methods have the advantage that they autonomously learn patterns in the data in a hierarchical manner. For example, edges are useful patterns for pictures of shapes such as squares, triangles and circles \citep{patrick2010ai}. The edges can be combined to corners and the number of corners can be counted for distinguishing between the different shapes. 

Since deep learning encompasses a broad set of machine learning algorithms, it can be readily combined with other areas of machine learning. One such area is reinforcement learning, where general goal-directed decision-making problems are studied. Reinforcement learning (RL) methods train models that are interacting sequentially with their environments to maximize a reward signal. Deep learning is frequently combined with RL techniques, allowing the models to map the inputs, such as high-dimensional image data, directly to actions. This combination has yielded promising results in different areas, ranging from recommender systems \citep{zhang2019deep} over autonomous driving \citep{kiran2021deep} to playing games \citep{silver2017mastering}. 

\section{Open problems}
\label{openproblems}
A known problem of deep neural networks is that they require a large amount of data for adequate performance. This data can be prohibitively expensive to obtain -- either in terms of time needed to create data and train the models for reinforcement learning methods or monetary cost of acquiring human-labeled training data for supervised learning methods -- which has encouraged the development of methods that learn representations from streams of more readily available, unlabeled data. 

This problem increases in severity in deep reinforcement learning (DRL). In the uncompromisingly titled blog post, \textit{Deep reinforcement learning doesn’t work yet}, \cite{irpan2018deep} identifies several fundamental problems of DRL. One of them is the problem of sample inefficiency, where many highly publicized state-of-the-art results on video games require hundreds of millions of frames of experience to achieve performance that humans reach in a matter of minutes. 

Another problem is the one of instability. Deep neural networks are highly expressive and optimize large numbers of parameters. This makes the design of DRL models difficult, as the search of hyperparameters\footnote{A hyperparameter is broadly speaking any design choice made by the programmer before the learning of the "regular" parameters starts.} that solve the problem can be quite time-consuming. Even when a promising set of hyperparameters is found, the difference between the performance of different models learned from scratch can be significant, depending on the random seed. This increased variance comes from the new source of randomness that is introduced to RL models, compared to regular regression learning: the agents actions are stochastic, increasingly so in the beginning of learning\footnote{There is a tradeoff between exploring the environment and exploiting the expected reward signal. A common strategy for RL agents is to start the learning with a high chance of performing random actions to explore different states of the environment and then decrease this chance as the learning progresses.}.

\section{Research aim}
\label{researchaim}
In this dissertation, we propose methods for unsupervised and self-supervised learning of representations for goal-directed behavior. Self-supervised learning methods use a subset of the input to predict the rest of it, foregoing the need of annotations while taking advantage of the powerful machinery of supervised learning methods. 

Tackling the open problems of data inefficiency and instability outlined above in order to further the field is our intention with this thesis. We do so by developing and investigating three different approaches: (i) unsupervised learning of a representation for RL agents, (ii) a method of jointly learning a predictor for planning a representation that is good for the transition prediction, and (iii) learning a representation for RL agents as the byproduct of reward prediction. We relate the data needed to learn the representations for our methods to the available data in the context of RL in Table \ref{tableofmethods}. Our hypothesis is that suitable state representations that reduce the complexity of high-dimensional inputs in RL settings can support a more stable and data efficient learning than having deep RL algorithms learn state representations from scratch. 

\begin{table}[h]
\begin{center}
\caption[Result: Input data type per method]{ {\bf Input data type per method.} If an RL agent is in a state $s$ and performs the action $a$, it will receive a reward of $r$ and transition to the state $s'$. The methods proposed in this thesis learn state representations by processing different subsets of the data tuples $\{s,a,r,s'\}$.  } 
\label{tableofmethods}
 \begin{tabular}{ c  l  r} 
  \textbf{Subset of $\{s,a,r,s'\}$ required for learning} & \textbf{Method} & \textbf{Chapter}
 \\ [0.5ex] 
 \hline
   $\{s\}$        &  GrICA   & 3\\
   \hline
   $\{s,a,s'\}$   &  LARP    & 4\\
   \hline
   $\{s,r,s'\}$   &  RewPred & 5\\
\end{tabular}\end{center}\end{table}

\section{Thesis outline}
\label{thesisoutline}
Here we outline the structure of the thesis. Three of the chapters are adapted from the work that were published over the course of the doctoral work. 
\nobibliography*
\begin{itemize}
    \item \textbf{Chapter 2: Background.} In this chapter, we go in further details on the main topics in this thesis and discuss their fundamentals. We introduce the formalism of Markov decision processes and explain the difference between model-based and model-free reinforcement learning algorithms. The machinery behind deep learning is then explained and the main building blocks of deep neural networks are illustrated. The chapter concludes with a discussion of the main representation learning methods. 
    \item \textbf{Chapter 3: Learning gradient-based ICA by neurally estimating mutual information.} This chapter discusses an adaption of independent component analysis (ICA) for DL. We introduce a novel application of a neural method for mutual information estimation to learn a representation with statistically independent features. The chapter is an adapted version of 
    \begin{itemize}
        \item \bibentry{hlynsson2019learning} \citep{hlynsson2019learning}
    \end{itemize}
    \item \textbf{Chapter 4: Latent representation prediction networks.} This chapter discusses a method for manipulable environments for jointly learning a representation of observations and a model for predicting the next representation, given an action. We learn the representation in a self-supervised manner, without the need of a reward signal. We introduce a new environment that is akin to manipulating toy objects for a viewpoint matching task. The representation is combined with a graph-search algorithm to find the goal viewpoint. The chapter is an adapted version of 
    \begin{itemize}
        \item \bibentry{hlynsson2020latent} \citep{hlynsson2020latent}
    \end{itemize}
    
    \item \textbf{Chapter 5: Reward prediction for representation learning and reward shaping.} This chapter discusses a self-supervised learning method to map high-dimensional inputs to a lower dimensional space for RL agents. We introduce a technique where a representation learned for a reward predictor is used to shape the reward for the agents. The chapter is an adapted version of 
    \begin{itemize}
        \item  \bibentry{hlynsson2021reward} \citep{hlynsson2021reward}

    \end{itemize}
    
    \item \textbf{Chapter 6: Comparison of our methods.} In this chapter, we directly compare the three different methods to state-of-the-art deep RL methods on four different environments: a visual pole-balancing environment, two goal-finding environment and an obstacle avoidance environment
    
        \item \textbf{Chapter 7: Summary and conclusion.} This chapter closes the dissertation with a brief summary of the thesis, concluding remarks and possible future work.

The following work was also published over the course of the doctoral studies:

\begin{itemize}
    \item \bibentry{hlynsson2019measuring} \citep{hlynsson2019measuring}
\end{itemize}

The paper compares supervised learning methods, but it is too dissimilar in topic from the rest of the work and is thus chosen to be omitted from this dissertation.
    
\end{itemize}

\clearpage

\chapter{Background}


This chapter lays out the fundamental concepts of machine learning that forms the focal point of the rest of the dissertation. In Section \ref{sec:rl}, we lay out the main object of study in reinforcement learning (RL), partially observable Markov decision processes, and present a brief taxonomy of reinforcement learning algorithms. We explain the basics of artificial neural networks and deep learning in Section \ref{sec:dlrn}. The most commonly used type of neural network used for processing visual data, the convolutional neural network, is then described, along with some of its main building blocks. In Section \ref{sec:repr_learn}, we discuss representation learning (also known as \textit{feature learning}) and motivate it in the context of reinforcement learning. 

For a more in-depth discussion of these topics, we refer the reader to the comprehensive textbook on RL by \cite{sutton2018reinforcement}, the deep learning book by \cite{Goodfellow-et-al-2016} and the excellent survey by \cite{bengio2013representation} on representation learning.

\section{Reinforcement learning} \label{sec:rl}



In this section, we formalize RL for the rest of the thesis. RL is one of the main disciplines of machine learning, and it covers how agents can learn to behave optimally in an environment to maximize a cumulative reward.

\subsection{Partially observable Markov decision processes}

A partially-observable Markov decision process (POMDP) is a general framework for modeling sequential decision processes in environments that can be stochastic, complex and contain hidden information. Formally, it is a tuple 

\begin{equation} \label{eq:pomdp}
(\mathcal{S}, \mathcal{A}, \mathcal{P}, \mathcal{R}, P(s_0), \Omega, \mathcal{O}, \gamma)\end{equation}

 which we also refer to as the \textit{environment}. The tuple is made up of the following elements:

\begin{itemize}
\setlength{\itemindent}{0.65cm}
    \item[$\mathcal{S}$:] The state space defines the possible configurations of the environment 
    \item[$\mathcal{A}$:] The action space describes how the agent is able to interact with the environment
    \item[$\mathcal{P}$:] The transition function $\mathcal{P}: \mathcal{S} \times \mathcal{A} \rightarrow P(\mathcal{S})$ dictates the effects of different actions in different states
    \item[$\mathcal{R}$:] The reward function $\mathcal{R}:  \mathcal{S} \times \mathcal{A} \times \mathcal{S} \rightarrow \mathbb{R}$ determines the immediate reward given to the agent for transitioning between any two states with any action
    \setlength{\itemindent}{1.3cm}

    \item[$P(s_0)$:] The initial state distribution
    \setlength{\itemindent}{0.65cm}

    \item[$\Omega$:] The observation space defines the aspects of the environment that the agent can perceive 
    \item[$\mathcal{O}$:] The observation function $\mathcal{S} \times \mathcal{A} \rightarrow P(\Omega)$ defines what (potentially transformed) subset of the environment the agent receives after acting in a given state
    \item[$\gamma$:] The reward discount factor
\end{itemize}
 
The environment starts in a state drawn from $P(s_0)$, from which the agent interacts sequentially with the environment by choosing action $a_t$ from action space $\mathcal{A}$ at time steps $t$. The agent receives an observation $o_t$ and a reward $r_t$ after each action. 

 The objective of an RL agent is to learn a \textit{policy} $\pi$ that determines the behavior of the agent in the environment by mapping states to a probability distribution over $\mathcal{A}$, written $\pi(a, s) = P(a_t = a | s_t = s)$. A discount factor $\gamma \in (0, 1)$ is usually included in the definition of POMDPs, and it comes into play in the optimization function of the agent. Namely, the policy should maximize the expected discounted future sum of rewards, or the expected \textit{return}, where the return is defined as

\begin{equation} \label{eq:ret}
R = \sum_{t=0}^\infty \gamma^t r_t
\end{equation}


The \textit{value function} is defined as the expectation of the return (Eq.\ref{eq:ret}), given a policy $\pi$ and an initial state $s_0 = s$ 

\begin{equation} \label{eq:vf}
 v_\pi(s) = \mathbb{E}\left[ R | s_0 = s, \pi\right] =  \mathbb{E}\left[ \sum_{t=0}^\infty \gamma^t r_t | s_0 = s, \pi\right] 
 \end{equation}

There is at least one \textit{optimal policy} $\pi^*$ that is better than or equal to others: $v_{\pi^*}(s) \geq v_{\pi'}(s)$ for all states $s$ and all other policies $\pi'$. 

\subsection{Model-free algorithms}

Model-free reinforcement learning learns the policy or a value function directly from experience without attempting to approximate the dynamics of the environment. Two popular classes of model-free methods are \textit{value-based} methods and \textit{policy-based} methods. 

Value-based methods approximate either the value function \citep{sutton1988learning} or another useful function that is similar to the value function, the \textit{action-value function} $q$. This function is defined as the expected return of following the policy $\pi$ after taking an action $a$ in a state $s$:

\begin{equation} \label{eq:saf}
 q_\pi(s, a) = \mathbb{E}_\pi\left[ R_t | s_t = s, a_t=a\right] =  \mathbb{E}_\pi\left[ \sum_{k=0}^\infty \gamma^k r_{t+k+1} | s_t = s, a_t=a\right] 
 \end{equation}

 Estimating the action-value function is a pivotal step for algorithms such as Q-learning \citep{watkins1992q}. A simple one-step Q-learning updating rule is \citep{sutton2018reinforcement}:
 
 \begin{equation} \label{eq:qlearn}
 q_\pi(s_t, a_t) \leftarrow q(s_t, a_t) + \eta\left[r_{t+1}+ \gamma \max_a q(s_{t+1}, a) -    q(s_t, a_t) \right] 
 \end{equation}
 where $\eta$ is a positive learning rate parameter and the initial values of $q_\pi(s_t, a_t)$ are chosen arbitrarily. Q-learning is guaranteed to converge to the optimal policy's  action-value function $q_{\pi^*}$, under certain conditions\footnote{ This depends on a good learning rate schedule and exploration techniques, which are difficult to determine in practice}, which in turn yields the optimal policy: $\pi^* = \argmaxB_a q_{\pi^*}(a, s)$. This method tabulates the values and thus works with discrete actions and state spaces. Q-learning has been combined with deep neural networks to work for actions and state spaces of higher dimensions \citep{mnih2015human}.
 
 Policy-based methods do not learn a value function, but rather learn the policy directly by optimizing an objective function with respect to $\pi$. We describe two of those methods that we employ in this work: (1) proximal policy optimization (PPO) \citep{schulman2017proximal} and (2) actor critic using Kronecker-factored trust region (ACKTR) \citep{wu2017scalable}. 
 
 PPO optimizes the objective function
 
 \begin{equation}
    L^{CLIP}(\theta) = \hat{\mathbb{E}}_t \left[ \min(\text{ratio}_t(\theta)\hat{A}_t, \text{clip} (\text{ratio}_t(\theta), 1-\epsilon, 1+\epsilon)\hat{A}_t \right]
 \end{equation}
 
 \noindent where $\epsilon$ is a hyperparameter, $\text{ratio}_t(\theta) = \frac{\pi_\theta(a_t | s_t)}{\pi_{\theta_{\text{old}}}(a_t, s_t)}$ and $\hat{A}$ is an estimator of the advantage function $A = q_\pi(s, a) - v_\pi(s)$. The clip term returns $\text{ratio}_t(\theta)$ if $1-\epsilon < \text{ratio}_t(\theta) < 1+\epsilon$, otherwise the value is clipped to the closer boundary value. The full PPO algorithm is shown in Algorithm \ref{algo: ppo}.

\begin{algorithm}[thb]
\caption{Proximal policy optimization}\label{algo: ppo}
\begin{algorithmic}[1]
\FOR{iteration $ = 1, 2, \dots$}
    \FOR{actor $= 1, 2, \dots, n$} 
        \STATE Run policy $\pi_{\theta_\text{old}}$ in environment for t time steps
        \STATE Compute advantage estimates $\hat{A}_1 , \dots , \hat{A}_t$
    \ENDFOR
    \STATE Optimize $L^{CLIP}$ with respect to $\theta$, with k epochs and minibatch size $m \leq nt$
    \STATE $\theta_{old} \leftarrow \theta$
\ENDFOR
\end{algorithmic}
\end{algorithm}

ACKTR applies the policy gradient updates

\begin{equation}
    \theta \leftarrow \theta - \eta \hat{F}^{-1} \nabla_\theta L
\end{equation}

\noindent where $\hat{F} \approx   \mathbb{E}[\nabla_\pi \log \pi (a_t | s_t) (\nabla_\theta \log \pi (a_t | s_t) )^T]$, $L$ is the log-likelihood of the output distribution of the policy and the learning rate $\eta = \min (\eta_{\max}, \sqrt{
\frac
{2\delta}
{\Delta \theta^T \hat{F} \Delta \theta }
}) $ is controlled dynamically with the trust region parameter $\delta$ to prevent the policy from converging prematurely to a poor policy.

\subsection{Model-based algorithms}
Model-based reinforcement learning (MBRL) algorithms learn the optimal policy $\pi^*$ by first estimating the transition function $\Tilde{\mathcal{P}}\approx \mathcal{P}$ and the reward function $\Tilde{\mathcal{R}}\approx \mathcal{R}$. These functions are usually called the environment \textit{dynamics} or \textit{world model} and are learned in a supervised fashion from a data set of observed transitions, $\mathcal{D} = \{(s_t, a_t, r_t, s_{t+1})_i\}$. The world models can be used in multiple different ways, depending on the algorithm, to derive the optimal policy. 

For example, sampling-based planning algorithms use $\Tilde{\mathcal{P}}$ and $\Tilde{\mathcal{R}}$ to sample action sequences and calculate their expected values:

\begin{equation} \label{eq:rsmbrl}
(A_t, \dots, A_{t+\tau}) = \argmaxB_{A_{t:t+\tau}} \mathbb{E}\left[ \sum_{k=t}^{t+\tau} \gamma^k \Tilde{\mathcal{R}}(s_k, a_k) | s_t = s, s_{t+1}, \dots, s_{t+\tau} \sim \Tilde{\mathcal{P}}\right]
 \end{equation}
 
The agent follows the action sequence associated with the highest expected reward in Equation \ref{eq:rsmbrl}. This is often combined with model-predictive control (MPC), where a new action sequence is calculated after taking the first action in the last sequence. There are different ways of choosing candidate action sequences, with the simplest being the random shooting algorithm \citep{richards2005robust}, that draws the actions from a uniform distribution.





\section{Deep learning}
\label{sec:dlrn}


For the last few years, deep learning has been on the center stage of machine learning research. We make extensive use of deep learning in this work because of its parallelizability, its efficient scaling with large data sets and its capability to approximate complex functions. 

The most basic type of deep learning method is the feedforward deep network, which comprises layers of artificial neurons. The theoretical capabilities of artificial neural networks were guaranteed by \cite{cybenko1989approximation}: his Universal Approximation Theorem has the implication that any continuous function of real numbers with values in a Euclidean space can be approximated by a neural network with one hidden layer. Unfortunately, it is a pure existence theorem, leaving the task of constructing the network to the engineer.
\subsection{The artificial neuron}
An artificial neuron \citep{rosenblatt1958perceptron} is a mathematical function that multiplies each input with a constant, adds a bias to the linear combination and then applies a non-linearity to the outcome: 
\begin{equation}
    y = \varphi\left( \sum_{i=1}^m w_i x_i  + b\right)
    \label{anneq}
\end{equation}

The non-linearity $\varphi$ is known as the \textit{activation function}, the coefficients $w_i$ are known as the \textit{weights} and the term $b$ is known as the \textit{bias}.

We now briefly discuss some commonly used activation functions that are employed in this thesis. For a more comprehensive overview of recent trends in the usage of activation functions, we encourage the reader to look at a comparison by \cite{nwankpa2018activation}.

The logistic function can be used for binary classification.

\begin{equation}
     \varphi_{\text{logistic}}(x) = \frac{1}{1+e^{-x}}
    \label{logistic}
\end{equation}

This function "squashes" the inputs to lie between $0$ and $1$, giving the output a probabilistic interpretation. 

The softmax function is an extension of the logistic function for several classes

\begin{equation}
     \varphi_{\text{softmax}}(x)_i =\frac{e^{x_i}}{\sum_{i=1}^n e^{x_i}}
    \label{softmax}
\end{equation}

The output of the softmax function is a vector of the same dimensionality as the input vector and sums to $1$.

The hyperbolic tangent function (tanh) squashes the input to lie between -1 and 1

\begin{equation}
     \varphi_{\text{tanh}}(x) = \frac{e^x - e^{-x}}{e^x - e^{-x}}
    \label{tanh}
\end{equation}

This has the computational advantage over the logistic function that biases in the gradients are avoided and 0-centered data gives rise to larger derivatives during optimization of the networks \citep{lecun2012efficient}, making them a more frequent choice as an activation function in hidden layers. 

The most popular nonlinearity for deep neural networks is the rectifier function

\begin{equation}
     \varphi_{\text{ReLU}}(x) = \max(0, x)
    \label{relu}
\end{equation}

The rectifier function offers the same advantages as the tanh function but at a lower cost, as evaluating exponentials and performing division is avoided. The rectifier function is also called a rectified linear unit, and it is commonly abbreviated as "ReLU". 
\subsection{Feedforward neural networks}

\begin{figure*}[h]
	\centering
	\begin{minipage}{.96\columnwidth}
		\centering
	\includegraphics[width=\textwidth]{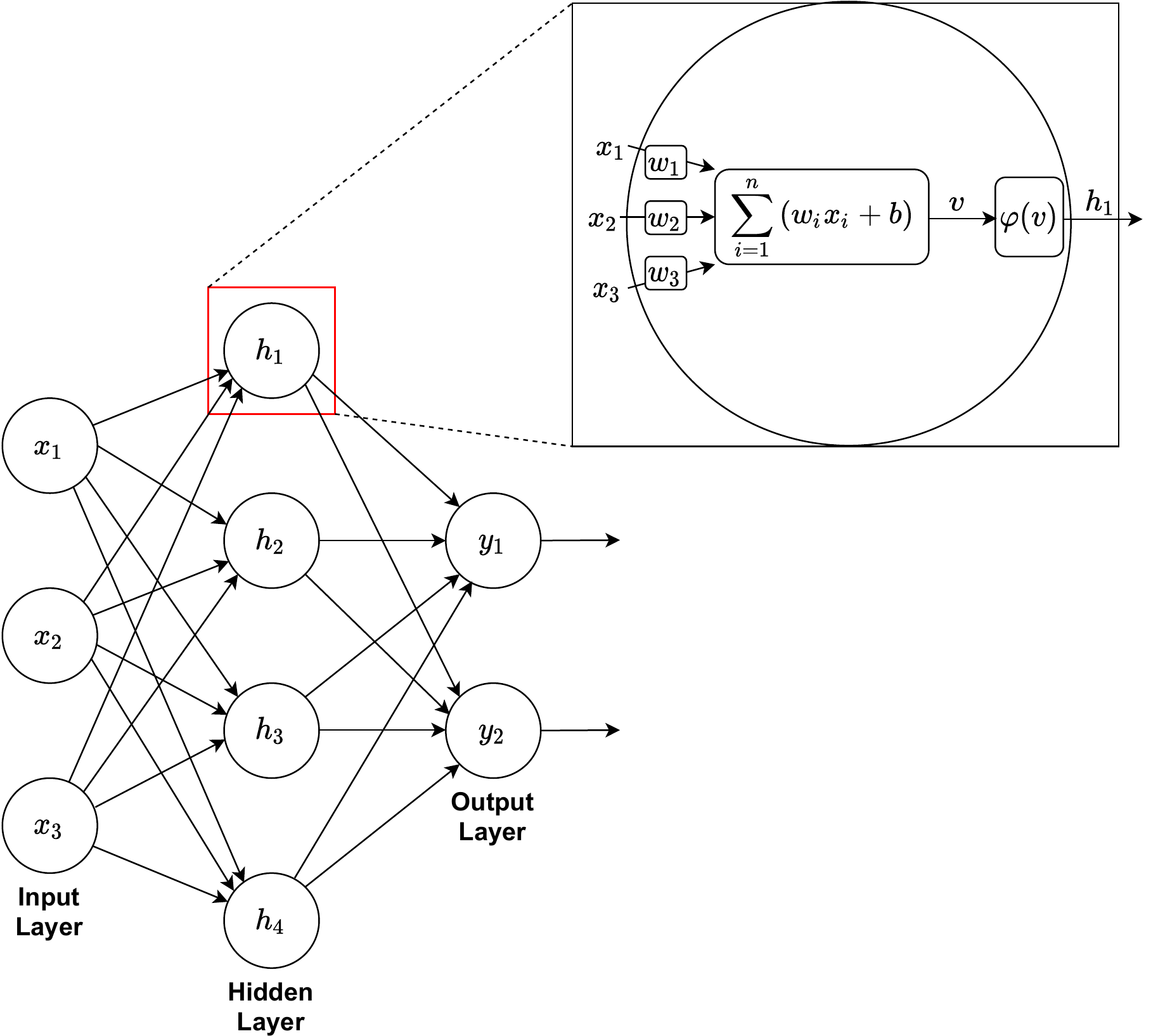}
	\end{minipage}%
		\caption[Illustration: A fully-connected neural network]{ \textbf{A Fully-connected neural network.} This network has three inputs and two layers: one hidden layer with four units and an output layer with two units.    }	\label{feedforward_nn}
\end{figure*}

Computational units implementing the function in Eq. \ref{anneq} can be arranged hierarchically, with the input of a neuron consisting of the output of other neurons. An example feedforward neural network or \textit{multilayer perceptron} (MLP)\footnote{Feedforward neural networks are sometimes loosely referred to as multi-layer perceptrons (MLPs), named after an early artificial neuron model called the \textit{perceptron}. However, perceptrons use a hard threshold activation function while modern MLPs can use any differentiable activation, so they are often not perceptrons, in the strict meaning of the word.} is depicted in Figure \ref{feedforward_nn}.

The figure shows a network with one hidden layer, but it can in principle have any number of hidden layers. The same is true for the number of inputs and outputs.

\subsection{Optimizing neural networks}

Training a deep neural network involves training data and a loss function. For training an artificial neural network, an appropriate loss function has to be found to match both the task at hand along with the final layer's activation function. The loss function measures the difference between the output of the network, when the data is passed through it, and the desired outcome. The parameters $\theta$ of the network are then adjusted toward the optimal $\theta^*$ that minimize the loss function over the data

\begin{equation}
\theta^* = \argminB_\theta \sum_{i=1}^n \mathcal{L}(f_\theta(x_i), y_i)
    \label{general loss}
\end{equation}


\noindent where $n$ is the number of data points and $f_\theta(x_i)$ is the prediction of a neural network with parameters $\theta$, for sample $x_i$ with the true value $y_i$. The quantity $\sum_{i=1}^n \mathcal{L}(f_\theta(x_i), y_i)$ is also known as the \textit{empirical risk}. One such example is the mean-squared error (MSE) loss
\begin{equation}
\mathcal{L}_{\text{MSE}}(f_\theta(x_i), y_i) =  \left( y_i - f_\theta(x_i) \right)^2
    \label{mse0}
\end{equation}

  Maximizing the likelihood of Gaussian data with respect to the parameters of the assumed model that generated the data is equivalent to minimizing the MSE, making it a popular choice for regression tasks (i.e. when the output layer activation is linear or ReLU). 

For classification networks with a logistic or sigmoid activation output, a suitable loss function is the cross-entropy loss function 

\begin{equation}
\mathcal{L}_{\text{CE}}(f_\theta(x_i), y_i) = - \left( \sum_{j=1}^C y_{ij} \cdot \log (f_\theta(x_i) ) \right)
    \label{celoss}
\end{equation}

\noindent where $C$ is the number of classes\footnote{If $C=3$, then the label vector could, for example, take the form $y_1 = (0, 1, 0)$.}. Similarly to MSE, this loss function is also motivated by the fact that minimizing the cross-entropy loss is equivalent to maximizing the likelihood of uniformly distributed i.i.d. data \citep{yao2019negative}. 

So far, the loss functions we have seen require a label $y_i$ as a part of the input. This makes them \textit{supervised} learning losses. Many commonly used loss functions exist that do not require labels, those are called \textit{unsupervised} learning losses. 









Once we have decided on a loss function to minimize, the next step is to choose the optimization algorithm. The most popular ones are implemented in software libraries such as Keras \citep{chollet2015keras}, MXNet \citep{chen2015mxnet}, Tensorflow \citep{abadi2016tensorflow}, Pytorch \citep{NEURIPS2019_9015} and several others. 

The most common way of training deep neural networks is by employing a variation of the \textit{gradient descent} \citep{curry1944method} algorithm. Gradient descent methods take advantage of the fact that a function decreases the fastest in the negative direction of its gradient, converging at a local minimum. 

An algorithm called \textit{backpropagation} \citep{linnainmaa1970representation} computes the gradient of the loss function with respect to each parameter (e.g. weights and biases) via the chain rule from calculus. These gradients are then used for an update step for each parameter:

\begin{equation}
\theta^{[i+1]} \leftarrow \theta^{[i]} - \eta \frac{\partial \mathcal{L}}{\partial \theta^{[i]}}
    \label{gradientdescent}
\end{equation}

\noindent where $i$ keeps track of the index of the iteration. The parameter $\eta$ is known as the \textit{learning rate} of the optimization algorithm. 

The classical gradient descent method in Equation \ref{gradientdescent} calculates the average loss over the entire data set. This can be made faster, without losing convergence guarantees, by performing a weight update using the gradient from only a subset of the training data -- or a \textit{training batch} -- in each iteration. This stochastic approximation of gradient descent is called \textit{stochastic gradient descent.}

A good learning rate is important for the practical convergence of stochastic gradient descent: if it is very small, then the time it takes to converge can be too long. However, if it is too large, then there is a risk of overshooting the local minima. 
It is generally good to start off with a larger learning rate and then make it smaller with time. 

Determining exactly when to decrease the size of the learning rate, and by how much, can be laborious in practice. For this reason, there have been proposed several gradient descent methods that automatically find this learning rate schedule with adaptive learning rates, for example, rmsprop \citep{tieleman2012lecture} and Adam \citep{kingma2014adam}.



\subsection{Convolutional neural networks}

The network in Fig. \ref{feedforward_nn} is a \textit{fully-connected} or \textit{dense} neural network, because every unit is connected to every unit in the preceding layer. There are other, more specialized, neural networks that are not fully-connected, one of the most important class being \textit{convolutional} neural networks. For input data with a spatial structure, for example images, convolutional neural networks are very efficient.

In contrast to dense neural networks, each unit in convolutional neural networks only receives as input a subset of the outputs from the previous layer. More specifically, each unit only receives inputs from units that are in spatial proximity of one another. Waldo Tobler's First Law of Geography captures succinctly the motivation behind convolutional networks \citep{tobler1970computer}: "everything is related to everything else, but near things are more related than distant things". Another key property of convolutional neural networks is the one of shared weights -- each computational unit in the same layer has the same set of weights, even though they receive different inputs. 

These properties have the practical consequence that the number of parameters is cut down substantially: each neuron processing a $64 \times 64$ grayscale image would require $64 \cdot 64 + 1 = 4097$ weights, which is then multiplied again by the number of neurons in the layer for the total parameter count. On the other hand, a convolutional layer where each neuron processes a $7 \times 7$ window (a relatively large window size) around a pixel would require $7\cdot7+1 = 50$ parameters -- for the whole layer, due to the shared weights. Thus, processing the image input in this example with a convolutional layer instead of a dense layer with a single neuron reduces the number of parameters by a factor of $80$. 

\begin{figure}
\centering
\begin{subfigure}[b]{0.95\textwidth}
   \includegraphics[width=1\linewidth]{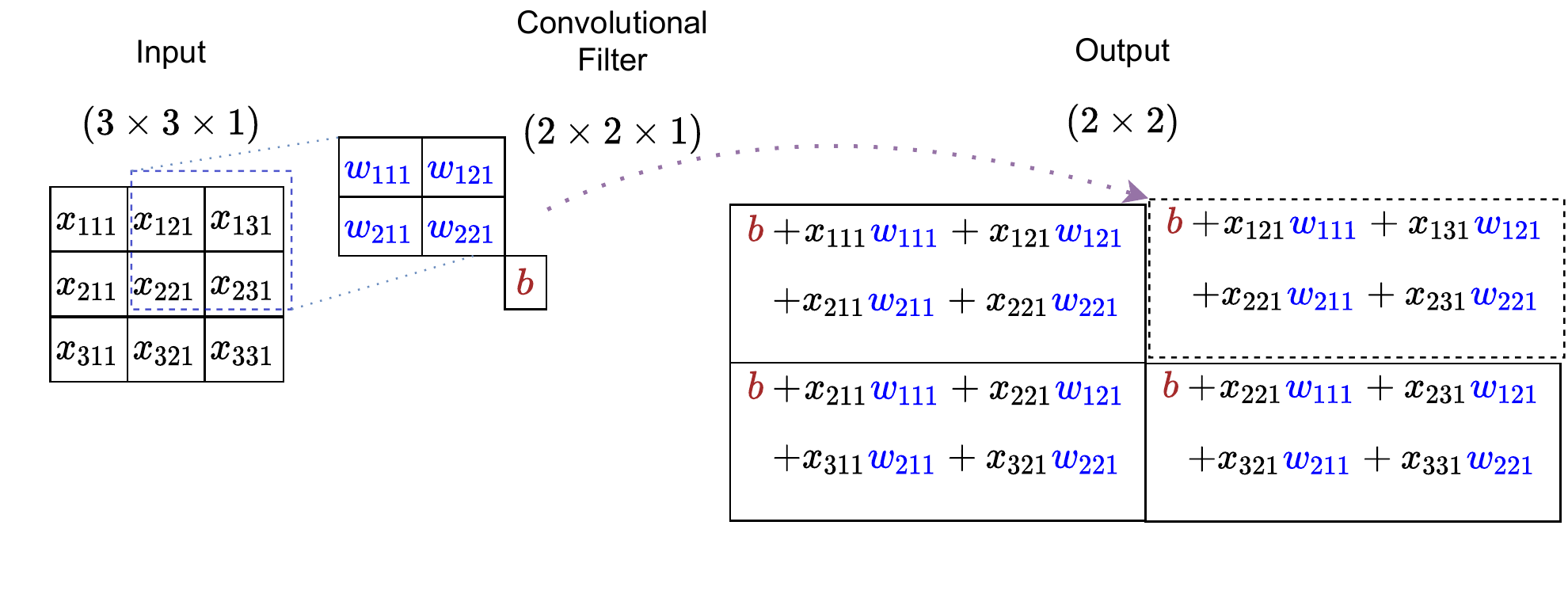}
   \caption{Input with (height $\times$ width  $\times$ depth) dimensions of ($3\times3\times1$). }
   \label{fig:cnn_example_lighter} 
\end{subfigure}

\begin{subfigure}[b]{0.95\textwidth}
   \includegraphics[width=1\linewidth]{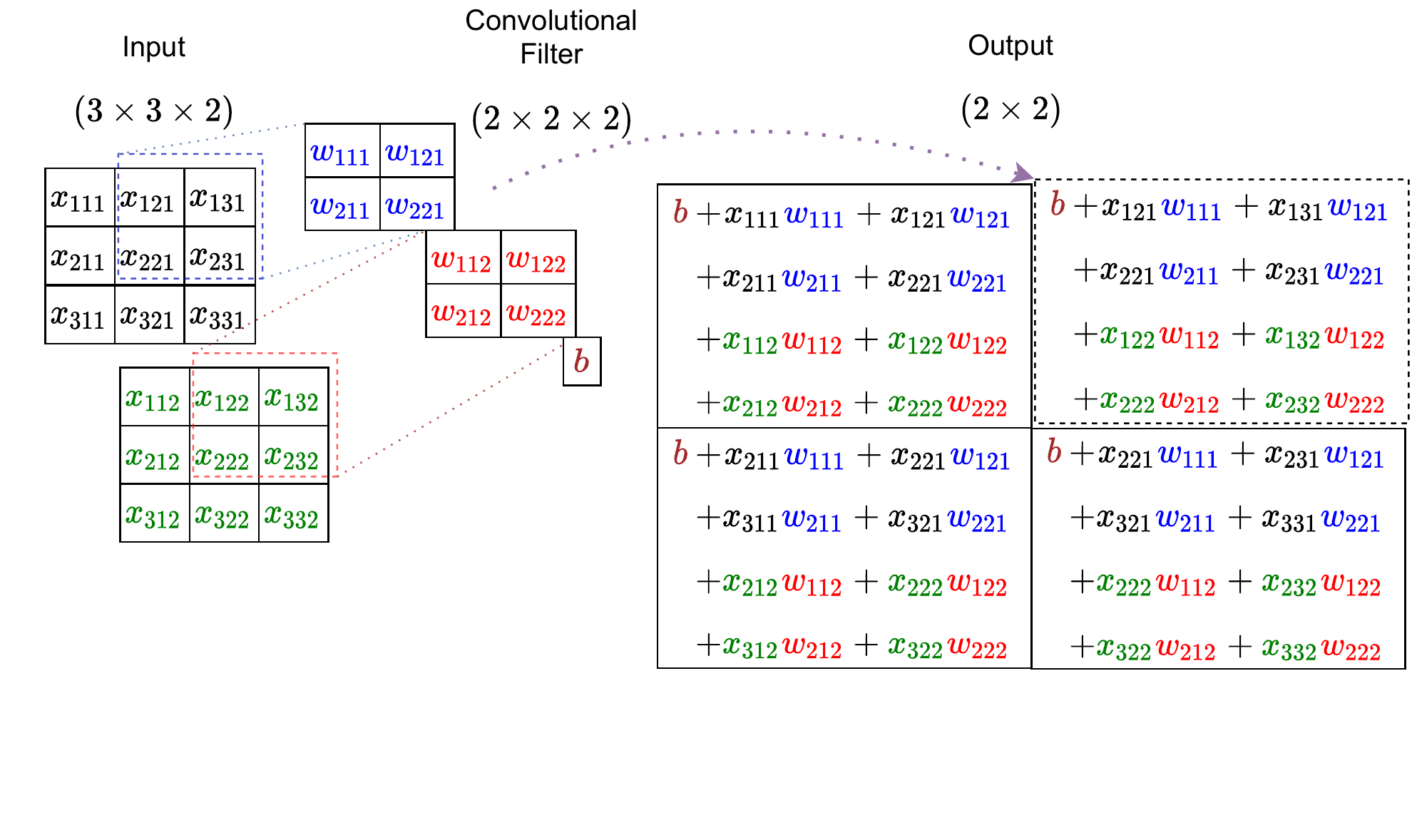}
   \caption{Input with (height $\times$ width  $\times$ depth) dimensions of ($3\times3\times2$). }
   \label{fig:cnn_example}
\end{subfigure}
\caption[Illustration: A convolution layer]{ \textbf{A convolutional layer.} The input is processed by a single convolutional filter with a receptive field of $2\times2$, a stride of one, no zero padding and the identity activation function $\varphi(x)=x$. The subset of the input that is included in the calculation of the upper right element of the output is highlighted by the dotted boxes.}
\end{figure}

In addition to the activation function, we specify the value of four hyperparameters for convolutional layers when we describe specific network architectures in this thesis: the number of \textit{filters} to slide along the height and width of the input, the size, or the \textit{receptive field}, of the filters, the \textit{stride} and how much \textit{zero padding} to use. The receptive field dictates the sizes of the spatial dimensions (height, width) of the input that the neuron takes. In Figure \ref{fig:cnn_example}, for instance, we would say that the filter size is ($2\times2$), despite the dimension of the weights being ($2\times2\times2$) -- this is due to the size of the input depth. Note that even though these height and width values are constrained, the neuron always processes the full depth of the input. 

The stride controls how many steps the filters take as the input is processed along its spatial dimensions. Zero padding amounts to adding rows and columns around the input, composed entirely of zeros, also along the depth. Padding is often done to make the output valid, for instance, if the stride or filter size would potentially cause a neuron to process inputs that are "out of bounds". This is often called \textit{valid} padding. Another purpose of padding is to pad the input with zeros to keep the original spatial dimensions unchanged, this is called \textit{same} padding. For example, zeros can be added around an image of size $12\times12$ before it is processed by a network that requires an input of $16\times16$.

The relationship between the input and output of a convolutional filter is illustrated in Figure~\ref{fig:cnn_example_lighter}. In Figure~\ref{fig:cnn_example} we add a depth dimension. In our example, the stride is 1. However, in the example, if we would want to increase the stride value then we would have to introduce zero padding.

Generally, any deep neural network is called a convolutional neural network if it has one or more convolutional layers. This holds true even if not all the layers are convolutional layers. Some popular layer types include:

\begin{itemize}
    \item \textit{Subsampling} layers that keep only every $n$th row and column to reduce the computational complexity. This is usually only done as a first preprocessing step for very high dimensional inputs, where throwing away the information is not as harmful as in the intermediate layers of the network.
    \item \textit{Max pooling} layers slide along the width and height of each depth slice in the input and return the largest single value in their window. They reduce the computational complexity by reducing the input dimension, and they make the representation approximately invariant to small translations.
    \item \textit{Flattening} layers are technical layers that re-shape tensor or array inputs to vectors.
    \item \textit{Normalization} layers for re-centering and re-scaling inputs to layers. They help speeding up and stabilizing the learning process.
\end{itemize}

Deep neural networks often have a number of parameters in the thousands or billions. This makes the interpretation of the calculations difficult, especially due to the number of layers. Visualizations of the first few convolutional layers in trained networks has been done \citep{zeiler2014visualizing}, with the result that the first layer's filters usually capture edges, corners, and color combinations. The second layer then combines these features into more complicated patterns. Higher filters then combine these features further into textures, object parts or even whole objects.

\section{Representation learning}
\label{sec:repr_learn}

In computer science in general, and machine learning in particular, the choice of the representation of the data that is being processed is crucial. This could mean choosing the right data structure for the task, such as designing a database for fast searching. This could also mean choosing the right independent variables for a statistical model.

The extraction of useful information about the data is thus an important task. This is especially true if the input is from a high-dimensional space, with the term \textit{curse of dimensionality}  \citep{bellman1957dynamic} being used since the late 1950s for describing problems of this nature: the amount of data needed to make statistically significant claims grows exponentially with the dimensionality of the space that the data resides in. This makes the discovery of methods for reducing the dimensionality of the input, without discarding important information, an attractive prospect.
\subsection{Supervised representation learning}

Representations arise in artificial neural networks (ANNs) when they are trained for a regression or classification objective. One view of ANNs is that the hidden layers perform feature extraction on the input, transforming it to a more suitable form for the output layer that performs the final calculations for the task. \cite{sharif2014cnn} made use of this insight by pre-processing inputs for supervised learning models with the intermediate layer outputs of a convolutional network that was pre-trained on an object classification task. They achieved impressive results on a diverse range of tasks, such as image retrieval and scene recognition. The hierarchical structure of ANNs also has the theoretical and practical advantage that the features at each level are re-used for the different features at the higher level.

\subsection{Unsupervised representation learning}
For hierarchical methods like ANNs, representations are generated at the same time as the whole system is trained to minimize error on human tagged data. Most other representation learning methods are \textit{unsupervised} and are able to learn useful features on unlabeled data.

\subsubsection{PCA}

Principal component analysis (PCA) is an unsupervised learning method invented by \cite{pearson1901liii}. The method finds an orthogonal linear transformation $f(X) = W^TX$ for zero-mean, $d$-dimensional data $X$. 

The columns $w_i$ of $W$ are the \textit{principal components} of the data $X$, which point to the direction of the greatest variance in the data. The first principal component is the solution to the equation

\begin{equation} \label{eq:pca}
w_i = \argmaxB_{||w||=1} ||Xw||^2
\end{equation}

The second component is found by applying Equation \ref{eq:pca} again to the transformed data $\hat{X}$ that is given by removing the contribution of the first component from $X$, $\hat{X} = X - ((w_1)^TX)w_1$, and so on.

All the components can also be found simultaneously by computing the eigendecomposition of the data's covariance matrix, as it has been shown that the principal components are equal to the resulting eigenvectors \citep{shlens2014tutorial}.

\begin{figure}[h]
\begin{adjustwidth}{0in}{0in}
    \centering
    
    \subfloat[{\bf Samples from the digits database.}]{{\includegraphics[width=.47\columnwidth]{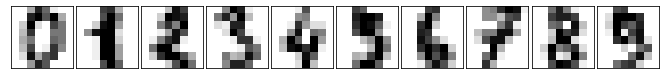}}}

        \vspace{1em}

    \subfloat[{\bf PCA visualization of digits.}]{{\includegraphics[width=.45\columnwidth]{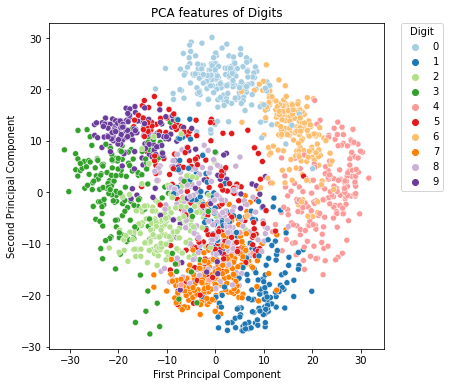}}}
    \qquad
    \subfloat[{\bf t-SNE visualization of digits.}]{{\includegraphics[width=.45\columnwidth]{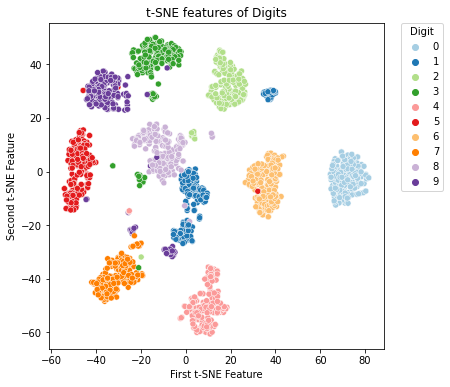}}}

    \caption[Example: Dimensionality reduction by PCA and t-SNE]{\textbf{Dimensionality Reduction by PCA and t-SNE} (a) The digits database consists of $8 \times 8$ pixel images of hand-drawn digits. (b) We plot the first two principal components of the digits data set and color by digit label. (c) We visualize the two-dimensional embedding of the digit data set as learned by t-distributed stochastic neighbor embedding. }
    \label{fig:vizcomparison}
\end{adjustwidth}
\end{figure}

Dimensionality reduction can be done by creating a matrix $W_L$, consisting only of the first $L$ principal components, and applying it to the data. This yields the lower-dimensional, transformed data matrix $T_L = W_{L}^T X $. By doing this, the data is projected onto the subspace with the maximum variance. This matrix $W_L$ of principal components has the property of minimizing the reconstruction error $||X -  W_L T_L  ||^2$. In Figure \ref{fig:vizcomparison}, we show a visualization of the UCI ML digits data set, which consists of $8\times8$ grayscale images of hand-written digits. The figure shows the result after the data is projected on its first two principal components, showing clear clustering of the digits. We also show the clustering found by t-distributed stochastic neighbor embedding (discussed below), which separates the clusters more cleanly for this data set.

\subsubsection{t-SNE}

Over the last few years, t-distributed stochastic neighbor embedding (t-SNE) \citep{maaten2008visualizing} has become one of the most popular dimensionality reduction techniques for visualization \citep{arora2018analysis}. The assumption behind the algorithm is that the high-dimensional input data lies on a locally connected manifold. First, an auxiliary asymmetric measure between each pair of data points is calculated according to the equation

\begin{equation} \label{eq:tsne}
p_{j|i} = \frac{\exp(-||x_i-x_j||^2 / 2 \sigma^2_i)}{\sum_{k\neq i}\exp(-x||x_i-x_k||^2/2\sigma^2_i)}
\end{equation}

\noindent where $p_{i|i} = 0$ as it is of primary interest to model pairwise similarities. The constant $\sigma_i$ is the variance of a Gaussian that is centered around $x_i$ and controls how influential nearby data points are in contrast to far away data points. Then the pairwise similarities are computed using the formula

\begin{equation} \label{eq:tsne2}
p_{ji} = \frac{p_{j|i}+p_{i|j}}{2}
\end{equation}

\noindent these similarities are defined to be symmetrized conditional probabilities to ensure that each data point makes a significant contribution to the cost function.

Next, a lower-dimensional vector of data points Y is created and initialized randomly. Each element in Y corresponds to an element in X: similarities $q_{ij}$ between data points $y_i$ an  $y_j$ are calculated according to the formula

\begin{equation} \label{eq:tsne3}
q_{ij} = \frac{(1+||y_i-y_j||^2)^{-1}}{\sum_k \sum_{l \neq k} (1 + ||y_k - y_l ||^2)^{-1}}\end{equation}


The low-dimensional data points $y_i$ are then moved around to minimize the KL divergence -- a measure of the difference between two probability distributions\footnote{Note that $q$ and $p$ can be interpreted as probabilities since $\sum_{i, j} p_{ij} = \sum_{i, j} q_{ij} = 1$ and $p_{ij} > 0$ and $q_{ij} > 0$ for all $i$ and $j$. } -- between $p$ and $q$

\begin{equation} \label{eq:tsne4}
\text{KL} \left( P~\middle\|~ Q\right) = \sum_{i \neq j }\log \frac{p_{ij}}{q_{ij}}
\end{equation}

Equation \ref{eq:tsne4} is minimized using gradient descent, ensuring that points that are similar in the high-dimensional space are also similar in the new, low-dimensional space.

\subsubsection{Autoencoders}

\begin{figure*}[ht]
		\centering
	\includegraphics[width=\textwidth]{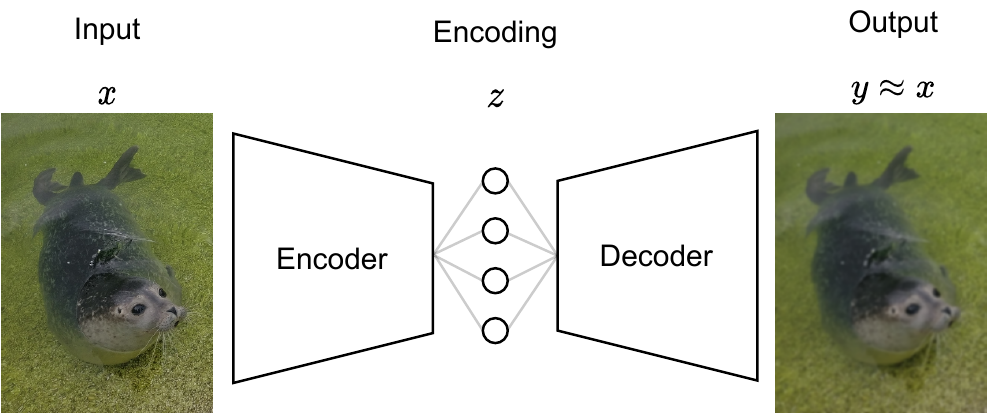}
		\caption[Illustration: An autoencoder]{ \textbf{An autoencoder. } This is an example of a convolutional autoencoder that is trained for reconstructing seal photographs. The encoding vector $z$ is usually much smaller than the total number of pixels in the input.}	\label{snorri}
\end{figure*}

The \textit{autoencoder} is a type of neural network\footnote{Autoencoders can consist of any types of neural network layers. For example, an autoencoder made up of convolutional layers is called a $\textit{convolutional autoencoders}$. } that consists of an \textit{encoder} part, which maps the input $x$ to an encoding (usually of a smaller size than the input), and a \textit{decoder} part, that outputs a reconstruction $y$ of the input (Figure \ref{snorri}). 

The objective function is the squared error between the input and the output 

\begin{equation} \label{eq:ae}
\mathcal{L}(x, y) = || x - y ||^2
\end{equation}

Useful representations arise in this process if the correct constraints are placed on the system. Without constraints, the system could end up learning the identity function, that trivially satisfies the objective function: $||x-y|| = ||x-x|| = 0$. This problem can be overcome by including a hidden layer in the autoencoder of a lower dimensionality than the input space. In this case, the autoencoder is said to be \textit{undercomplete.} 

Undercomplete, single-layer autoencoders with linear activation functions are almost equivalent to PCA. The $p$-dimensional hidden layer spans the same subspace as the first $p$ principal components, or the principal subspace, of the data \citep{baldi1989neural}. Unlike PCA, however, the weights of the hidden layer are not guaranteed to be orthonormal nor ordered. If the smallest dimensionality of a hidden layer is larger than the size of the input, the autoencoder is said to be \textit{overcomplete}. Overcomplete autoencoders can be prevented from learning the identity function if the objective function (Equation \ref{eq:ae}) is combined with a regularization term.

For example, instead of reconstructing the original input as it is, \cite{vincent2008extracting} propose that the goal could be to recover the input after it has been corrupted with noise (e.g. Gaussian noise or salt-and-pepper noise). The idea behind this is that the autoencoder has to learn representations that are stable and robust under the corruption of the input, and that the denoising task extracts a useful structure of the input distribution \citep{vincent2010stacked}.

\subsection{Self-supervised learning}

Approaches that learn representations by way of solving auxiliary tasks, in the sense that the representation that arises from the optimization is more important than achieving a good performance on the task itself, is sometimes called \textit{self-supervised} learning in the literature \citep{gogna2016semi}. For example, \cite{ha2018world} train an autoencoder to reconstruct the observations in an RL environment, but they do not take advantage of the reconstructive capabilities of the network when they train their RL policies, and they use only the encoder part of the network for visual pre-processing. Another example is the denoising autoencoder from the previous chapter.

 One direction of self-supervised learning that has been followed in the literature is to invent a task that requires an understanding of the domain to solve correctly, such as the reconstructive network used by \cite{ha2018world}. Another example is the work by \cite{gidaris2018unsupervised}, who learn representations by applying random rotations to natural images and train a network to predict which rotation was applied. This encourages the network to learn high-level concepts, such as beaks, wings and talons, and their relative positions. More commonly, self-supervised learning methods consist of obscuring some part of the input and train a model to predict that part given some other subset of the input, as we do in Chapter \ref{chap:larp} and Chapter \ref{chap:rewpred}. Some variations of this idea include, for example, colorization \citep{zhang2016colorful}, where colorful images are converted to grayscale with the goal of predicting the original colors. 
\clearpage

\chapter{Learning gradient-based ICA by neurally estimating mutual information}
\label{chap:grica}
In this chapter\footnote{This chapter is adapted from \citep{hlynsson2019learning}, which was published in the Joint German/Austrian Conference on Artificial Intelligence (Künstliche Intelligenz).}, we introduce a novel method of training neural networks in an unsupervised manner to output statistically independent components, a method we call GrICA. We use a mutual information neural estimation (MINE) network \citep{belghazi2018mine} to guide the learning of an encoder to produce statistically independent outputs. This is a recent method of estimating the mutual information of random variables in a deep learning setting, and we apply it to get a qualitatively equal solution to FastICA on blind-source-separation of noisy sources. We investigate the usefulness of our method in contrast to a representation learned by a convolutional autoencoder for preprocessing visual inputs for an RL agent, but the comparison is unfavorable for our approach.

The rest of this chapter has the following organization: Section~\ref{icaintroduction} motivates the design of representations with independent components. Section~\ref{icabackground} explains the ICA problem formulation. Section~\ref{icarelatedwork} briefly discusses related work. Section \ref{icamethod} introduces our method of using a mutual information neural estimator to teach a neural network to output independent components. Section \ref{icaresults} shows the experimental evaluation of our method. Finally, we conclude with a discussion in Section~\ref{icaconclusion}.

\section{Introduction}
\label{icaintroduction}

The general objective of training an encoder to learn statistically independent, factorial codes of the data has been called the "holy grail" of unsupervised learning \citep{schmidhuberunsupervised}. We suggest that learning to recover few, statistically independent, latent variables of an RL environment can speed up the training of RL agents. For environments where high-dimensional observations are created from a small set of statistically independent latent variables, this technique could reduce the dimensionality of the observations without discarding unnecessary information.

Another theoretical advantage of using this kind of approach in RL settings, compared to the other methods we develop in this PhD dissertation, is that it requires only out-of-context observation data from the environment. Learning the GrICA representation does not require full transitions $(s, a, r, s')$ tuples\footnote{We use $s$ to denote the state, $a$ to denote the action, $r$ to denote the reward and $s'$ to denote the next state. } and can thus be used when the transition or reward dynamics of the environments change.

Learning representations that output statistically independent features can be done in any number of ways, for example, by trying to make each output as unpredictable as possible given the other output units \citep{schmidhuber1992learning}. We take the approach of minimizing the mutual information, as estimated by a MINE network, between the output units of a differentiable encoder network. This is done by simple alternate optimization of the two networks.

\section{Background}
\label{icabackground}

Independent component analysis (ICA) aims at estimating unknown \textit{sources} that have been mixed together into an \textit{observation}. The usual assumptions are that the sources are statistically independent and no more than one is Gaussian \citep{jutten2003advances}. The now-cemented metaphor is one of a cocktail party problem: several people (sources) are speaking simultaneously, and their speech has been mixed together in a recording (observation). The task is to unmix the recording such that all dialogues can be listened to clearly. 

In linear ICA, we have a data matrix $S$ whose rows are drawn from statistically independent distributions, a mixing matrix $A$, and an observation matrix $X$:

$$X = AS$$

\noindent and we want to find an unmixing matrix $U$ of $A$ that recovers the sources up to a permutation and scaling:

$$ Y = UX $$

The general non-linear ICA problem is ill-posed \citep{hyvarinen1999nonlinear, darmois1953analyse} as there is an infinite number of solutions if the space of mixing functions is unconstrained. However, post-linear \citep{taleb1999source} (PNL) ICA is solvable. This is a particular case of non-linear ICA where the observations take the form

$$X = f(AS)$$

\noindent where $f$ operates componentwise, i.e. $X_{i, t} = f_i \left( \sum_m^n A_{i, m}S_{m, t}\right) $. The problem is solved efficiently if $f$ is at least approximately invertible \citep{ziehe2003blind} and there are approaches to optimize the problem for non-invertible $f$ as well \citep{ilin2004post}.
For signals with time-structure, however, the problem is not ill-posed even though it is for i.i.d. samples \citep{blaschke2007independent, sprekeler2014extension}. 

To frame ICA as an optimization problem, we must find a way to measure the statistical independence of the output components and minimize this quantity. There are two main ways to approach this: either minimize the mutual information between the sources \citep{amari1996new, bell1995non, cardoso1997infomax}, or maximize the sources' non-Gaussianity \citep{hyvarinen2000independent, blaschke2004cubica}. 

\section{Related work}
\label{icarelatedwork}

There has been an interest in combining neural networks with the principles of ICA for several decades. In Predictability Maximization \citep{schmidhuber1992learning}, a game is played where one agent tries to predict the value of one output component given the others, and the other tries to maximize the unpredictability. More recently, Deep InfoMax (DIM) \citep{hjelm2018learning}, Graph Deep InfoMax \citep{velivckovic2018deep} and Generative adversarial networks \citep{goodfellow2014generative}, utilize the work of Brakel et al. \citep{brakel2017learning} to deeply learn ICA. Our work differs from these adversarial training methods in the rules of the minimax game being played to achieve this: one agent directly minimizes the lower-bound of the mutual information, as derived from the Donsker-Varadhan characterization of the KL-Divergence, as the other tries to maximize it.

\section{Method}
\label{icamethod}

\subsection{Reinforcement learning environment}

\begin{figure*}[htbp]
	\centering\begin{minipage}{.525\columnwidth}\centering
	\includegraphics[width=\textwidth]{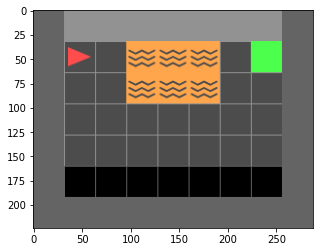}\label{label1icaa}
	\end{minipage}%
	\begin{minipage}{.3525\columnwidth}\centering
	\includegraphics[width=\textwidth]{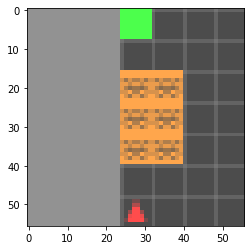}\end{minipage}
	\begin{minipage}{.525\columnwidth}\centering
	\includegraphics[width=\textwidth]{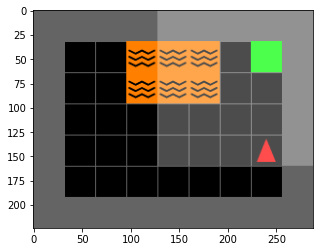}\label{label1icab}
	\end{minipage}%
	\begin{minipage}{.3525\columnwidth}\centering
	\includegraphics[width=\textwidth]{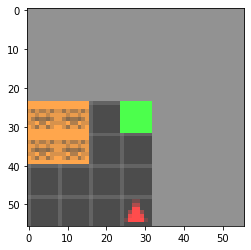}    \end{minipage}\caption[Example: The lavafield environment]{ \textbf{The lavafield environment.} The world has $5\times7$ tiles and is surrounded by impassable walls. The agent (red arrow) is tasked with reaching the goal, represented by the green tile, which yields a reward and terminates the episode. The episode ends without reward if the agent touches an orange lava tile. The full world state can be seen on the right, with a slightly lighter box containing the agent. This box highlights the subset of the world that is perceived by the agent, seen on the left.}\label{clusteringz}
\end{figure*}

Our representation is tested on a 2D environment where the agent is supposed to avoid a field of lava and reach a goal on the other side of the room. The full state of the environment is the whole room (Fig \ref{clusteringz}, left) and the observation is an isometric view of the agent and its point of view (Fig \ref{clusteringz}, right). The observations are $56 \times 56$ RGB images and the agent can take a step forward, turn left or turn right.

The episode terminates if the agent steps toward lava or the goal. The agent receives a positive reward if it reaches the goal, but the episode terminates with zero reward if it steps into the lava. There is no change if the agent is faced toward the wall and takes a step forward. 

\subsection{Learning the independent components}

We train an encoder $E$ to generate an output $\left(z_1, z_2, \dots, z_k\right)$ such that any one of the output components is statistically independent of the union of the others, i.e. $P(z_i, \boldsymbol{z_{-i}}) = P(z_i)P(\boldsymbol{z_{-i}})$, where $$\boldsymbol{z_{-i}} := \left(z_1, \dots, z_{i-1}, z_{i+1}, \dots, z_k\right)$$ The statistical independence of $z_i$ and $\boldsymbol{z_{-i}}$ can be maximized by minimizing their mutual information 
\begin{equation} \label{mi_definition}
I\left(Z_i; \boldsymbol{Z_{-i}} \right) =
\int_{z} \int_{\boldsymbol{z_{-i}}} 
P(z_i, \boldsymbol{z_{-i}}) \log \left( 
\frac{P(z_i, \boldsymbol{z_{-i}})}{P(z_i)P(\boldsymbol{z_{-i}})} \right) dz_i d\boldsymbol{z_{-i}} 
\end{equation}

 This quantity is hard to estimate, particularly for high-dimensional data. Note that Equation~\ref{mi_definition} can be more succinctly as the \textit{KL divergence} between $Z_i$ and $Z_{-i}$:
 
 \begin{equation} \label{mi_definition_kld}
I\left(Z_i; \boldsymbol{Z_{-i}} \right) = D_{KL} \left( P (Z_i, \boldsymbol{Z_{-i}}) \vert| P(Z_i) P (\boldsymbol{Z_{-i}})  \right) 
\end{equation}

\cite{donsker1975asymptotic} famously proved that the KL Divergence admits the representation

 \begin{equation} \label{dual_kld}
D_{KL} \left( X \vert| Y  \right)  = \sup _{T: \Omega \rightarrow \mathcal{R}}  \mathbb{E}_X[T] - \log (  \mathbb{E}_Y [e^T])
\end{equation}

\noindent where the domain $\Omega$ is a closed and bounded subset of $\mathbb{R}^d$.

\cite{belghazi2018mine} introduce a method of using the Donsker-Varadhan representation to estimate mutual information with neural networks, with an architecture they call mutual information neural estimation (MINE) networks. 


 To learn representations with independent components, we therefore estimate the lower bound of Eq.\ (\ref{mi_definition}) using a MINE network $M$: 
\begin{equation} \label{mine_objective}
I\left(Z_i; \boldsymbol{Z_{-i}} \right) \geq L_i = \mathbb{E}_{\mathbb{J}} \left[M \left( z_i, \boldsymbol{z_{-i}} \right) \right] - \log \left( \mathbb{E}_{\mathbb{M}} \left[ e^{M \left( z_i, \boldsymbol{z_{-i}} \right) } \right]  \right)
\end{equation}

\noindent where $\mathbb{J}$ indicates that the expected value is taken over the joint and similarly $\mathbb{M}$ for the product of marginals. The networks $E$ and $M$ are parameterized by $\theta_E$ and $\theta_M$. The encoder takes the observations as input and the MINE network takes the output of the encoder as an input.

The $E$ network minimizes $L := \sum_i L_i$ in order for the outputs to have low mutual information and therefore be statistically independent. In order to get a faithful estimation of the lower bound of the mutual information, the $M$ network maximizes $L$. Thus, in a push-pull fashion, the system as a whole converges to independent output components of the encoder network $E$. 
In practice, rather than training the $E$ and $M$ networks simultaneously it proved useful to train $M$ from scratch for a few iterations after each iteration of training $E$, since the loss functions of $E$ and $M$ are at odds with each other. When the encoder is trained, the MINE network's parameters are frozen and \textit{vice versa.}

\begin{figure}[htbp] 
\centering
 \captionsetup{width=.99\linewidth}
  \resizebox*{0.99   \textwidth}{!}{\includegraphics
{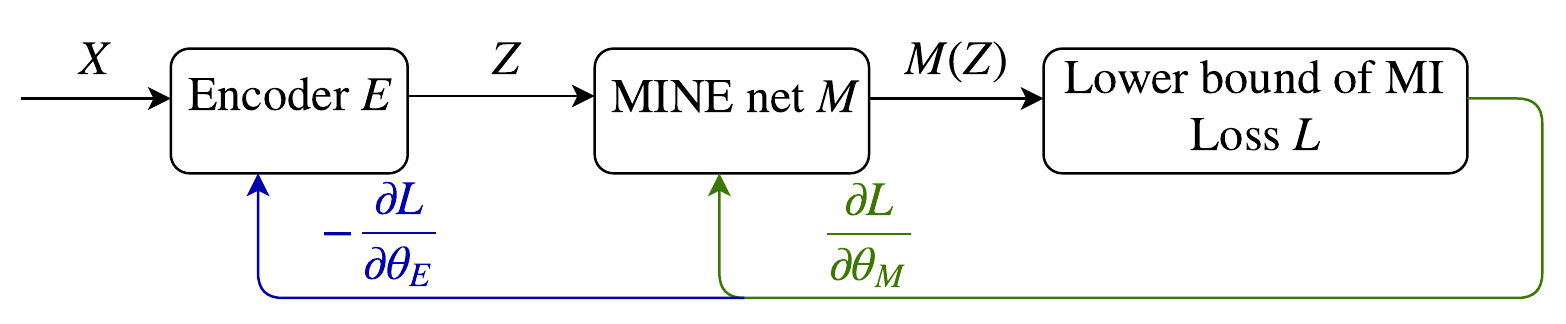}}
\caption[Illustration: Our independent feature learning system]{\textbf{Our independent feature learning system.} The system learns statistically independent outputs by alternate optimization of an encoder $E$ and a MINE network $M$ parameterized by $\theta_E$ and $\theta_M$. The MINE objective (Eq.\ \ref{mine_objective}) is minimized with respect to $\theta_E$ for weight updates of the encoder, but it is \textit{maximized} with respect to $\theta_M$ for weight updates of the MINE network.}
\label{our_method}
\end{figure}

\section{Results}
\label{icaresults}

We try our method on two scenarios: (1) we compare it to canonical implementations of ICA on a textbook example of estimating sources from noisy data and (2) we use our method with a more complex function approximator for preprocessing observations in an RL setting.

\subsection{Recovering noisy signals}

\begin{figure}[h]
\centering
 \captionsetup{width=.98\linewidth}
\begin{subfigure}[b]{.49\linewidth}
\includegraphics[width=\linewidth]{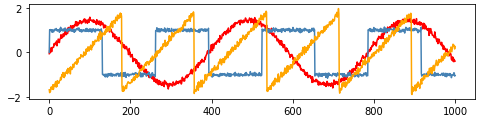}
\caption{The original sources.}\label{sources}
\end{subfigure}
\begin{subfigure}[b]{.49\linewidth}
\includegraphics[width=\linewidth]{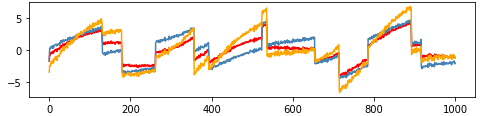}
\caption{Linear mixture of sources.}\label{mixed}
\end{subfigure}

\begin{subfigure}[b]{.49\linewidth}
\includegraphics[width=\linewidth]{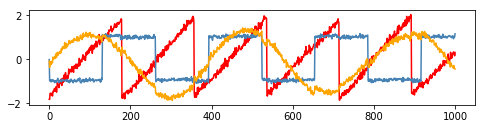}
\caption{Sources recovered by our method.}\label{ours_bss}
\end{subfigure}
\begin{subfigure}[b]{.49\linewidth}
\includegraphics[width=\linewidth]{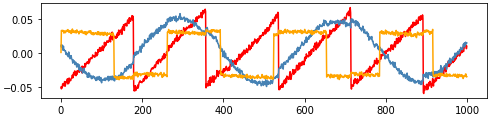}
\caption{Sources recovered by FastICA.}\label{fastica_bss}
\end{subfigure}
\caption[Result: Noisy signal recovery]{\textbf{Noisy signal recovery.} Three independent, noisy sources (a) are mixed linearly (b). Our method recovers them (c) to the same extent as FastICA (d).}
\label{fig:animals}
\end{figure}

We validate the method\footnote{Full code for the noisy signal recovery experiment is available at \url{github.com/wiskott-lab/gradient-based-ica/blob/master/bss3.ipynb}} a for linear noisy ICA example \citep{sklearn}.
Three independent, noisy sources --- sine wave, square wave and saw tooth signal (Fig.\ \ref{sources}) --- are mixed linearly (Fig.\ \ref{mixed}): 

$$ Y =   \begin{bmatrix}
    1 & 1 & 1 \\
    0.5 & 2 & 1\\
    1.5 & 1 & 2 
  \end{bmatrix} S$$
  \\
The encoder is a single-layer neural network with linear activation, with a differentiable whitening layer \citep{schuler2018gradient} before the output. The whitening layer is a key component for performing successful blind source separation for our method. Statistically independent random variables are necessarily uncorrelated, so whitening the output by construction beforehand simplifies the optimization problem significantly. 

The MINE network $M$ is a seven-layer neural network. Each layer but the last one has 64 units with a rectified linear activation function. Each training epoch of the encoder is followed by seven training epochs of $M$. Estimating the exact mutual information is not essential, so few iterations suffice for a good gradient direction.

Since the MINE network is applied to each component individually, to estimate mutual information (Eq.\ \ref{mine_objective}), we need to pass each sample through the MINE network $n$ times --- once for each component. Equivalently, one could conceptualize this as having $n$ copies of the MINE network and feeding the samples to it in parallel, with different components singled out. Thus, for sample $(z_1, z_2, \dots, z_n)$ we feed in $(z_i ; z_{-i})$, for each $i$. Both networks are optimized using Nesterov momentum ADAM \citep{dozat2016incorporating} with a learning rate of $0.005$. 

For this simple example, our method (Fig.\ \ref{ours_bss}) is equivalently good at unmixing the signals as FastICA as implemented in the scikit-learn package \citep{scikit-learn} (Fig.\ \ref{fastica_bss}). Note that, in general, the sources can only be recovered up to permutation and scaling.

\subsection{Lavafield environment}

For these experiments, we learn our ICA features using a convolutional neural network. We roll out 100 episodes with a fully random policy to gather data for learning the independent features: an agent is placed in the upper right corner of the environment and turns left, right or takes a step forward with equal probabilities. The result observations are then gathered until the episode terminates. This gives us 4130 observations to train the representation on for this experiment. 

The representation we learn is 32-dimensional. The MINE network is the same as above, but the encoder network is a five-layer convolutional neural network. The first two layers are convolutional layers each with 32 filters, a rectified linear unit activation and no padding. The first layer has a stride of 4 and the second one a stride of 3. The output of layer 2 is then passed to a flattening layer, reshaping the tensor output to a vector input for a linear dense layer with 32 units. The output of the dense layer is then finally passed to a sphering layer, giving us the encoding. The network description is summarized in Table~\ref{tab:icarepnet}.

\begin{table}[ht]
    \centering \begin{tabular}{|c c c c c c c|} 
     \hline\rule{0pt}{2.2ex}
     \textbf{Layer} & \textbf{Filters} & \textbf{Kernel} & \textbf{Stride}  & \textbf{Padding} & \textbf{Output }  & \textbf{Learnable} \\
     \textbf{} & \textbf{} & \textbf{} & \textbf{}  & \textbf{} & \textbf{Shape}  & \textbf{Parameters} \\ [0.5ex] 
     \hline\rule{0pt}{2.2ex}
      Input & - & -  &-&-&$56{\times}56{\times}5$ & 0 \\[1ex]
 \hline \rule{0pt}{2.2ex}
      Conv. & 32 & 3x3  & 4 & None & $14{\times}14{\times}32$ & 896 \\[.5ex] 
      ReLU & - & -  & -&- & $4{\times}4{\times}32$ & 0 \\[.5ex]

     \hline \rule{0pt}{2.2ex}
     Conv.     & 32 & 3x3  & 3&None & $4{\times}4{\times}32$ & 9248  \\[.5ex]
       ReLU & - & -  & -&-& $4{\times}4{\times}32$ & 0\\[.5ex] 
       \hline\rule{0pt}{2.2ex}
       Flatten & - & -  & - &-& 512 & 0\\[.5ex] 
       Dense & - & -  & - &-&32 & 16416\\[.5ex]
      ReLU & - & -  & -&-& $32$ & 0\\[.5ex] 
            \hline \rule{0pt}{2.2ex}
Sphering & - & -  & - &-&32 & 0\\[.5ex] 
    \hline
    \end{tabular}
    \vspace{0.1cm}
    \caption[Result: Our convolutional ICA network]{\textbf{Our convolutional ICA network.} The layout is inspired by the topology of Mnih's deep Q networks \citep{mnih2013playing} except for the last layer, where we have a differentiable sphering operation instead of a dense layer.}
    \label{tab:icarepnet} 
\end{table}

The training of our ICA representation follows the same scheme as before: we train the estimator $M$ for seven epochs after each training epoch of the encoder. We trained the encoder for 100 epochs and the estimator for 700 epochs for this experiment.

Our trained representation is used to preprocess the visual input for a RL agent. We choose Actor Critic using Kronecker-Factored Trust Region (ACKTR) as implemented by Stable Baselines with default parameters and model. The ACKTR default model is a fully-connected neural network with two layers of 64 units each and a tanh activation function. 

\begin{figure*}[ht]
	\centering
	\begin{minipage}{.83\columnwidth}
		\centering
	\includegraphics[width=\textwidth]{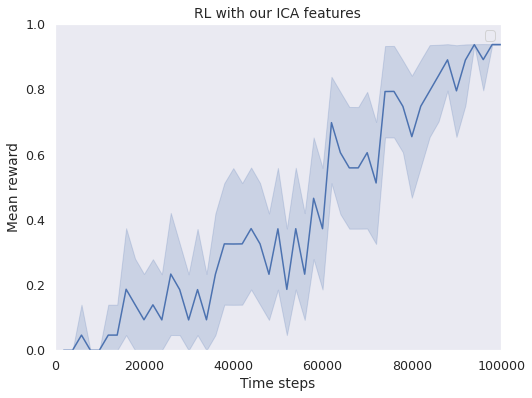}
	\end{minipage}%
		\caption[Result: Reward during training on the lavafield environment]{ \textbf{GrICA reward during training on the lava field environment.} The line indicates the average reward (every 1500 training steps) over 20 different agents trained from scratch. The error bands indicate one standard deviation from the mean. }	\label{rewzies}
\end{figure*}

We trained an ACKTR model from scratch twenty times on our ICA representation, and show the results in Fig~\ref{rewzies}. This indicates that we are able to learn the environment using our method to preprocess the input for a reinforcement learning method. 

\begin{figure*}[ht]
	\centering
	\begin{minipage}{.33\columnwidth}
		\centering
	\includegraphics[width=\textwidth]{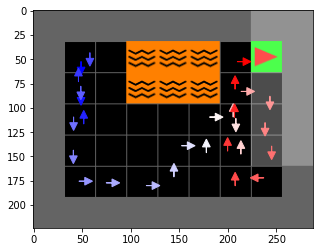}
	\end{minipage}%
	\hspace{-0.45em}
	\begin{minipage}{.33\columnwidth}
		\centering
	\includegraphics[width=\textwidth]{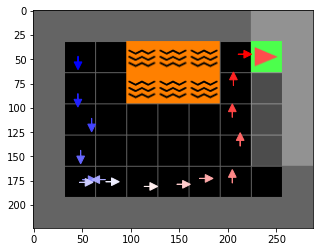}
	\end{minipage}
		\hspace{-0.45em}
	\begin{minipage}{.33\columnwidth}
		\centering
	\includegraphics[width=\textwidth]{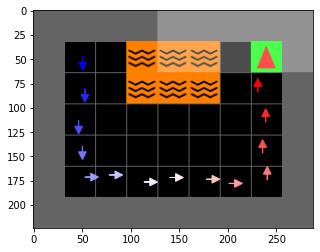}
	\end{minipage}
	\begin{minipage}{.33\columnwidth}
		\centering
	\includegraphics[width=\textwidth]{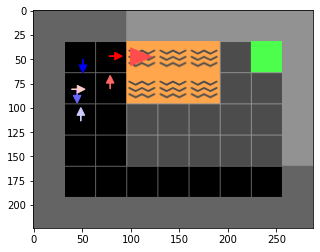}
	\end{minipage}%
	\hspace{-0.45em}
	\begin{minipage}{.33\columnwidth}
		\centering
	\includegraphics[width=\textwidth]{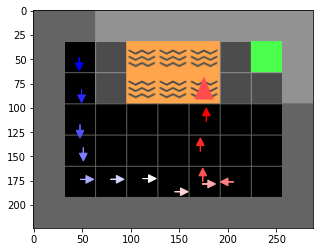}
	\end{minipage}
		\hspace{-0.45em}
	\begin{minipage}{.33\columnwidth}
		\centering
	\includegraphics[width=\textwidth]{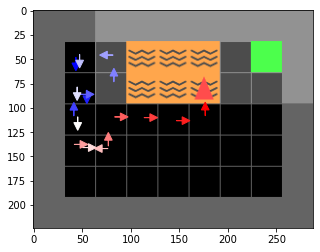}
	\end{minipage}

	\begin{minipage}{.25\columnwidth}
		\centering
	\end{minipage}%
	\hspace{-0.45em}
	\begin{minipage}{.49\columnwidth}
		\centering
	\includegraphics[width=\textwidth]{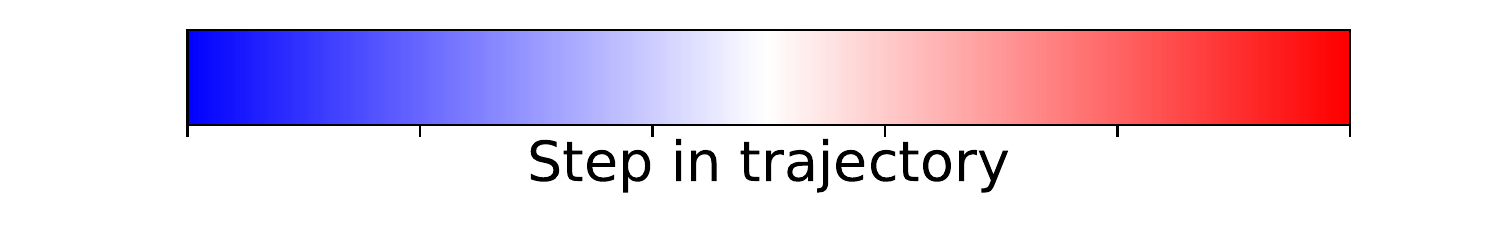}
	\end{minipage}
		\hspace{-0.45em}
	\begin{minipage}{.25\columnwidth}
		\centering
	\end{minipage}	

		\caption[Result: Trajectories in the lavafield environment]{ \textbf{Trajectories in the lava field environment.} The first step in the trajectory is indicated by blue, then the color warms up with each step until it becomes a more saturated red color in the final step.   }	\label{trajectories1}
\end{figure*}

To visualize the behavior of our agent (Figure~\ref{trajectories1}), we choose three successful episodes from a fully-trained model after 100 thousand time steps of training and three unsuccessful ones from a model with 80 thousand time steps of training. It is noteworthy that the agent prefers a wide margin between itself and the lava field as it passes it, even though a more optimal strategy would have the agent walk to the right with the lava left immediately on its left-hand side. The agent also sometimes doubles back before continuing toward the goal again. 

\begin{figure*}[ht]
	\centering
	\begin{minipage}{.33\columnwidth}
		\centering
	\includegraphics[width=\textwidth]{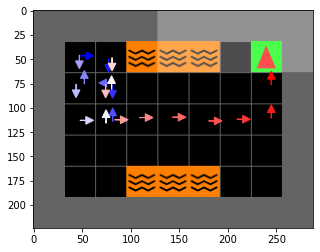}
	\end{minipage}%
	\hspace{-0.45em}
	\begin{minipage}{.33\columnwidth}
		\centering
	\includegraphics[width=\textwidth]{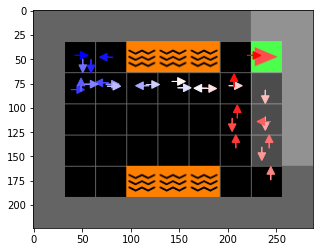}
	\end{minipage}
		\hspace{-0.45em}
	\begin{minipage}{.33\columnwidth}
		\centering
	\includegraphics[width=\textwidth]{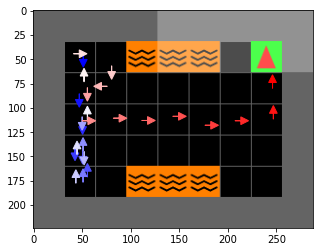}
	\end{minipage}
		\begin{minipage}{.33\columnwidth}
		\centering
	\includegraphics[width=\textwidth]{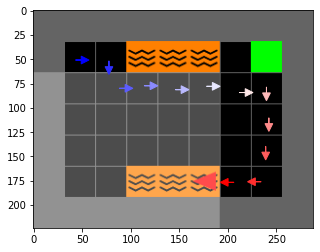}
	\end{minipage}%
	\hspace{-0.45em}
	\begin{minipage}{.33\columnwidth}
		\centering
	\includegraphics[width=\textwidth]{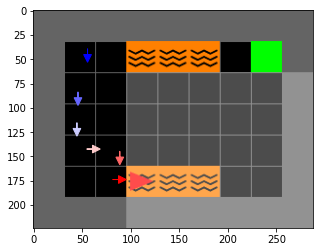}
	\end{minipage}
		\hspace{-0.45em}
	\begin{minipage}{.33\columnwidth}
		\centering
	\includegraphics[width=\textwidth]{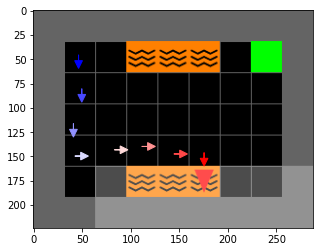}
	\end{minipage}
	
	\begin{minipage}{.25\columnwidth}
		\centering
	\end{minipage}%
	\hspace{-0.45em}
	\begin{minipage}{.49\columnwidth}
		\centering
	\includegraphics[width=\textwidth]{ica_assets/cmap.pdf}
	\end{minipage}
		\hspace{-0.45em}
	\begin{minipage}{.25\columnwidth}
		\centering
	\end{minipage}	
		\caption[Result: Trajectories with shifted lava fields]{ \textbf{Trajectories with shifted lava fields.} This variant punishes strategies -- such as the one learned using our representation -- where the agent seeks to go all the down for safety as it goes across the room. }	\label{failfield}
\end{figure*}

We also tried to see whether our method generalizes to a variant of the environment where the lower row of lava is moved to the bottom, punishing our strategy. There were only 3 successes in a thousand test iterations, shown in Figure~\ref{failfield}, along with three of the unsuccessful episodes.

For comparison, we also trained a convolutional autoencoder (CAE) for reconstruction on the same data set we used to train our ICA representation. We ran the experiment again with the resulting encoding after the CAE was trained. The encoder is the same as the network used for our representation, except that it does not have the sphering layer. 

The decoding portion of the network consists of a 392-unit dense layer whose output is reshaped to a $7\times7\times 8$ tensor and passed to a convolutional layer with 32 filters of size $3\times3$. The output is then up-sampled to quadruple the width and height of the tensor. This is then followed by another convolutional layer, of the same kind as the previous one, and another up-sampling layer that doubles the width and height of the tensor. The output then finally goes through a convolutional layer with 3 filters of size $3\times3$. Each layer in the decoder has a ReLU activation, except for the last which has a logistic activation to reconstruct pixel values that have been normalized lie in the range $[0, 1]$. Each convolutional layer does zero-padding to preserve the width and the height of the input tensor. See Table \ref{tab:caenet} for an overview of the architecture. 
\definecolor{lightblue}{rgb}{0.8,0.85,0.9}
\definecolor{lightred}{rgb}{0.85,0.9,0.8}
\begin{table}[ht]
    \centering \begin{tabular}{|c c c c c c c|} 
     \hline\rule{0pt}{2.2ex}
     \textbf{Layer} & \textbf{Filters} & \textbf{Kernel} & \textbf{Stride}  & \textbf{Padding} & \textbf{Output }  & \textbf{Learnable} \\
     \textbf{} & \textbf{} & \textbf{} & \textbf{}  & \textbf{} & \textbf{Shape}  & \textbf{Parameters} \\ [0.5ex] 
       \rowcolor{lightblue} \hline\rule{0pt}{2.2ex}
    Input & - & -  &-&-&$56{\times}56{\times}5$ & 0 \\[1ex]
 \rowcolor{lightblue}\hline \rule{0pt}{2.2ex}
      Conv. & 32 & 3x3  & 4 & None & $14{\times}14{\times}32$ & 896 \\[.5ex] 
      \rowcolor{lightblue}ReLU & - & -  & -&- & $4{\times}4{\times}32$ & 0 \\[.5ex]

     \rowcolor{lightblue}\hline \rule{0pt}{2.2ex}
     Conv.     & 32 & 3x3  & 3&None & $4{\times}4{\times}32$ & 9248  \\[.5ex]
       \rowcolor{lightblue}ReLU & - & -  & -&-& $4{\times}4{\times}32$ & 0\\[.5ex] 
       \rowcolor{lightblue}\hline\rule{0pt}{2.2ex}
       Flatten & - & -  & - &-& 512 & 0\\[.5ex] 
       \rowcolor{lightblue}Dense & - & -  & - &-&32 & 16416\\[.5ex]
       \rowcolor{lightblue}ReLU & - & -  & -&-& 32 & 0\\[.5ex] 
       \rowcolor{lightred}Dense & - & -  & - &-&392 & 12936\\[.5ex]
       \rowcolor{lightred}ReLU & - & -  & -&-& $392$ & 0\\[.5ex] 
       \rowcolor{lightred}Reshape & - & -  & - &-&$7{\times}7{\times}8$ & 0\\[.5ex]
\rowcolor{lightred}\hline \rule{0pt}{2.2ex}
     Conv.     & 32 & 3x3  & 1&Same & $7{\times}7{\times}32$ & 2336  \\[.5ex]
       \rowcolor{lightred}ReLU & - & -  & -&-& $7{\times}7{\times}32$ & 0\\[.5ex] 
       \rowcolor{lightred}\hline \rule{0pt}{2.2ex}
       Upsampling & - & 4x4  & - & - & $28{\times}28{\times}32$ & 0  \\[.5ex]
\rowcolor{lightred}\hline \rule{0pt}{2.2ex}
     Conv.     & 32 & 3x3  & 1&Same & $28{\times}28{\times}32$ & 9248  \\[.5ex]
       \rowcolor{lightred}ReLU & - & -  & -&-& $28{\times}28{\times}32$ & 0\\[.5ex] 
       \rowcolor{lightred}\hline \rule{0pt}{2.2ex}
       Upsampling & - & 4x4  & - & - & $56{\times}56{\times}32$ & 0  \\[.5ex]
\rowcolor{lightred}\hline \rule{0pt}{2.2ex}
     Conv.     & 3 & 3x3  & 1&Same & $28{\times}28{\times}32$ & 9248  \\[.5ex]
       \rowcolor{lightred}Tanh & - & -  & -&-& $56{\times}56{\times}3$ & 867\\[.5ex] 
    \hline
    \end{tabular}
    \vspace{0.1cm}
    \caption[Result: Convolutional autoencoder network architecture]{\textbf{Convolutional autoencoder network architecture.} The blue part highlights the encoder portion and the green part highlights the decoder portion. The encoder portion was used to preprocess the input to the RL learner for the experiment.}
    \label{tab:caenet}
\end{table}

\begin{figure*}[ht]
	\centering
	\begin{minipage}{.35\columnwidth}
		\centering
	\includegraphics[width=\textwidth]{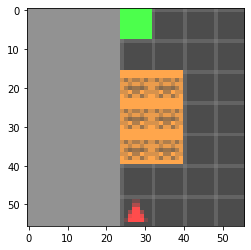}
		\label{icacae1}
	\end{minipage}%
	\begin{minipage}{.305\columnwidth}
		\centering
	\includegraphics[width=\textwidth]{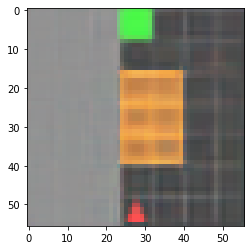}
	\end{minipage}
	\begin{minipage}{.35\columnwidth}
		\centering
	\includegraphics[width=\textwidth]{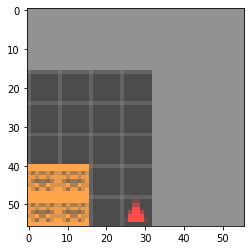}
		\label{icacae3}
	\end{minipage}%
	\begin{minipage}{.305\columnwidth}
		\centering
	\includegraphics[width=\textwidth]{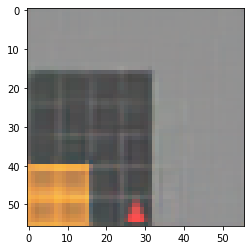}    
	\end{minipage}
		\caption[Result: Reconstruction by autoencoder]{ \textbf{Reconstruction by autoencoder.} The 32-dimensional latent space captures enough information to reconstruct the original observations quite accurately, even after a modest number of training epochs.}
		\label{recons}
\end{figure*}

We train the CAE for 50 epochs. Even though this is a low number of epochs compared to the training for our method, its reconstructive properties are already quite good (Figure \ref{recons}).

\begin{figure*}[ht]
	\centering
	\begin{minipage}{.93\columnwidth}
		\centering
	\includegraphics[width=\textwidth]{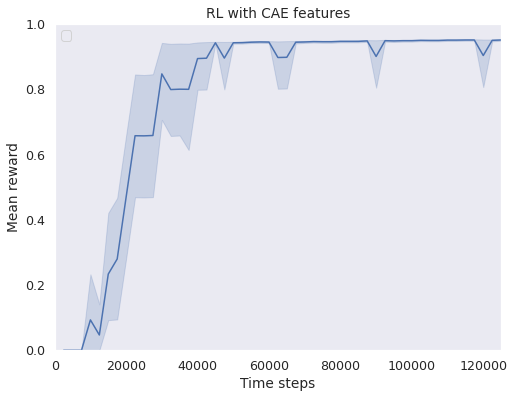}
	\end{minipage}%
		\caption[Result: Reward during training on the lavafield environment (convolutional autoencoder)]{ \textbf{CAE reward during training} The line indicates the average reward (every 1500 training steps) over 20 different agents trained from scratch. The error bands indicate one standard deviation from the mean.  }	\label{stonks2}
\end{figure*}

We repeat the experiment as before, but now with the CAE features instead of our ICA representation. The training curve is shown in Figure \ref{stonks2}. This straightforward baseline algorithm learns to solve the environment twice as fast as our method.

\section{Conclusion} 
\label{icaconclusion}
We have introduced a novel technique for training a differentiable function to perform ICA. The method consists of alternating the optimization of an encoder and a neural mutual information neural estimation (MINE) network. The mutual information estimate between each encoder output and the union of the others is minimized with respect to the encoder's parameters. 

The solution learned by our approach agrees with the one learned by the canonical ICA algorithm, FastICA. An advantage of our method, however, is that it is trivially extended for overcomplete or undercomplete ICA by changing the number of output units of the neural network. We apply our algorithm on high-dimensional data to test the representation learned by our method for dimensionality reduction of visual inputs for an RL agent. The agent is able to use our representation to learn how to solve a simple navigation task, but the preprocessing offered by our approach is outperformed by a convolutional autoencoder.

Our method works in principle, as can be seen by the noisy signal recovery experiment, but its effectiveness for learning representations for RL agents remains unproven. Even though the observations of the lava field environment are fully determined by three latent variables that are statistically independent, the agent's x and y positions, along with its direction, our representation was still not useful enough to beat the relatively simple baselines.

\clearpage

\chapter{Latent representation prediction networks}
\label{chap:larp}

In this chapter\footnote{This chapter is adapted from \citep{hlynsson2020latent}.}, we introduce a  representation learning technique for RL settings that we name Latent Representation Prediction (LARP). This novel system takes advantage of more information given by the environment than our GrICA method from the previous chapter, that only learned from static observations without taking advantage of the knowledge that the system will be used in a dynamic setting. That is to say, we will now utilize the  $(s, a, s') = (\text{state}, \text{action}, \text{next state})$ triplets for training. Our algorithm  learns a state representation, along with a function that predicts how the representation changes when the agent takes given actions in the environment. 

Instead of using our system to preprocess inputs for a model-free reinforcement learner, as we did in the previous chapter, now we take advantage of a prediction function, which is used as a forward model for search on a graph in a viewpoint-matching task. Using a representation that is learned to be maximally predictable for the predictor is found to outperform pretrained representations. The data-efficiency and overall performance of our approach is shown to rival standard reinforcement learning methods, and our learned representation transfers successfully to novel environments.

The rest of the chapter is organized as follows: in Section~\ref{sec:larpintro}, we motivate the usefulness of representations that are predictable in the scope of visual planning, and we mention how we intend to overcome a common pitfall in their design. We move on with discussing the main classes of related work in Section~\ref{sec:larprelatedwork}, and summarize the most relevant articles from among them. In Section~\ref{sec:larpmethods}, we discuss in concrete detail the design of our representation, how we use it for planning, and we introduce an experimental environment of our own design. The results of our experiments are presented in Section~\ref{larpresults}, and we conclude with a discussion of the proposed methodology and prospects for future work in Section~\ref{larpdiscussion}.
\section{Introduction}
\label{sec:larpintro}

Deeply-learned planning methods are often based on learning representations that are optimized for unrelated tasks. For example, they might be trained to reconstruct observations of the environment, such as the convolutional autoencoder from the previous chapter. These representations are then combined with predictor functions for simulating rollouts to navigate the environment. We propose to rather learn representations such that they are directly optimized for the task at hand: to be maximally predictable for the predictor function. This results in representations that are well-suited, by design, for the downstream task of planning, where the learned predictor function is used as a forward model.

While modern reinforcement learning algorithms reach super-human performance on tasks such as game playing, they remain woefully sample inefficient compared to humans. An algorithm that is data-efficient~\citep{hlynsson2019measuring} requires only few samples for good performance and the study of data-efficient control is currently an active research area \citep{corneil2018efficient, buckman2018sample,du2019good,saphal2020seerl}. 

 Dimensionality reduction is a powerful tool for increasing the data-efficiency of machine learning methods. There has been much recent work on methods that take advantage of compact, low-dimensional representations of states for search and exploration~\citep{kurutach2018learning, corneil2018efficient, xu2019regression}. One of the advantages of this approach is that a good representation aids in faster and more accurate planning.
This holds in particular when the latent space is of much lower dimensionality than the state space \citep{hamilton2014efficient}. For high-dimensional inputs, such as image data, a representation function is frequently learned to reduce the complexity for a controller.

\begin{figure}[htp]

\centering
\includegraphics[width=0.9\textwidth]{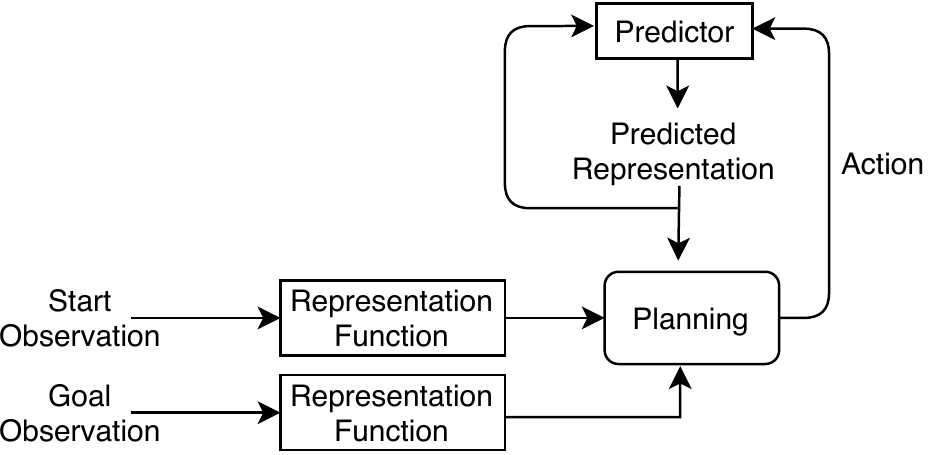}
\caption[Illustration: Latent representation prediction network]{{\bf Conceptual overview of our method.} The important components are the representation network along with the predictor network. Together, they comprise a LARP network, which is utilized by a planning algorithm.}
\label{Conceptual1.pdf}
\end{figure}

In deep reinforcement learning, the representation and the controller are learned simultaneously. Similarly, a representation can in principle be learned along with a forward model for classical planning in high-dimensional space. We do this with our LARP network, which is a neural network-based method for learning a state representation and a transition function for planning within the learned latent space (Fig.~\ref{Conceptual1.pdf}).

During training, the representation and the predictor are learned simultaneously from transitions in a self-supervised manner. We train the predictor to predict the most likely future representation, given a current representation and an action. The predictor is then used for planning by navigating the latent space defined by the representation to reach a goal state.

Optimizing control in this manner, after learning an environment model, has the advantage of allowing for learning new reward functions in a fast and data-efficient manner.
After the representation is learned, we find said goal state by conventional path planning. Disentangling the reward from the transition function in such a way is helpful when learning for multiple or changing reward functions, and aids with learning when there is no reward available at all. Thus, it is also good for a sparse or a delayed-reward setting.

A problem that can arise in representation learning is the one of trivial features. This can happen when the method is optimizing an objective function that has a straightforward, but useless, solution. For example, Slow Feature Analysis (SFA) \citep{wiskott2002slow} has the objective of extracting the features of time series data that vary the least with time. This is easily fulfilled by constant functions, so SFA requires that the representations have a variance of $1$ -- which constant functions cannot fulfill.

Constant features would similarly be maximally predictable representations for our system. Therefore, we study three different approaches to prevent this trivial representation from being learned, we either: \textbf{(i)} design the architecture such that the output is sphered, \textbf{(ii)} regularize it with a contrastive loss term, or \textbf{(iii)} include a reconstruction loss term along with an additional decoder module. 

We compare these approaches and validate our method experimentally on a visual environment: a  viewpoint-matching task using the NORB data set \citep{lecun2004learning}, where the agent is presented with a starting viewpoint of an object and the task is to produce a sequence of actions such that the agent ends up with the goal viewpoint. As the NORB data set is embeddable on a cylinder \citep{hadsell2006dimensionality, schuler2018gradient} or a sphere \citep{wang2018toybox}, we can visualize the actions as traversing the embedded manifold. Our approach compares favorably to state-of-the-art methods on our test bed with respect to data-efficiency, but our asymptotic performance is still outclassed by other approaches.

\section{Related work}
\label{sec:larprelatedwork}

Most of the related work falls into the categories of reinforcement learning, visual planning, or representation learning.  The primary difference between ours and  other model-based methods is that the representation is learned by optimizing auxiliary objectives which are not directly useful for solving the main task.

\subsection{Reinforcement learning}
There are many works in the literature that also approximate the transition function of environments, for instance by performing explicit latent-space planning computations \citep{tamar2016value, gal2016improving, henaff2017model, srinivas2018universal, chua2018deep, hafner2019learning} as part of learning and executing policies. \cite{gelada2019deepmdp} train an RL agent to simultaneously predict rewards as well as future latent states. Our work is distinct from these, as we are not assuming a reward signal during training. 
\cite{ha2018world} combine vision, memory, and controller for learning a model of the world before learning a decision model. A predictive model is trained in an unsupervised manner, permitting the agent to learn policies completely within its learned latent space representation of the environment. The main difference is that they first approximate the state distribution using a variational autoencoder, producing the encoded latent space. In contrast, our representation is learned such that it is maximally predictable for the predictor network.

Similar to our training setup, \cite{oh2015action} predict future frames in ATARI environments conditioned on actions. The predicted frames are used for learning the transition function of the environment, e.g. for improving exploration by informing agents of which actions are more likely to result in unseen states. Our work differs as we are acting within a learned latent space and not the full input space, and our representations are used in a classical planning paradigm with start and goal states instead of a reinforcement learning one. 
\subsection{Visual planning}

We define visual planning as the problem of synthesizing an action sequence to generate a target state from an initial state, and all the states are observed as images.  Variational State Tabulations~\citep{corneil2018efficient} learn a state representation in addition to a transfer function over the latent space. However, their observation space is discretized into a table using a variational approach, as opposed to our continuous representation. A continuous representation circumvents the problem of having to determine the size of such a table in advance or during training. Similarly, \cite{cuccu2018playing} discretize visual input using unsupervised vector quantization and use that representation for learning controllers for Atari games.

Inspired by classic symbolic planning, Regression Planning Networks \citep{xu2019regression} create a plan backward from a symbolic goal. We do not have access to high-level symbolic goal information for our method, and we assume that only high-dimensional visual cues are received from the environment.

Topological memories of the environment are built in Semi-parametric Topological Memories  \citep{savinov2018semi} after being provided with observation sequences from humans exploring the environment. Nodes are connected if a predictor estimates that they are close. The method has problems with generalization, which are reduced in Hallucinative Topological Memories \citep{liu2020hallucinative}, where the method also admits a description of the environment, such as a map or a layout vector, which the agent can use during planning. Our visual planning method does not receive any additional information on unseen environments and does not depend on manual exploration during training.

Causal InfoGAN \citep{kurutach2018learning} and related methods \citep{wang2019learning} are based on generative adversarial networks (GANs) \citep{goodfellow2014generative}, inspired by InfoGAN in particular \citep{chen2016infogan}, for learning a plannable representation. A GAN is trained for encoding start and goal states, and they plan a trajectory in the representation space as well as reconstructing intermediate observations in the plan. Our method is different as it does not need to reconstruct the observations and the forward model is directly optimized for prediction.


\subsection{Prediction-based representation learning}

 In Predictable Feature Analysis \citep{richthofer2015predictable}, representations are learned that are predictable by autoregression processes. Our method is more flexible and scales better to higher dimensions as the predictor can be any differentiable function.

Using the output of other networks as prediction targets instead of the original pixels is not new. The case where the output of a larger model is the target for a smaller model is known as knowledge distillation \citep{bucilua2006model, hinton2015distilling}. This is used for compressing a model ensemble into a single model. \cite{vondrick2016anticipating} learn to make high-level semantic predictions of future frames in video data. Given a current frame, a neural network predicts the representation of a future frame. Our approach is not constrained only to pretrained representations, we learn our representation together with the prediction network. Moreover, we extend this general idea by also admitting an action as the input to our predictor network.
\section{Materials and methods}
\label{sec:larpmethods}
In this work, we study different representations for learning the transition function of a partially observable MDP (POMDP) and propose a network that jointly learns a  representation with a prediction model and apply it for latent space planning. We summarize here the different ingredients of the LARP network -- our proposed solution. More detailed descriptions will follow in later sections.

\textbf{Training the predictor network:} We use a two-stream fully connected neural network to predict the representation of the future state given the current state's representation and the action bridging those two states. The predictor module is trained with a simple mean-squared error term.

\textbf{Handling constant solutions:} The representation could be transferred from other domains or learned from scratch on the task. If the representation is learned simultaneously with an estimate of a Markov decision process's (MDP) transition function, precautions must be taken such that the prediction loss is not trivially minimized by a representation that is constant over all states.  We consider three approaches for tackling the problem: sphering the output, regularizing with a contrastive loss term, and regularizing with a reconstructive loss term.

\textbf{Searching in the latent space:} Combining the representation with the predictor network, we can search in the latent space until a node is found that has the largest similarity to the representation of the goal viewpoint using a modified best-first search algorithm.

\textbf{NORB environment:} We use the NORB data set~\citep{lecun2004learning} for our experiments.  This data set consists of images of  objects
from different viewpoints, and we create viewpoint-matching tasks from the data set. 



\subsection{On good representations}
We rely on heuristics to provide sufficient evidence for a good --- albeit not necessarily optimal --- decision at every time step to reach the goal. Here, we use the Euclidean distance in representation space: a sequence of actions is preferred if their end location is closest to the goal. The usefulness of this heuristics depends on how well and how coherently the Euclidean distance encodes the actual distance to the goal state in terms of the number of actions.

A learned predictor network approximates the transition function of the environment for planning in the latent space defined by some representation. This raises the question: what is the ideal representation for latent space planning? Our experiments show that an openly available, general-purpose representation, such as a pretrained VGG16 \citep{simonyan2014very}, can already provide sufficient guidance to apply such heuristics effectively. Better still are representation models that are trained on the data at hand, for example, uniform manifold approximation and projection (UMAP) \citep{mcinnes2018umap} or variational auto-encoders (VAEs) \citep{kingma2013auto}.

One might, however, ask what a particularly suited representation might look like when attainability is ignored. It would need to take the topological structure of the underlying data manifold into account, so that the Euclidean distance becomes a good proxy for the geodesic distance. One class of methods that satisfy this are spectral embeddings, such as Laplacian Eigenmaps (LEMs) \citep{belkin2003laplacian}. Their representations are smooth and discriminative which is ideal for our purpose. However, they do not easily produce out-of-sample embeddings, so they will only be applied in an in-sample fashion to serve as a control experiment, yielding optimal performance.

\subsection{Predictor network}

As the representation is used by the predictor network, we want it to be predictable. Thus, we optimize the representation learner simultaneously with the predictor network, in an end-to-end manner.

Suppose we have a representation map $\phi$ and a training set of $N$ labeled data tuples
$(X_t = [o_t, a_t], Y_t = o_{t+1})$, where $o_t$ is the observation at time step $t$ and $a_t$ is an action resulting in a state with observation $o_{t+1}$. We train the predictor $f$, parameterized by $\theta$, by minimizing the mean-squared error loss over $f$'s parameters:
\begin{equation}
\underset{\theta}{\text{argmin}} \ \mathcal{L}_{\text{prediction}}(\mathcal{D}, \theta) = \underset{\theta}{\text{argmin}} \  \frac{1}{N} \sum_{t=1}^N \big \Vert \phi(o_{t+1}) - f_\theta(\phi(o_t), a_t) \big \Vert ^2    
\end{equation}
\noindent where $\mathcal{D} = \{ (X_t, Y_t) \}_{t=0}^N $ is our set of training data.

We construct $f$ as a two-stream, fully connected, neural network. Using this predictor we can carry out planning in the latent space defined by $\phi$. By planning, we mean that there is a start state with observation $o_{\text{start}}$ and a goal state with $o_{\text{goal}}$ and we want to find a sequence of actions connecting them.

The network outputs the expected representation after acting. Using this, we can formulate planning as a classical pathfinding or graph traversal problem.

\subsection{Avoiding trivial solutions}

In the case where $\phi$ is trainable and parameterized by $\eta$, the loss for the whole system that only cares about maximizing predictability is
\begin{equation}
\underset{\theta, \eta}{\text{argmin}} \ \mathcal{L}_{\text{prediction}}(\mathcal{D}, \theta, \eta) = \underset{\theta, \eta}{\text{argmin}} \  \frac{1}{N} \sum_{t=1}^N \left(\phi_\eta(o_{t+1}) - f_\theta(\phi_\eta(o_t), a_t) \right)^2
\label{uselessloss}
\end{equation}
for a given data set $\mathcal{D}$. With no constraints on the family of functions that $\phi$ can belong to, we run the risk that the representation collapses to a constant. Constant functions $\phi = c$ trivially yield zero loss for any set $\mathcal{D}$ if $f_\theta$ outputs the input state again for any $a$, i.e $f(\phi(\cdot), a) = \phi(\cdot)$:

Constant representations are optimal with respect to predictability, but they are unfortunately useless for planning, as we need to discriminate different states. This objective is not present in the proposed loss function in Eq.~\eqref{uselessloss} and we must thus add a constraint or another loss term to facilitate differentiating the different states.

There are several ways to limit the function space such that constant functions are not included, for example with decoder \citep{goroshin2015learning} or adversarial \citep{denton2017unsupervised} loss terms. In this work, we do this with \textbf{(i)} a sphering layer, \textbf{(ii)} a contrastive loss, or \textbf{(iii)} a reconstructive loss.

\subsubsection{(i) Sphering the output} 
\begin{figure}[htp]

\centering
\includegraphics[width=1.0\textwidth]{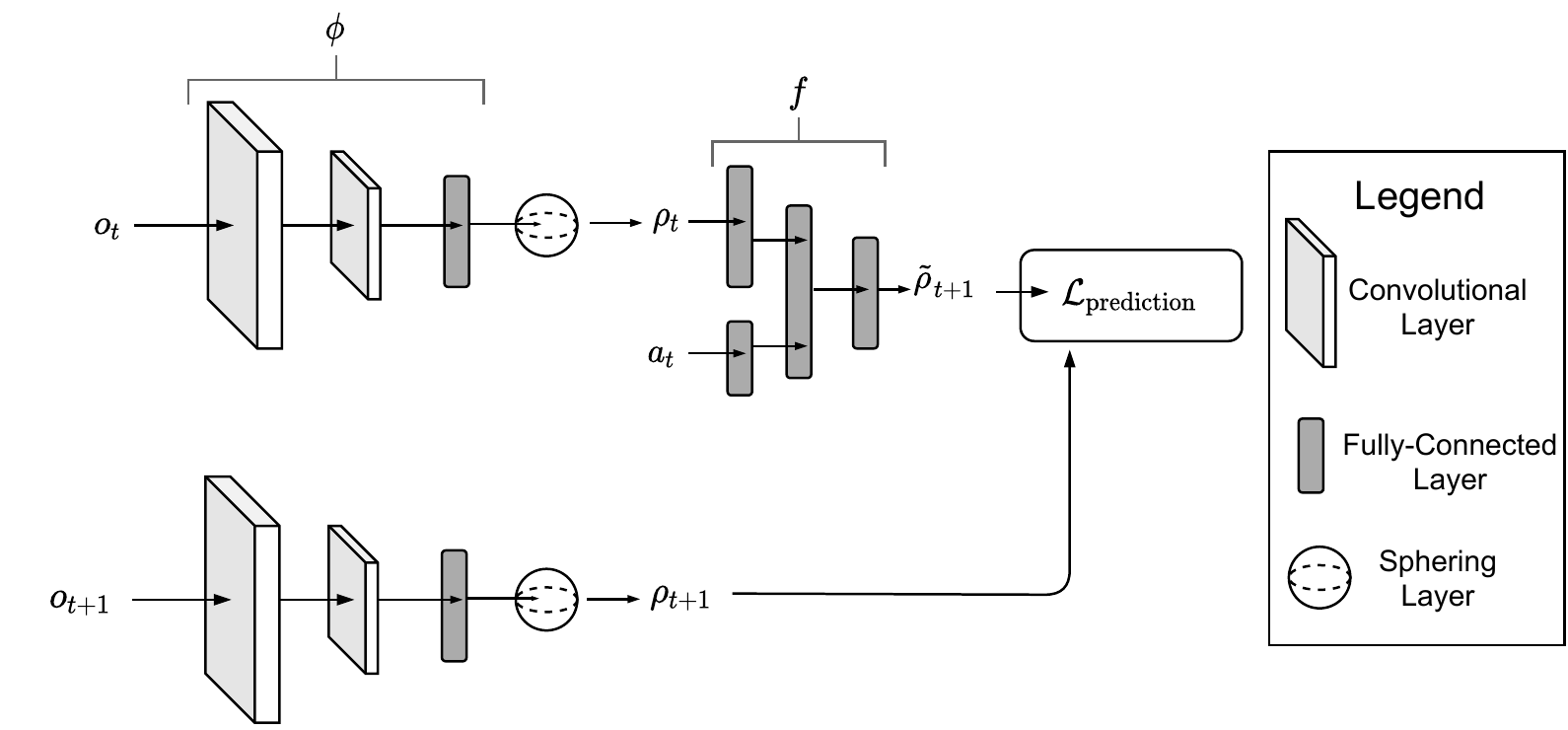}
    \caption[Illustration: Predictive representation learning with sphering regularization]{{\bf Predictive representation learning with sphering regularization.} The observations $o_t$, and the resulting observation $o_{t+1}$ after the action $a$ has been performed in $o_t$, are passed through the representation map $\phi$, whose outputs are passed to a differentiable sphering layer before being passed to $f$. The predictive network $f$ minimizes the loss function $\mathcal{L}$, which is the mean-squared error between $ \phi(o_t) = \rho_t$ and $ \phi(o_{t+1}) = \rho_{t+1}$.}
\label{spheringdiagram}
\end{figure}

The problem of trivial solutions is solved in SFA \citep{wiskott2002slow} and related methods \citep{escalante2013solve, schuler2018gradient} by constraining the overall covariance of the output to be $I$. Including this constraint to our setting yields the optimization formulation:
\begin{equation}
\begin{aligned}
& \underset{\eta, \theta}{\text{minimize}}
& & \mathcal{L}_{\text{prediction}}(\mathcal{D}, \theta, \eta) \\
& \text{subject to}
& & \mathbb{E}_{\mathcal{D}}\left[ \phi_\eta  \right] \hspace{7pt}= 0 \hspace{47pt}\hspace{0.1em}\text{(zero mean)} \\
&&&  \mathbb{E}_{\mathcal{D}}\left[ \phi_\eta\phi_\eta^T \right]= I\, \ \ \   \ \ \  \ \ \ \text{(unit covariance)} \\
\end{aligned}
\end{equation}
\noindent We enforce this constraint in our network via architecture design. The last layer performs differentiable sphering \citep{schuler2018gradient, hlynsson2019learning} of the second-to-last layer's output using the whitening matrix $\bm{W}$. We get $\bm{W}$ using power iteration of the following iterative formula:
\begin{equation}
\bm{u}^{[i+1]} = \frac{\bm{T}\bm{u}^{[i]}}{|| \bm{T}\bm{u}^{[i]} ||}
\end{equation}
\noindent 

\noindent where the superscript $i$ tracks the iteration number and $\bm{u}^{[0]}$ can be an arbitrary vector. The power iteration algorithm
converges to the largest eigenvector  $\bm{u}$ of a matrix $\bm{T}$ in a few hundred, quick iterations. The eigenvalue $\lambda$ is determined, and we subtract the eigenvector from the matrix:
\begin{equation}
\bm{T} \leftarrow \bm{T} - \lambda \bm{u} \bm{u}^T
\end{equation}
\noindent the process is repeated until the sphering matrix is found
\begin{equation}\bm{W} = \sum_{j=0} \frac{1}{\sqrt{\lambda_j}}\bm{u}_j \bm{u}_j^T
\end{equation}
\noindent The whole system, including the sphering layer, can be seen in Fig~\ref{spheringdiagram}, with an abstract convolutional neural network as the representation $\phi$ and a fully-connected neural network as the prediction function $f$.

\subsubsection{(ii) Contrastive loss} Constant solutions can also be dealt with in the loss function instead of via architecture design. \cite{hadsell2006dimensionality} propose to solve this with a loss function that pulls together the representation of similar objects (in our case, states that are reachable from each other with a single action) but pushes apart the representation of dissimilar ones:
\begin{equation}
L_{\text{contrastive}}{(o, o^{'} )} = \begin{cases} || \phi (o) - \phi(o^{'})  || & \text{if } o, o^{'} \text{ are similar} \\
                      \max(0, m - || \phi (o) - \phi(o^{'})  || )          & \text{otherwise} \end{cases}
\end{equation}
\noindent where $m$ is a margin and $|| \cdot||$ is some --- usually the Euclidean --- norm. The representation of dissimilar objects is pushed apart only if the inequality
\begin{equation}
|| \phi (o) - \phi(o^{'})  || < m
\end{equation}
is violated. During each training step, we compare each observation to a similar and a dissimilar observation simultaneously~\citep{schroff2015facenet} by passing a triplet of (positive, anchor, negative) observations during training to three copies of $\phi$. In our experiments, the positive corresponds to the predicted embedding of $o_{t+1}$ given $o_t$ and $a_t$,  the anchor, is the true resulting embedding after an action $a_t$ is performed in state  $o_t$ and $\phi(o_n)$, the negative, is the representation of an arbitrarily chosen observation that is not reachable from $\phi(o_t)$ with a single action. For environments where this is determinable, such as in our experiments, this can be assessed from the environment's full state. When this information isn't available, ensuring for $\phi(o_n)$ and $\phi(o_t)$ that $|n-t|>2$ is a good proxy, even though this can result in some incorrect triplets. For example, when the agent runs in a self-intersecting path.

We define the representation of the observation at time step $t$ as $\rho_t = \phi(o_t)$ and the next-step prediction $\Tilde{\rho}_{t+1} := f\left(\phi(o_t), a_t)\right)$ for readability and our (positive, anchor, negative) triplet is thus $\left(\Tilde{\rho}_{t+1}, \phi(o_{t+1}), \phi(o_n) \right)$ and we minimize the triplet loss:
\begin{equation}
\mathcal{L}_{\text{contrastive}}(o_t, o_{t+1}, o_n, a_t) = || \rho_{t+1} - \Tilde{\rho}_{t+1}  || + \max(0, m - || \rho_{t+1} - \rho_n  || )  
\label{ourcontrastive}
\end{equation}

\noindent It would seem that $\rho _{t+1}$ and $\Tilde{\rho}_{t+1}$ are interchangeable, since the second term is included only to prevent the representation from collapsing into a constant. However, if the loss function is
\begin{equation}
\mathcal{L}_{\text{contrastive}}(o_t, o_{t+1}, o_n, a_t) = || \rho_{t+1} - \Tilde{\rho}_{t+1}  || + \max(0, m - ||  \Tilde{\rho}_{t+1} - \rho_n  || )  
\label{badcontrastive}
\end{equation}
then the network is rewarded during training for making $f$ poor at predicting the next representation instead of simply pushing the representation of $o_{t}$ and $o_n$ away from each other.

\begin{figure}[htp]
\centering
\includegraphics[scale=.99]{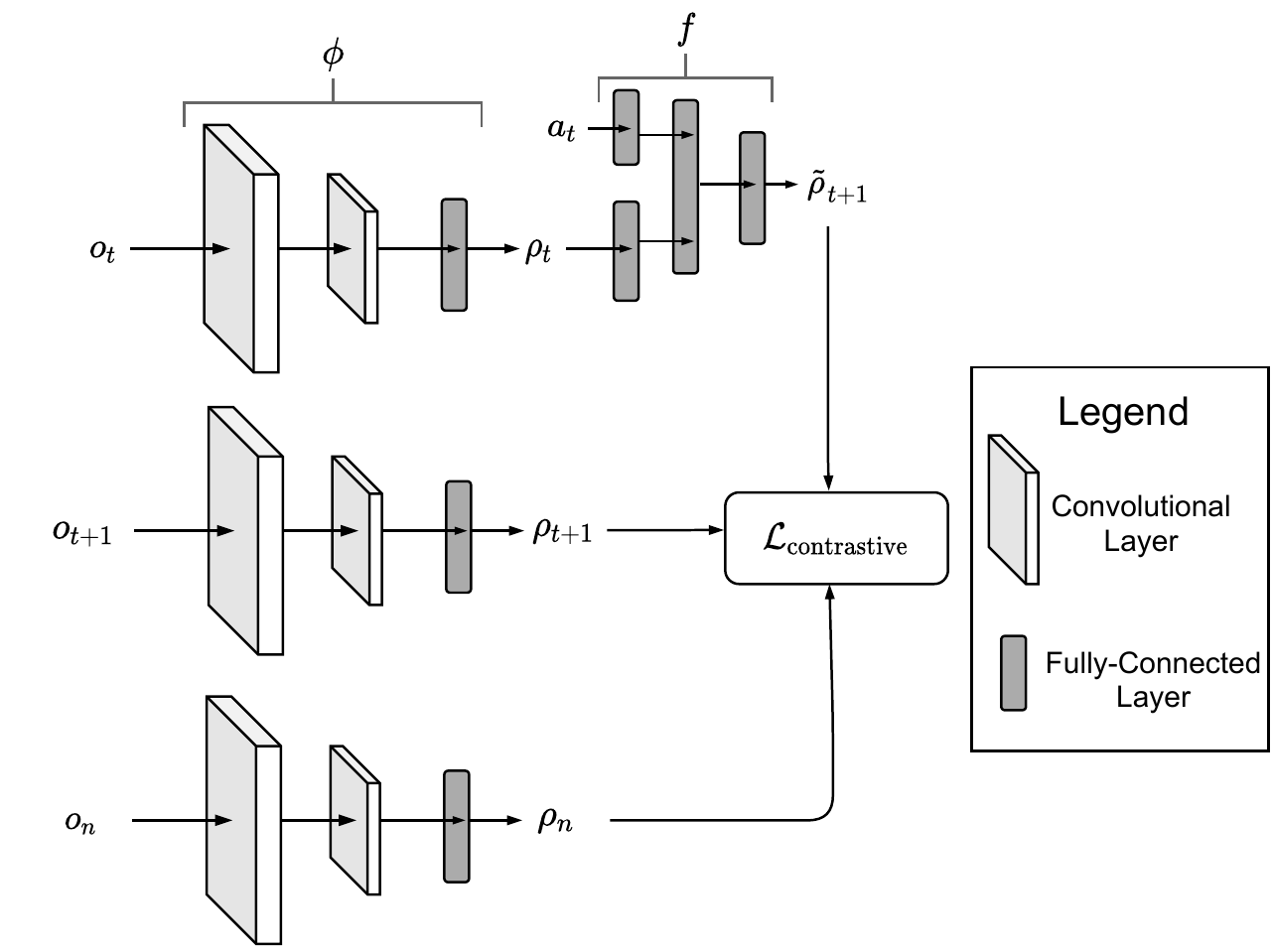} 
\caption[Illustration: Predictive representation learning with contrastive loss regularization]{ {\bf Predictive representation learning with contrastive loss regularization.} We minimize the contrastive loss function $\mathcal{L}_{\text{contrastive}}$ (Eq. \ref{ourcontrastive}). The predicted future representation $\Tilde{\rho}_{t+1}$ is pulled toward the next step's representation $\rho_{t+1}$. However, $\rho_{t+1}$ is pushed away from the negative state's representation $\rho_{n}$ if the distance between them is less than $m$. The observation $o_n$ is randomly selected from those that are not reachable from $o_{t}$ with a single action.}
\label{contrastivefigure}
\end{figure}

There are two main ways to set the margin $m$, one is dynamically determining it per batch \citep{sun2014deep}. The other, which we choose, is constraining the representation to be on a hypersphere using $L_2$ normalization and setting a small constant margin such as $m = 0.2$ \citep{schroff2015facenet}. The architecture for the training scheme using the contrastive loss regularization is depicted in Fig.~\ref{contrastivefigure}.

\subsubsection{(iii) Reconstructive loss} 
\begin{figure}[htp]
\centering
\includegraphics[scale=.99]{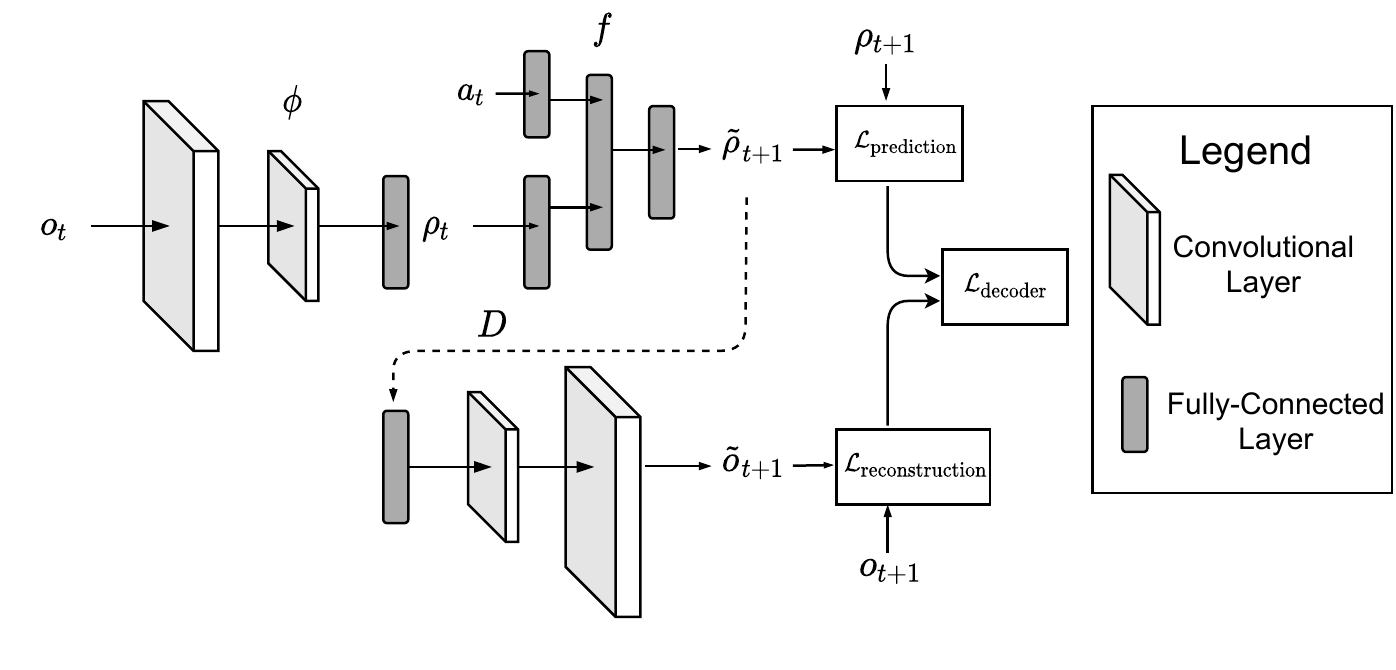} 
\caption[Illustration: Predictive representation learning with decoder loss regularization] {{\bf Predictive representation learning with decoder loss regularization.} At the time step $t$, the observation $o_t$ is passed to the representation $\phi$. This produces $\rho_t$ which is passed, along with the action $a_t$ at time step $t$, to the predictor network $f$. This produces the predicted $\Tilde{\rho}_{t+1}$ which is compared to $\rho_{t+1} = \phi(o_{t+1})$ in the mean-squared error term $\mathcal{L}_\text{prediction}$. The prediction $\Tilde{\rho}_{t}$ is also passed to the decoder network $D$. We then compare $\Tilde{o}_{t+1} = D(\Tilde{\rho}_{t+1})$ with $o_{t+1}$ in $\mathcal{L}_\text{decoder}$, another mean-squared loss term. The final loss is the sum of these two loss terms $\mathcal{L}_\text{total} = \mathcal{L}_\text{prediction}+\mathcal{L}_\text{decoder}$.}
\label{decoderlossgraphic}
\end{figure}

Trivial solutions are avoided by Goroshin et al. \citep{goroshin2015learning} by introducing a decoder network $D$ to a system that would otherwise converge to a constant representation. We incorporate this intuition into our framework with the loss function

\begin{align}
\label{decoderloss}
     \mathcal{L}_{\text{decoder}} (o_{t}, o_{t+1}, a_t) = \mathcal{L}_{\text{prediction}}(o_{t},o_{t+1}) + \mathcal{L}_{\text{reconstruction}}(o_{t},o_{t+1}) \nonumber   && \\
    =\left( \rho_t - \Tilde{\rho}_{t+1}\right)^2 + \alpha  \left(o_{t+1}  - D \left( \Tilde{\rho}_{t+1} \right) \right) ^2 \hspace{12pt} && \text{}
\end{align}

\noindent where $\alpha$ is a positive, real coefficient to control the regularization strength. Fig.~\ref{decoderlossgraphic} shows how the models and loss functions are related during the training of the representation and predictor using both a predictive and a reconstructive loss term.

The desired effect of the regularization can also be achieved by replacing the second term in Eq.~\eqref{decoderloss} with $\alpha  \left( o_{t+1}  - D \left( \rho_{t+1}\right) \right) ^2$. By doing this we would maximize the reconstructive property of the latent code in and of itself, which is not inherently useful for planning. We instead add an additional level of predictive power in $f$: in addition to predicting the next representation, its prediction must also be useful in conjunction with the decoder $D$ for reconstructing the new true observation.

This approach can have the largest computational overhead of the three, depending on the size of the decoder. We construct the decoder network $D$ such that it closely mirrors the architecture of $\phi$, with convolutions replaced by transposed convolutions and max-pooling replaced by upsampling.

\subsection{Training the predictor network}

We train the representation network and predictor network jointly by minimizing Eq.~(2), Eq.~(10) or Eq.~(12). The predictor network can also be trained on its own for a fixed representation map $\phi$.  In this case, $f$ is tasked as before with predicting $\phi(o_{t+1})$ after the action $a_t$ is performed in the state with observation $o_t$ by minimizing the mean-squared error between $f(\phi(o_{t}), a_t)$ and $\phi(o_{t+1})$. The networks are built with Keras \citep{chollet2015keras} and optimized with rmsprop \citep{tieleman2012lecture}. 

\subsection{Planning in transition-learned domain representation space}
We use a modified best-first search algorithm with the trained representations for our experiments (Algorithm \ref{algo: gs_algo}).

\begin{algorithm}[thb]
\caption{Perform a simulated rollout to find a state that is maximally similar to a goal state. Output a sequence to reach the found state from the start state.}\label{algo: gs_algo}
\begin{algorithmic}[1]
\REQUIRE $o_{\text{start}}$, $o_{\text{goal}}$, max trials $m$, action set $\mathcal{A}$,  representation map $\phi$ and predictor function $f$ 
\ENSURE A sequence of actions $(a_0, \dots, a_n)$ connecting the start state to the goal state
\STATE Initialize the set $Q$ of unchecked representations with the representation of the start state $\phi(o_{\text{start}})$
\STATE Initialize the dictionary $\mathbb{P}$ of representation-path pairs  with the initial representation mapped to an empty sequence: $\mathbb{P}[\phi(o_{\text{start}})]\leftarrow(\varnothing$)
\STATE Initialize the empty set of checked representations $C \leftarrow \varnothing$
\FOR{$k \leftarrow 0 \text{ to } m$}
    \STATE Choose $\rho'$ $\leftarrow$   $\underset{\rho \in Q}{\text{argmin}} ||\rho - \phi(o_{goal}) ||$
    \STATE Remove $\rho'$ from $Q$ and add it to $C$
    \FORALL{actions $a \in \mathcal{A}$} 
        \STATE Get a new estimated representation $\rho^* \leftarrow f(\rho', a)$ and add it to the set Q
        
        \STATE Concatenate $a$ to the end of $\mathbb{P}[p']$ and associate the resulting sequence with $\rho^*$ in the dictionary:  $\mathbb{P}[\rho^*] \leftarrow \mathbb{P}[\rho'] ^\frown  (a)$
        
        \STATE
    \ENDFOR
\ENDFOR
\STATE Find the most similar representation to the goal: $\rho_{\text{result}} \leftarrow  \underset{\rho \in Q \cup C}{\text{argmin}} ||\rho - \phi(o_{goal}) ||$
\RETURN{the sequence $\mathbb{P}[\rho_{\text{result}}]$}
\end{algorithmic}
\end{algorithm}

From a given state, the agent performs a simulated rollout to search for the goal state. For each action, the initial observation is passed to the predictor function along with the action. This results in a predicted next-step representation, which is added to a set. The actions taken so far and resulting in each prediction are noted also. The representation that is closest to the goal (using for example the Euclidean distance) is then taken for consideration and removed from the set. This process is repeated until the maximum number of trials is reached. The algorithm then outputs the sequence of actions resulting in the predicted representation that is the closest to the goal representation.

To make the algorithm faster, we only consider paths that do not take us to a state that has already been evaluated,
even if there might be a difference in the predictions from going this roundabout way. That is, if a permutation of the actions in the next path to be considered is already in an evaluated path, it will be skipped. This has the same effect as transposition tables used to speed up search in game trees. Paths might be produced with redundancies, which can be amended with path-simplifying routines (e.g. take one step forward instead of one step left, one forward then one right).

We do Model-Predictive Control \citep{garcia1989model}, that is, after a path is found, one action is performed and a new path is recalculated, starting from the new position. Since the planning is possibly over a long time horizon, we might have a case where a previous state is revisited. To avoid loops resulting from this, we keep track of visited state-action pairs and avoid an already chosen action for a given state.

\subsection{NORB viewpoint-matching experiments}

    For our experiments, we create an OpenAI Gym environment based on the small NORB data set \citep{lecun2004learning}. The code for the environment is available at https://github.com/wiskott-lab/gym-norb and requires the pickled NORB data set hosted at https://s3.amazonaws.com/unsupervised-exercises/norb.p. The data set contains 50 toys, each belonging to one of five categories: four-legged animals, human figures, airplanes, trucks, and cars. Each object has stereoscopic images under six lighting conditions, 9 elevations, and 18 azimuths (in-scene rotation). In all of the experiments, we train the methods on nine car class toys, testing on the other toys.

Each trial in the corresponding RL environment revolves around a single object under a given lighting condition. The agent is presented with a start and a goal viewpoint of the object and transitions between images until the current viewpoint matches the goal where each action operates the camera. To be concrete, the actions correspond to turning a turntable back and forth by $20^{\circ}$, moving the camera up or down by $5^{\circ}$ and, in one experiment, changing the lighting. The trial is a success if the agent manages to change viewpoints from the start position until the goal viewpoint is matched in fewer than twice the minimum number of actions necessary.

We compare the representations learned using the three variants of our method to five representations from the literature, namely  \textbf{(i)} Laplacian Eigenmaps  \citep{belkin2003laplacian}, \textbf{(ii)} the second-to-last
layer of VGG16 pretrained on ImageNet \citep{deng2009imagenet}, \textbf{(iii)} UMAP embeddings \citep{mcinnes2018umap}, \textbf{(iv)} convolutional encoder \citep{masci2011stacked} and \textbf{(v)} VAE codes \citep{kingma2013auto}. As fixed representations do not change throughout the training, they can be saved to disk, speeding up the training. As a reference, we consider three reinforcement learning methods working directly on the input images: \textbf{(i)} Deep Q-Networks (DQN)~\citep{mnih2013playing}, \textbf{(ii)} Proximal Policy Optimization (PPO)~\citep{schulman2017proximal} and \textbf{(iii)} World Models~\citep{ha2018world}.

The data set is turned into a graph for search by setting each image as a node and each viewpoint-changing action as an edge. The task of the agent is to transition between neighboring viewing angles until a goal viewpoint is reached. The total number of training samples is fixed at 25600. For our method, a sample is a single ($o_t$, $o_{t+1}$, $a_t$) triplet to be predicted while for the regular RL methods it is a ($o_t$, $o_{t+1}$, $a_t$, $\rho_t$) tuple.

\subsection{Model architectures}

\subsubsection{Input} The network $\phi$ encodes the full NORB input, a $96 \times 96$ pixel grayscale image, to lower-dimensional representations. The system as a whole receives the image from the current viewpoint, the image of the goal viewpoint and a one-hot encoding of the taken action. The image inputs are converted from integers ranging from 0 to 255 to floating point numbers ranging from 0 to 1.

\subsubsection{Representation learner $\phi$ architecture} We use the same architecture for the $\phi$ network in all of our experiments except for varying the output dimension,  Table~\ref{table:phiarchiteture}. 

\begin{table}[htb]\begin{center}
  \caption[Result: LARP representation network architecture]{{\bf Representation network architecture.}}
  \label{table:phiarchiteture}
 \begin{tabular}{l r r r l r} 
 Layer &   Filters / Units  & Kernel size & Strides & Output shape & Activation   \\ [0.5ex] 
 \hline \hline
     Input  &  &  &  & (96, 96, 1) &  \\ 
      \cline{1-6} 
     Convolutional  & 64 & $5 \times 5$ & $2 \times 2$ & (45, 45, 64) & ReLU \\ 
      \cline{1-6} 
     \cline{1-5} Max-pooling &  & & $2 \times 2$  & (22, 22, 64) &  \\ 
  \cline{1-6}  Convolutional & 128 & $5 \times 5$ & $2 \times 2$ & (9, 9, 128) & ReLU \\ 
     \cline{1-6} Flatten &  &  & & (10368) & \\ 
     \cline{1-6} Dense & 600 & && (600) & ReLU \\
     \cline{1-6} Dense & \#Features& && (\# Features) & Linear  \\
\end{tabular}
\end{center}\end{table} 

\subsubsection{Regularizing decoder architecture $D$} The decoder network $D$ has the architecture listed in Table~\ref{table:decoder}. It is designed to approximately inverse each operation in the original $\phi$ network.

\begin{table}[htb]
\begin{center}
  \caption[Result: LARP regularizing decoder architecture]{{\bf Regularizing decoder architecture.} The upsampling layer uses linear interpolation, BN stands for Batch Normalization and CT stands for Convolutional Transpose. }
  \label{table:decoder}
 \begin{tabular}{l r r r r l } 
 Layer &  Filters / Units & Kernel Size & Strides & Output shape & Activation  \\ [0.5ex] 
\hline \hline
     Input &   & &&(\# Features) & \\
    \cline{1-6}
     Dense & 512 & &&(512) & ReLU\\
     \cline{1-6}BN &  & & & (512) &   \\ 
    \cline{1-6} Dense & 12800 & &&(12800) &ReLU\\
    \cline{1-6} BN &  & & & (12800) &  \\ 
    \cline{1-6} Reshape &  & &   & (10, 10, 128) & \\ 
  \cline{1-6}  CT & 128 & $5 \times 5$ & $2 \times 2$ & (23, 23, 128) & ReLU \\ 
  \cline{1-6} Upsampling &  & &  $2 \times 2$  & (46, 46, 128) &\\ 
  \cline{1-6} BN &  & &&(46, 46, 128)    &\\ 
    \cline{1-6} CT  & 64 & $5 \times 5$ & $2 \times 2$ & (95, 95, 64) & ReLU\\ 
    \cline{1-6} BN &  & &  & (95, 95, 64) & \\ 
    \cline{1-6} CT & 1 & $2 \times 2$ & $1 \times 1$ & (96, 96, 1) & Sigmoid \\ 
\end{tabular}
\end{center}
\end{table} 

\subsubsection{Predictor network  $f$}  The predictor network $f$ is a two-stream dense neural network. Each stream consists of a dense layer with a rectified linear unit (ReLU) activation, followed by a batch normalization (BatchNorm) layer. The outputs of these streams are then concatenated and passed through 3 dense layers with ReLU activations, each one followed by a BatchNorm, and then an output dense layer, see Table \ref{table:predictornet}.

\begin{table}[htb]
\begin{center}
  \caption[Result: LARP representation predictor architecture]{{\bf Represention predictor architecture.} The $\phi$ stream receives the representation as input and the $A$ stream receives the one-hot action as input. Both streams are processed in parallel and then concatenated, with each operation applied from top to bottom sequentially. The number of hidden units in the last layer depends on the chosen dimensionality of the representation.}
  \label{table:predictornet}
 \begin{tabular}{l r r l} 
 Layer &  Filters / Units & Output shape & Activation  \\ [0.5ex] 
 \hline \hline
     \ \ \ \ $\phi$ Stream: Input  & & (\# Features) & ReLU\\
 \cline{1-4}
     \ \ \ \  $\phi$ Stream: Dense & 256 &  (256) & ReLU\\
     \cline{1-4}\ \ \ \  $\phi$ Stream: Batch Normalization &  &  (256) &    \\ 
    \cline{1-4}\ \  $A$ Stream: Input &  & (\# Actions) & ReLU\\

    \cline{1-4}\ \  $A$ Stream: Dense & 128  &(128)& ReLU\\
    \cline{1-4}\ \  $A$ Stream: Batch Normalization &   & (128) &    \\ 
    \cline{1-4} Concatenate $\phi$ and $A$ streams &   &  (384)  & \\ 

  \cline{1-4}  Dense & 256 &   (256)  & ReLU \\ 
      \cline{1-4} Batch Normalization &  & (256) &    \\

  \cline{1-4}  Dense & 256 &   (256)  & ReLU \\ 
      \cline{1-4} Batch Normalization &   & (256) &    \\
         \cline{1-4}  Dense & 128 &   (128)  & ReLU \\ 

    \cline{1-4} Batch Normalization &   &  (128) &  \\ 
    \cline{1-4}Dense & \# Features &   (\# Features)  & Linear \\ 
\end{tabular}
\end{center}
\end{table} 

\section{Results}
\label{larpresults}
With our empirical evaluation, we aim to answer the following research questions:
\begin{enumerate}
\item (Monotonicity) Is the Euclidean distance between a
suitable representation and the goal representation decreasing as the number of actions that separate them decreases?
\item (Trained predictability) Is training a representation for predictability, as proposed, feasible?
\item (Dimensionality) What is the best dimensionality of the latent space for our planning tasks?
\item (Solution constraints) In terms of planning performance, what are promising constraints to place on the representation to avoid trivial solutions?
\item (Benchmarking) How does planning with LARP compare to other methods from the RL literature?

\item (Generalization) How well does our method generalize to unseen environments?
\end{enumerate}
we will refer to these research questions by number below as they get addressed. 

\subsection{Latent space visualization}

When the representation and predictor networks are trained, we apply Algorithm \ref{algo: gs_algo} to the viewpoint-matching task. As described above, the goal is to find a sequence of actions that connects the start state to the goal state, where the two states differ in their configurations.

To support the qualitative analysis of the latent space, we plot heatmaps of similarity between the goal representation and the predicted representation of nodes during search (Fig~\ref{fig:lem_matching}). Of the 10 car toys in the NORB data set, we randomly chose 9 for our training set and test on the remaining one.

\subsubsection{In-sample embedding: Laplacian Eigenmaps}
\label{sec:in_sample}
First, we consider research question 1 (monotonicity). In order to get the best-case representation, we embed the toy using Laplacian Eigenmaps.  Embedding a single toy in three dimensions using Laplacian Eigenmaps results in a tube-like embedding that encodes both elevation and azimuth angles, see Fig \ref{fig:azimuth_cylinder}. Three dimensions are needed so that the cyclic azimuth can be embedded correctly as $\sin(\theta)$ and $\cos(\theta)$.

{
\captionsetup{aboveskip=-13pt}
\begin{figure}[htp]
\centering
\subcaptionbox*{}{\includegraphics[width=0.35\linewidth]{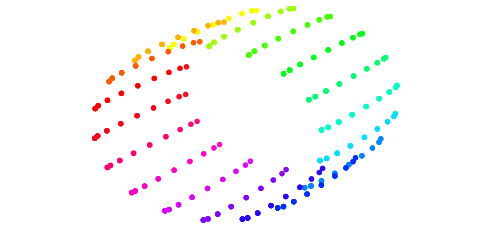}}
\hspace{-1.cm}
\subcaptionbox*{}{\includegraphics[width=0.35\linewidth]{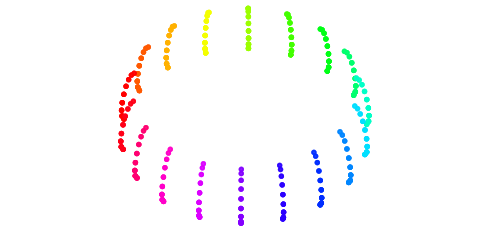}}
\hspace{-1.cm}
\vspace{-0.45cm}
\subcaptionbox*{}{\includegraphics[width=0.35\linewidth]{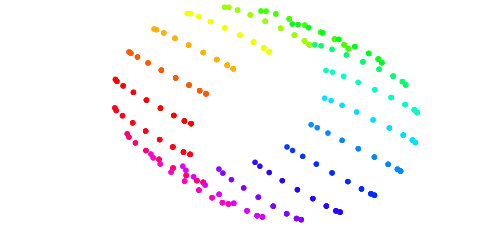}}
\vspace{5mm}
\subcaptionbox*{}{\includegraphics[width=0.35\linewidth]{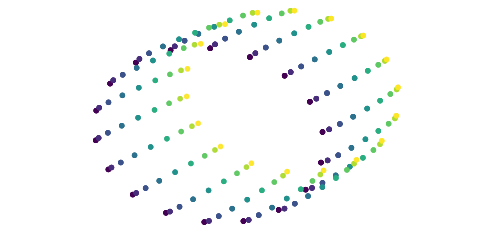}}
\hspace{-1.cm}
\subcaptionbox*{}{\includegraphics[width=0.35\linewidth]{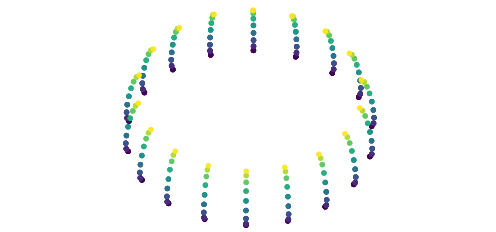}}
\hspace{-1.cm}
\subcaptionbox*{}{\includegraphics[width=0.35\linewidth]{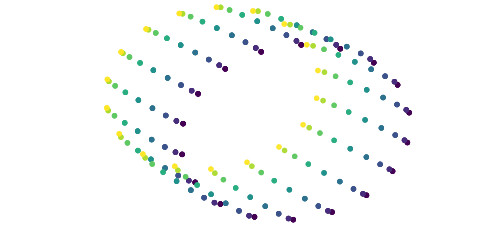}}
\vspace{5mm}
\caption[Result: Laplacian Eigenmap representation space of a NORB toy]{{\bf Laplacian Eigenmap representation space of a NORB toy.} The three-dimensional Laplacian Eigenmaps of a toy car where elements with the same azimuth values have the same color (top) and where elements with the same elevation values have the same color (bottom). Euclidean distance is a good proxy for geodesic distance in this case.}
\label{fig:azimuth_cylinder}
\end{figure}

}

\begin{figure}[htp]
\centering
 \captionsetup{width=1.0\linewidth}
  \resizebox*{1.   \textwidth}{!}{\includegraphics
{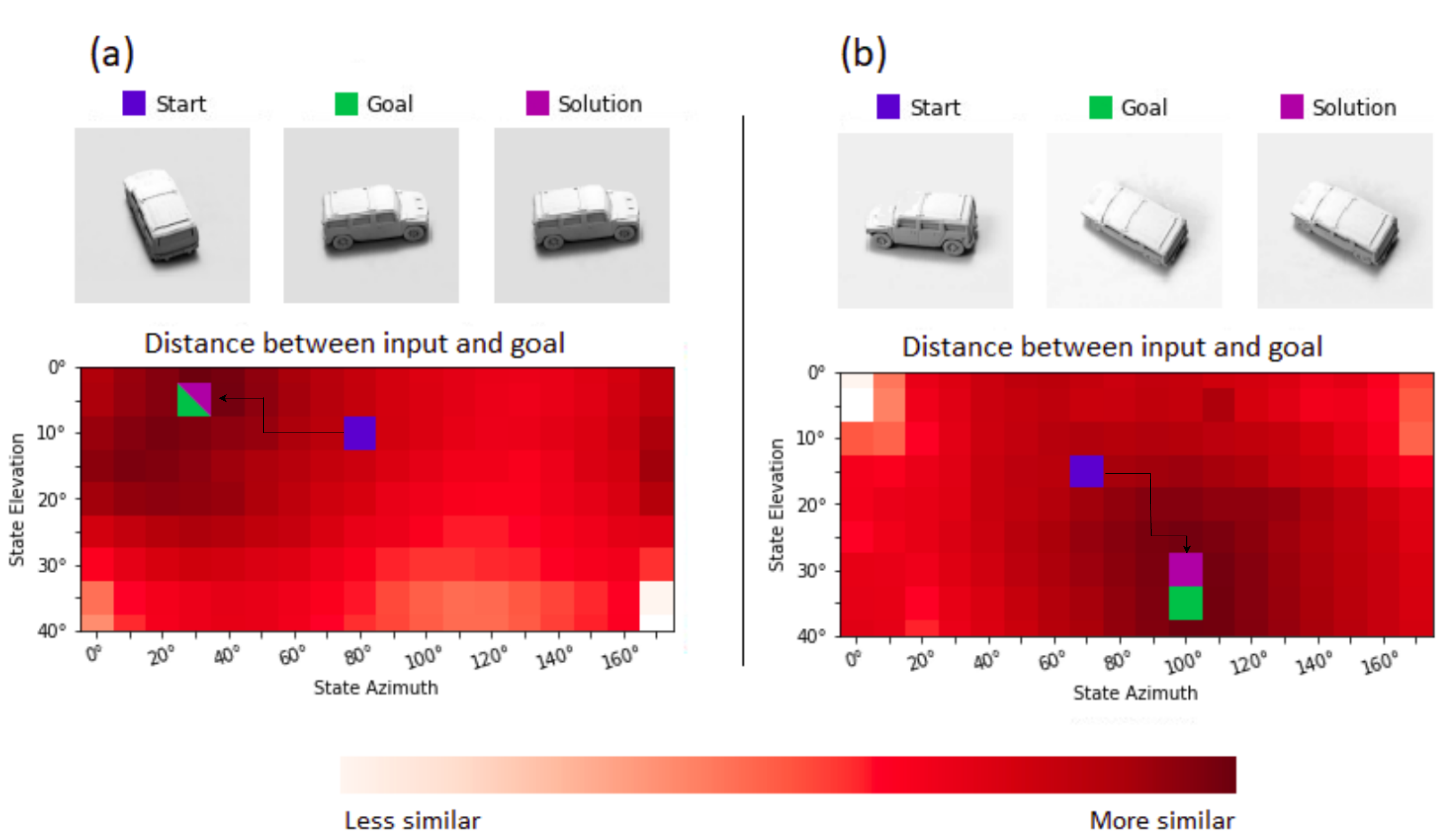}}
\caption[Result: Heatmap of Laplacian eigenmap latent space similarity]{ 
{\bf Heatmap of Laplacian Eigenmap latent space similarity.} Each pixel displays the difference between the predicted representation and the goal representation. Only the start and goal observations are given. The blue dot shows the start state, green the goal, and purple the solution state found by the algorithm.
The search algorithm can rely on an almost monotonically decreasing Euclidean distance between each state's predicted representation and the goal's representation to guide its search.
} 
\label{fig:lem_matching}
\end{figure}

If the representation is now used to train the predictor, one would expect that the representation becomes monotonically more similar to the goal representation as the state moves toward the goal. In Fig \ref{fig:lem_matching} we see that this is the case and that this behavior can be effectively used for a greedy heuristics. While the monotonicity is not always exact due to errors in the prediction, Fig \ref{fig:lem_matching} still qualitatively illustrates a best-case scenario.

We conclude from this that, for a suitable representation, the Euclidean distance between a current representation and the goal representation is monotonically increasing as a function of the number of actions that separate them. This supports the use of a prediction-based latent space search for planning.

\subsubsection{Out-of-sample embedding: pretrained VGG16 representation}
\begin{figure}[htp]
\centering
 \captionsetup{width=1.0\linewidth}
  \resizebox*{1.   \textwidth}{!}{\includegraphics
{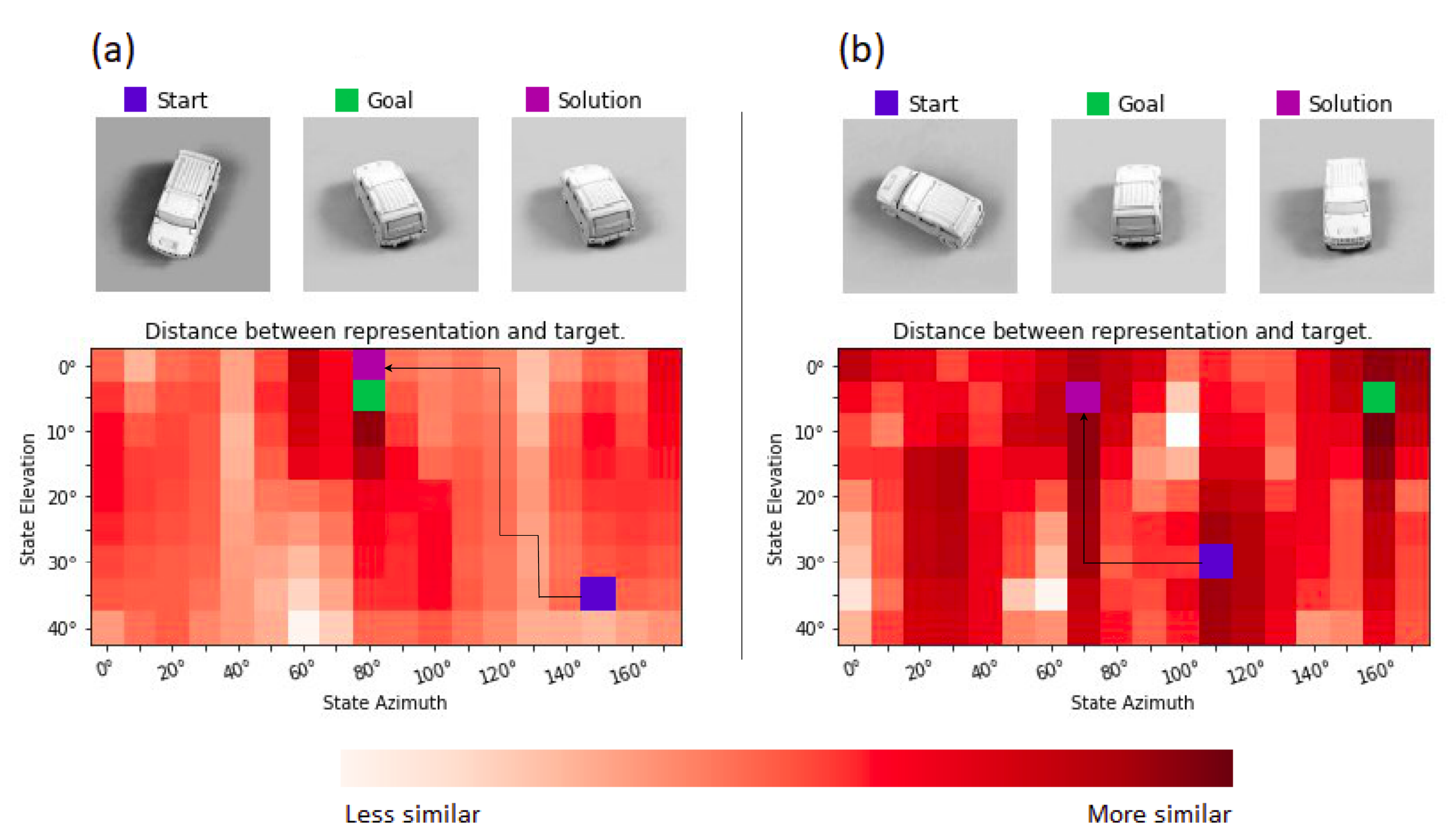}}
\caption[Result: Heatmap of VGG16 latent space similarity]{
{\bf Heatmap of VGG16 latent space similarity.}  The predictor network estimates the VGG16 representation of the resulting states as the object is manipulated.
 \textbf{(a)} The goal lies on a hill containing a maximum of representational similarity. \textbf{(b)} The accumulated errors of iterated estimations cause the algorithm to plan a path to a wrong state with a similar shape.
}
\label{fig:allinone}
\end{figure}

Next, we consider the pretrained representation of the VGG16 network to get a representation that generalizes to new objects. We train the predictor network and plot the heat map of the predicted similarity between each state and the goal state, beginning from the start state, in  Fig~\ref{fig:allinone}.

\begin{figure}[htp] 
\centering
 \captionsetup{width=.97\linewidth}
  \resizebox*{1.   \textwidth}{!}{\includegraphics
{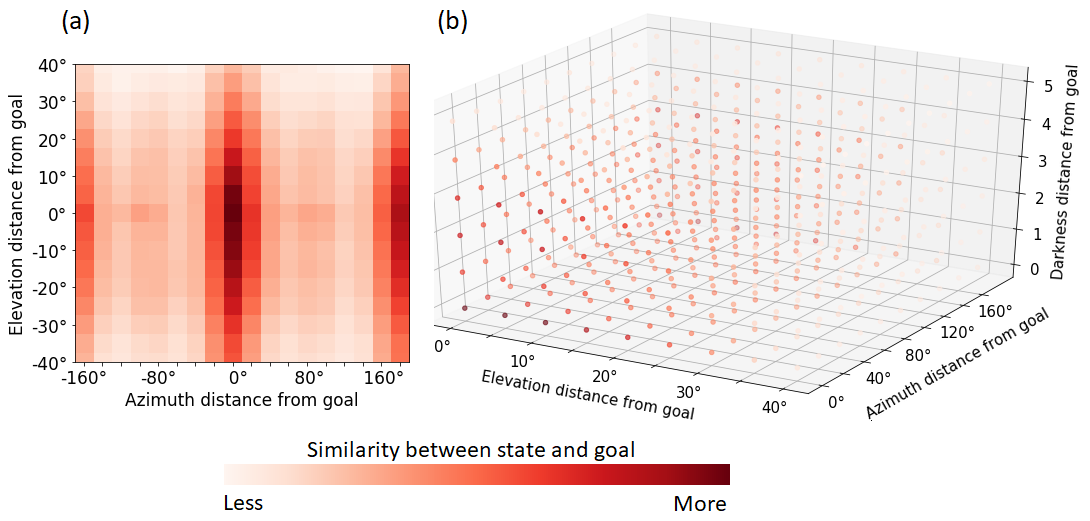}}
\caption[Result: Aggregate heat maps of VGG16 representation similarities on test data]{ {\bf Aggregate heat maps of VGG16 representation similarities on test data.} The data is collected as the state space is searched for a matching viewpoint. The pixels are arranged according to their elevation and azimuth difference from the goal state at $(0^{\circ}, 0^{\circ})$ on the left and $(0^{\circ}, 0^{\circ}, 0^{\circ})$ on the right.
\textbf{(a)} We see clear gradients toward the two basins of attraction. There is less change along the elevation due to less change at each step. \textbf{(b)} The agent can also change the lighting of the scene, with qualitatively similar results. In this graphic we only measure the absolute value of the distance.
}
\label{fig:all_seismic.png}
\end{figure}

The heat distribution in this case is more noisy. To get a view of the expected heat map profile, we average several figures of this type to show basins of attraction during the search. Each heat map is shifted such that the goal position is at the bottom, middle row (Fig~\ref{fig:all_seismic.png}, a).
Here, it is obvious that the goal and the $180^{\circ}$ flipped (azimuth)  version of the goal are attractor states. This is due to the representation map being sensitive to the rough shape of the object, but being unable to distinguish finer details. In (Fig~\ref{fig:all_seismic.png}, b) we display an aggregate heat map when the agent can also change the lighting conditions. 

Our visualizations show a gradient toward the goal state in addition to visually similar far-away-states, sometimes causing the algorithm to produce solutions that are the polar opposite of the goal concerning the azimuth. Prediction errors also prevent the planning algorithm from finding the exact goal for every task, even if it is not distracted by the polar-opposite.

\begin{figure}[htp]
\centering
 \includegraphics[width=1.\textwidth]{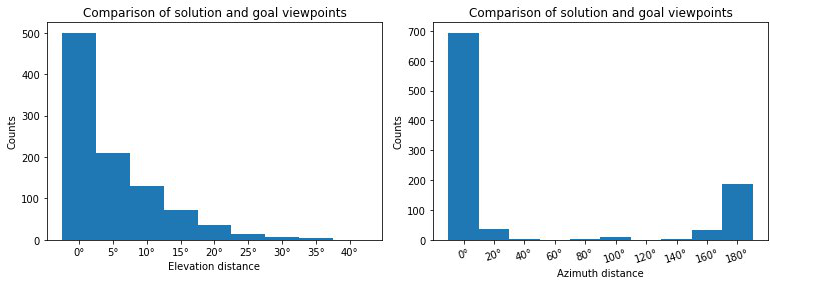}
\caption[Result: Histograms of elevation-wise and azimuth-wise VGG16 errors]{ {\bf Histograms of elevation-wise and azimuth-wise VGG16 errors.} The histograms display the counts of the distance between goal and solution states along elevation (left) and azimuth (right) on test data. The distance between the start and goal viewpoints is equally distributed across all the trials, along both dimensions. The goal and the $180^{\circ}$ flipped (azimuth) version of the goal are attractor states.
}
\label{fig:allinone_histos.png} 
\end{figure}

To investigate the accuracy of the search with respect to each dimension separately, we plot the histogram of distances between the goal states and the solution states in Fig~\ref{fig:allinone_histos.png}. The goal and start states are chosen randomly, with the restriction that the azimuth distance and elevation distance between them are each uniformly sampled.  For the rest of the chapter, all trials follow this sampling procedure.  The results look less accurate for elevation than azimuth because the elevation changes are smaller than the azimuth changes in the NORB data set.  The difference between the goal and solution viewpoints in Fig~\ref{fig:allinone} left, for example, is hardly visible. If one would scale the histograms by angle and not by bins, the drop-off would be similar.

\subsection{Latent space dimensionality}
With the next experiment, we aim to answer research question 2 (trained predictability). While tuning the details of the design, we also tackle research questions 3 (dimensionality) and 4 (solution constraints). 
We do an ablation study of the dimensionality of the representation for our method  (Table~\ref{featuresablation}). The test car is an unseen car toy from the NORB data set, and the train car comes from the training set. 

\begin{table}[h]
\begin{center}
  \captionof{table}[Result: Ablation study of the representation dimensionality]{{\bf Ablation study of the representation dimensionality.} We change the output dimension of the representation learner subnetwork and compare it to the VGG16 representation trained on ImageNet. The performance (mean success rate) is averaged over ten separate instantiations of our systems, where each instance is evaluated on a hundred trials of the viewpoint-matching task. A trial is a success if the goal is reached by taking less than twice the minimum number of actions needed to reach it. The standard deviations range between 0.1 and 0.3 for each table entry.}
  \label{featuresablation}
 \begin{tabular}{l r r r r} 
 Representation & Dimensions & Training Car ($\%$) & Test Car ($\%$)  \\ [0.2ex] 
\hline\\[-2.5ex]LARP (Contrastive)  & 96  &59.3& 56.8 \\ 
     & 64 &  64.1& 60.5\\ 
    & 32&72.3& 59.4  \\ 
     & 16 &74.1& 59.3  \\
  & 8&  82.7& 58.0   \\ [0.1ex]
  \hline\\[-2.5ex]LARP (Sphering) & 96 & 41.1& 37.8  \\ 
 &  64& \textbf{93.9}&53.7  \\ 
  & 32&  89.8 & 51.9  \\ 
 &16&85.2&42.6  \\
       &  8 &85.1    & 40.1 \\ [0.1ex] 
   \hline\\[-2.5ex]LARP (Decoder)     &  96& 58.0    &51.9  \\ 
   &   64& 79.5 & \textbf{63.0}  \\ 
  &  32   & 77.8 & 61.7\\ 
  &  16   &51.9 & 45.2 \\ 
  &  8  &51.1   & 42.4  \\ [.1ex] 
  \hline\\[-2.5ex]VGG16 &  902  & 62.4 & 55.1 \\ [.1ex]
 \hline\\[-2.5ex]Random Steps &  & 3.5  & 3.5 \\ [1ex] 
\end{tabular}
\end{center}\end{table}

There is no clear winner: the network with the sphering layer does the best on one of the cars used during training, while the reconstructive-loss network does the best on the held-out test car. The sharp difference in performance between 64 and 94 sphering-regularized representation can be explained by the numerical instability of the power iteration method for too large matrix dimensions. 

The VGG16 representation is not the highest performer on any of the car toys. Many of VGG16's representation values are 0 for all images in the NORB data set, so we only use those that are nonzero for any of the images. We suggest that this high number of dead units is due to the representation being too general for the task of manipulating relatively homogenous objects. Another drawback of using pretrained networks is that information might be encoded that is unimportant for the task. This has the effect that our search method is not guaranteed to output the correct solution in the latent space, as there might be distracting pockets of local minima. 

The random baseline has an average success rate of 3.5$\%$, which is very clearly outperformed by our method.  As 64 is the best dimensionality for the representation on average, we continue with that number for our method in the transfer learning experiment.

We conclude that the proposed method of training a representation for predictability is feasible. So far we have evidence that 64 is the best dimensionality of the representation's latent space for our planning tasks. However, it is not yet conclusive what the best restriction is to place on the representation to avoid the trivial solution, in terms of planning performance. 

\subsection{Comparison with other RL methods}
Now we divert our attention to research question 5 (benchmarking), where we compare our method to the literature. 
 For the comparison with standard RL methods, we use the default configurations of the model-free methods DQN and PPO as defined in OpenAI Baselines~\citep{baselines}. Our model-based comparison is chosen to be world models \citep{ha2018world} from https://github.com/zacwellmer/WorldModels. We make sure that the compared RL methods are similar to our system in terms of the number of parameters as well as architecture layout and compare them with our method on the car viewpoint-matching task.

 The task setup is the same as before and is converted to an OpenAI gym environment: a start observation and a goal observation are passed to the agent. If the agent manages to reach it within 2 times the minimum number of actions required (the minimum number is calculated by the environment), the agent receives a reward and the task is considered a success. Otherwise, no reward is given.

\begin{figure}[htp]
    \centering
    \subfloat[{\bf DQN performance.}]{{\includegraphics[width=.343\linewidth, angle=90]{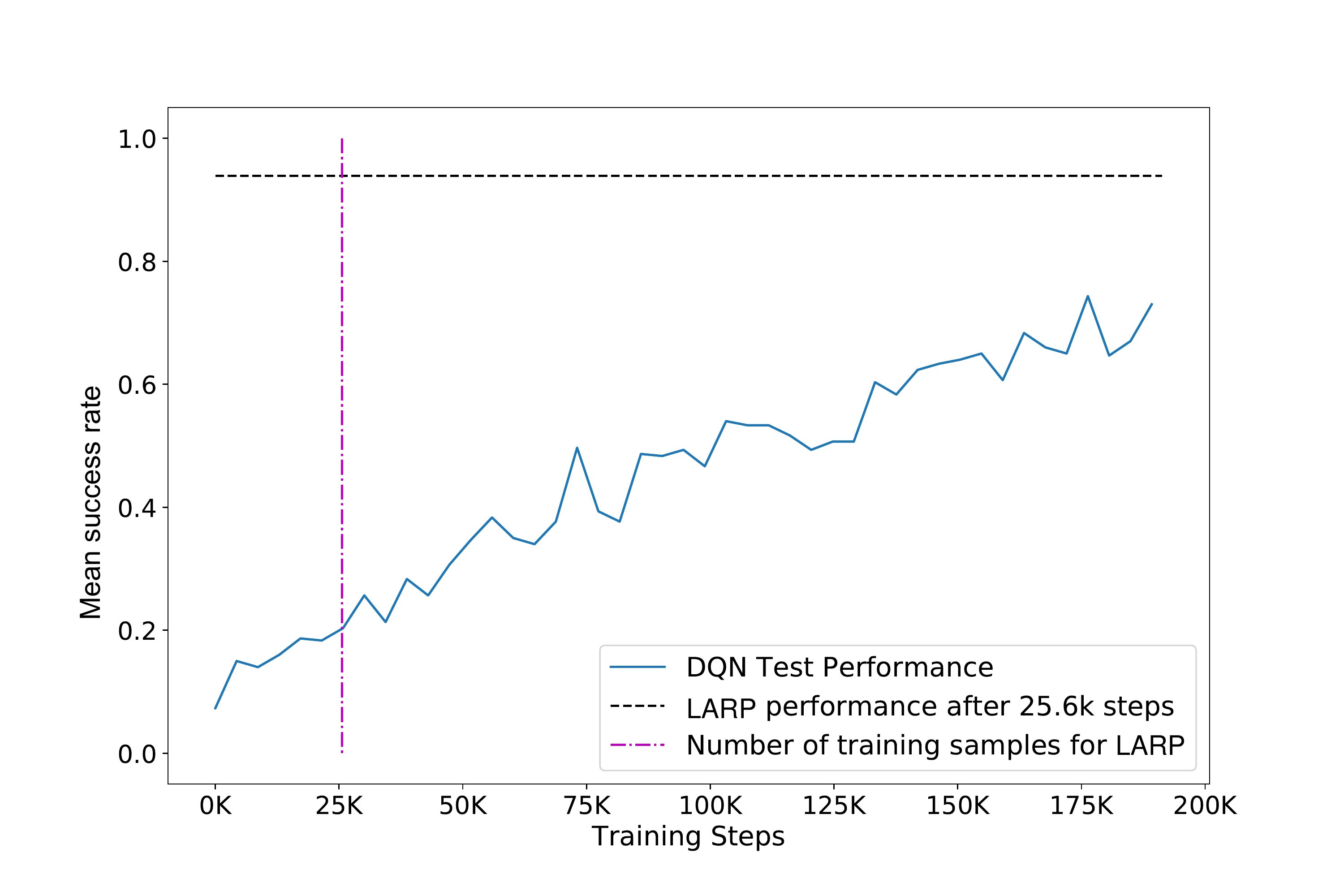}}}
    \qquad
      \hspace{-4em}
    \subfloat[{\bf PPO performance.}]{{\includegraphics[width=.343\linewidth, angle=90]{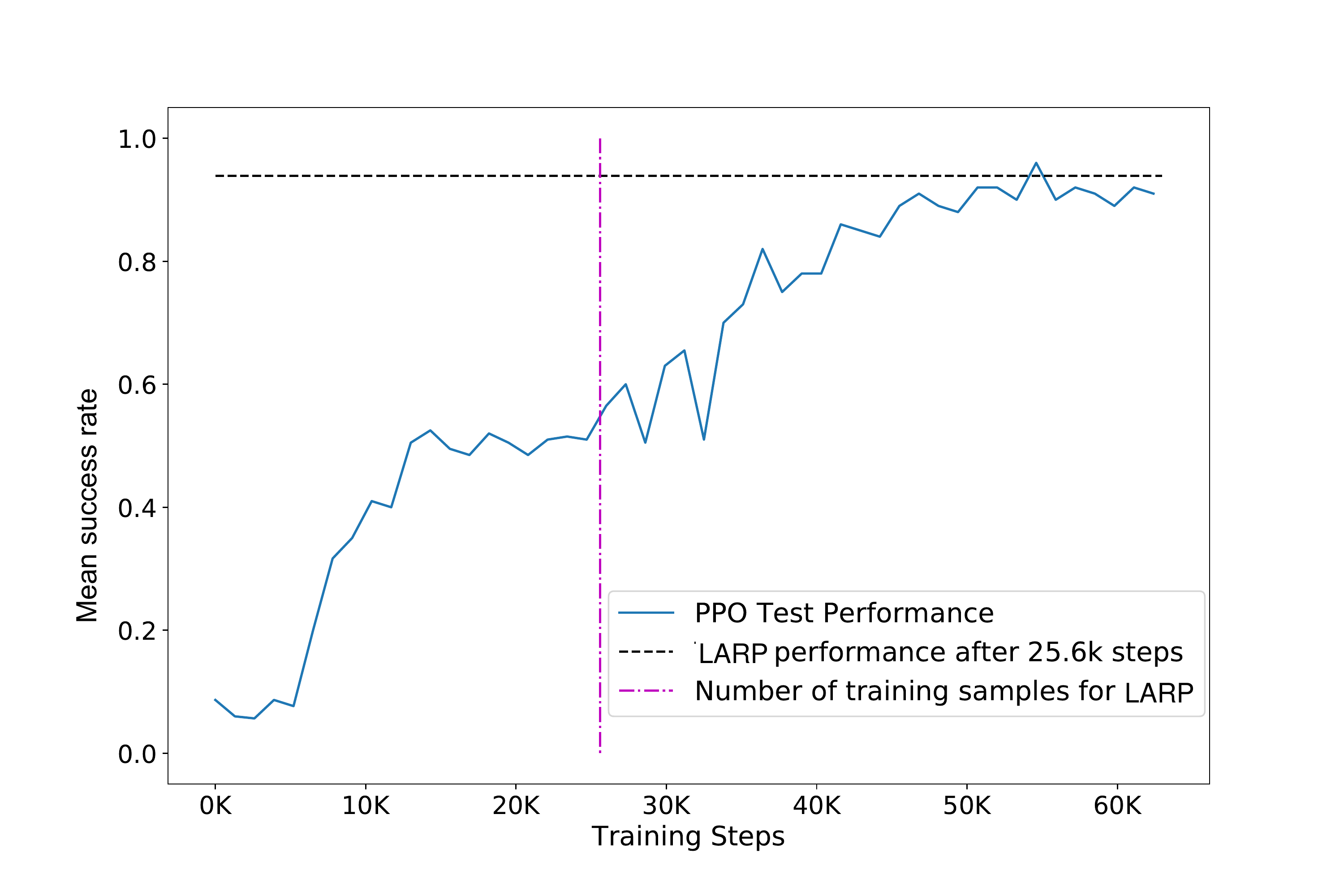}}}
    \qquad
    \subfloat[{\bf World models performance.}]{{\includegraphics[width=.343\linewidth, angle=90]{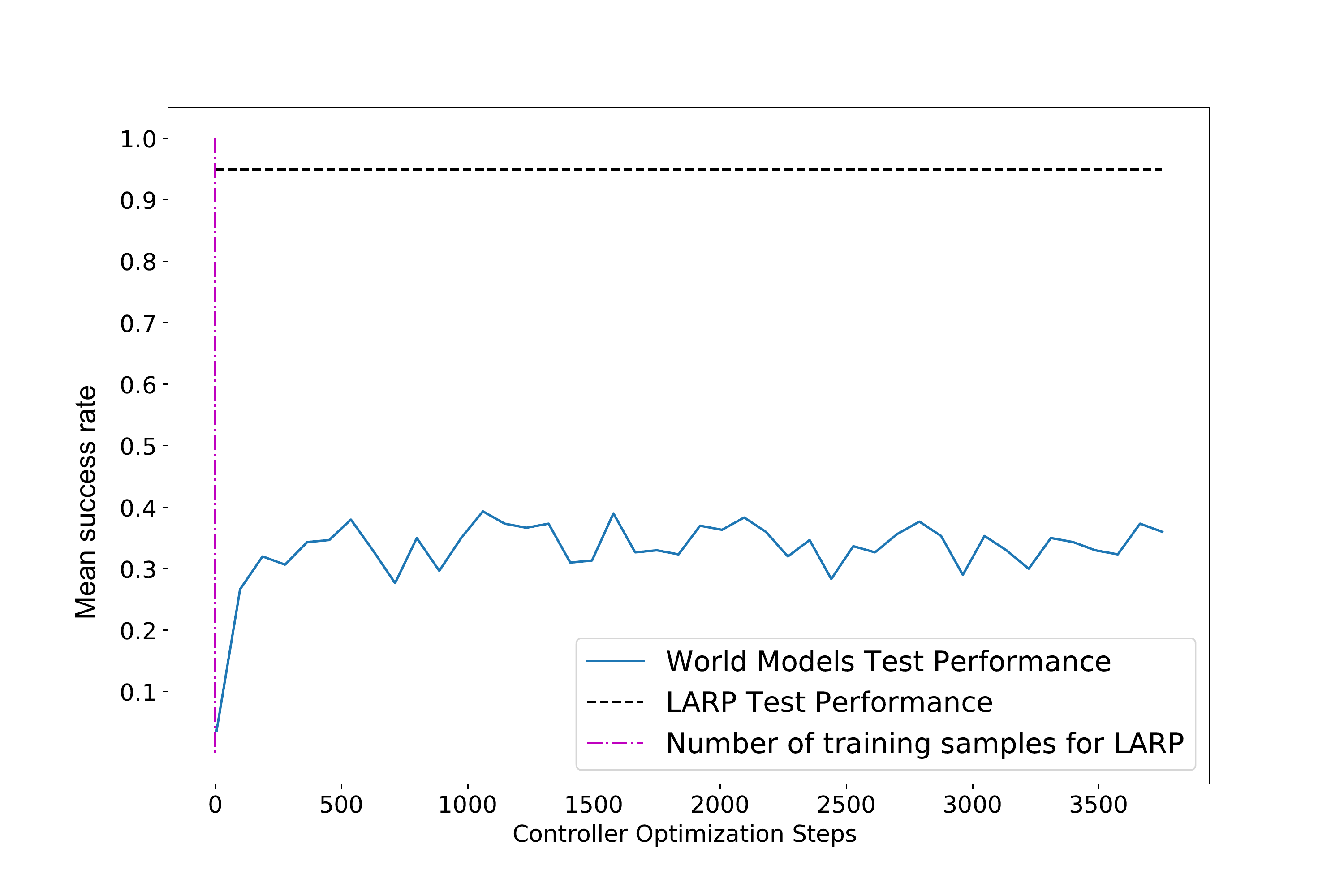}}}
    \qquad
      \hspace{-3em}
    \qquad
      \hspace{-5em}
    \caption[Result: LARP reinforcement learning comparison]{{\bf Reinforcement learning comparison.} The vertical dashed lines indicate when the compared algorithm has processed the same number of transitions as our method and the horizontal dotted line indicates the test performance of our method. Each data point is the   mean success rate of 100 test episodes after a varying number of training steps, averaged over 5 different seeds of each learner. The model-free methods in {\bf(a)} and {\bf(b)} train the representation and the controller simultaneously by acting in the environment and collecting new experiences. The representation in {\bf(c)} is trained on 25.6k transitions, which is the same number we use. The plot shows the optimization curve for the controller, using a Covariance-Matrix Adaptation Evolution Strategy, which hardly improves after 500 or so training steps. The horizontal line starts at 0 for world models because the representation has finished training on the observations before the controller is optimized. }
    \label{fig:rlfigs}
\end{figure}

 The results of the comparison can be viewed in Fig.~\ref{fig:rlfigs}. Each point in the curve contains each method's mean success rate: the average of the cumulative reward from 100 test episodes from 5 different instantiations of the RL learner, so it is the average reward over 500 episodes in total. The test episodes are done on the same environment as is used for training, except that the policy is maximally exploiting and minimally exploring.

 In our experiments, the DQN networks are much more sample inefficient than PPO, which in turn is more sample inefficient than our method. However, our method is more time-consuming during test time. We require a forward pass of the predictor network for each node that is searched before we take the next step, which can grow rapidly if the target is far away. In contrast, only a single pass through the traditional RL networks is required to compute the next action.

Our method reaches $93.9\%$ success rate on the train car (Table~\ref{featuresablation}) using 25.6k samples, but the best PPO run only reaches $70.5\%$ after training on the same number of samples. The best single PPO run needed 41.3k samples to get higher than $93.9\%$ success rate, and the average performance is higher than $93.9\%$ at around 55k samples. After that, some PPO learners declined again in performance. The world models policy quickly reaches the same level of performance as DQN got after 50k steps and PPO after approximately 8k steps, but it doesn't improve beyond that.

\begin{figure}[htp]
    \centering
    \subfloat[ {\bf}  ]
    {{\includegraphics[width=5.05cm]{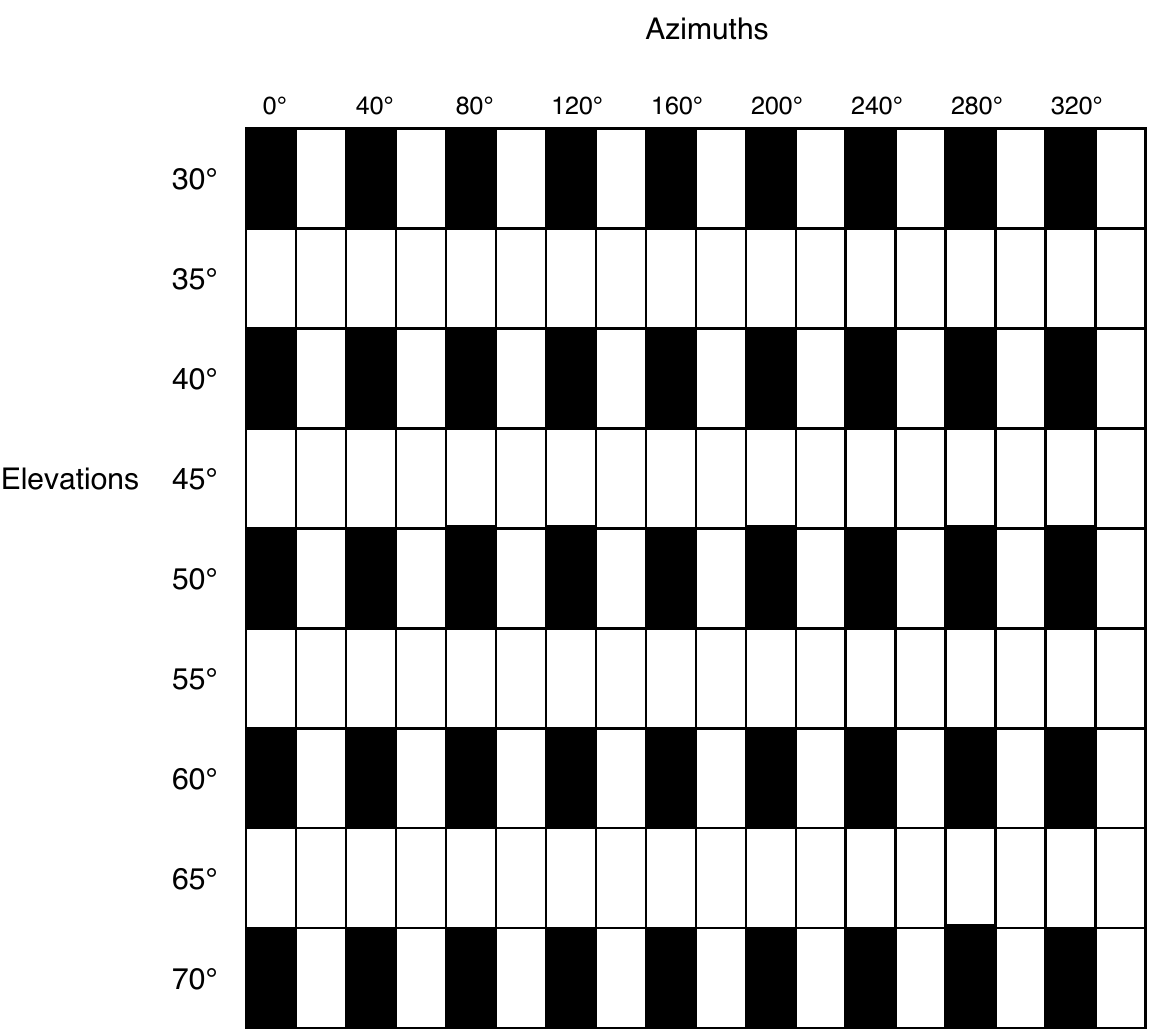}}}
    \qquad
      \hspace{-1.5em}
    \subfloat[{\bf}   ]
    {{\includegraphics[width=9.53cm]{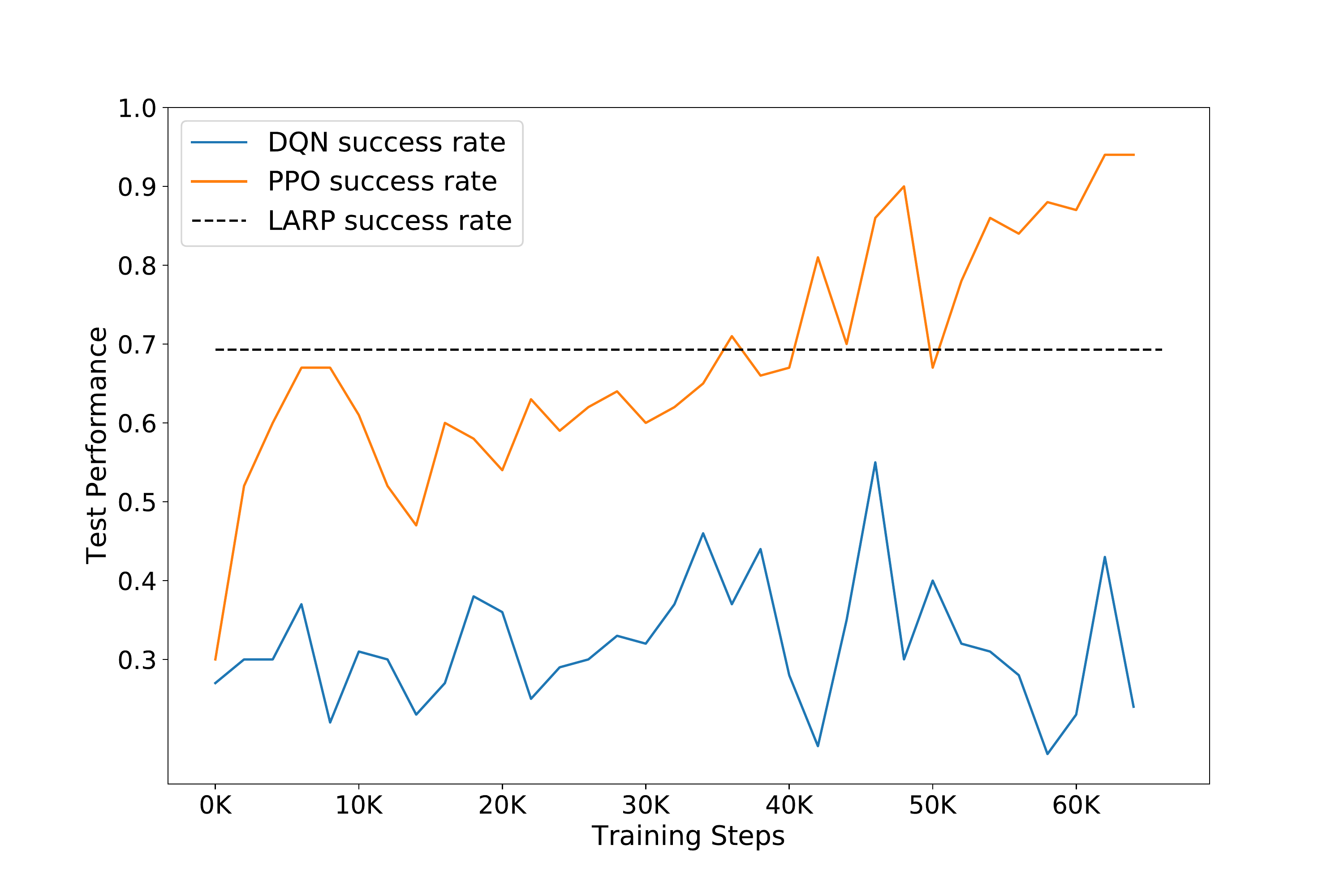}    \vspace{-1.0em}}}
    \caption[Result: LARP re-training after placing obstacles in a checkerboard pattern]{{\bf Re-training after placing obstacles in a checkerboard pattern.} {\bf (a)} The task is the same as before, but nothing happens if the agent attempts to move to a state containing a black rectangle. {\bf (b)}  After training the agents, we re-tested them after we  introduced the checkerboard pattern of obstacles. Our method does not allow for re-training in the new environment.}
    \label{fig:checkz}
\end{figure}

We conclude that our method compares favorably to other methods from the RL literature in terms of data-efficiency.

\subsection{Modifying the environment}
We now modify the environment to answer research question 6 (generalization).  To see how the methods compare when obstacles are introduced to the environment, we repeat the trial on one of the car objects except that the agent can no longer pass through states whose elevation values are divisible by 10 and azimuth values are divisible by 40 (Fig.~\ref{fig:checkz}, (a)).

As before, the goal and start locations can have any azimuth-elevation pair, but the agent cannot move into states with the properties indicated by the black rectangle. Every action is available to the agent at all locations as before, but the agent's state is unchanged if it attempts to move to a state with a black rectangle. 

We trained LARP using the contrastive loss, PPO, and DQN agents until they reached $80\%$ accuracy on our planning task and then tested them with the added obstacles. 
Our method loses about $10\%$ performance, but PPO loses $50\%$. Nevertheless, we can continue training PPO until it quickly reaches top performance again (Fig.~\ref{fig:checkz}, (b)). Our method is not re-trained for the new task, and DQN did not reach a good performance again in the time we allotted for re-training.  

Thus, we see that our method is quite flexible and generalizes well when obstacles are introduced to the environment.

\subsection{Transfer to dissimilar objects}

We now consider research question 6 (generalization) further by investigating how well our method transfers knowledge from one domain to another.

\begin{figure}[htp]
\centering
 \captionsetup{width=1\linewidth}
  \resizebox*{1.   \textwidth}{!}{\includegraphics
{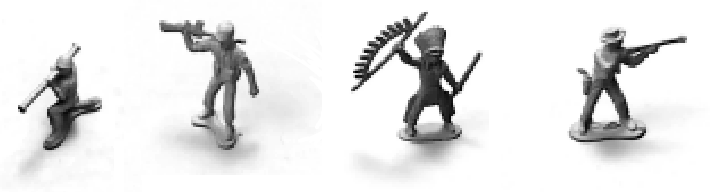}}
\caption[Example: Toys for transfer learning experiments]{ {\bf Toys for transfer learning experiments.} From left to right: Soldier (Kneeling), Soldier (Standing), Native American with Bow and Cowboy with Rifle. 
}
\label{fig:merge4.png}
\end{figure}

Selecting the best dimensionality for the representation from the previous set of experiments, we investigate further their performance in harder situations using unseen, non-car objects. The models are trained on the same car objects as in the previous experiment, but they are tested on an array of different plastic soldiers: a kneeling soldier holding a bazooka, a standing soldier with a rifle, a Native American with a bow and spear and a cowboy with a rifle (Fig.~\ref{fig:merge4.png}).

\subsubsection{Qualitative results}

\begin{figure}[htp]
\captionsetup[subfigure]{justification=centering,singlelinecheck=false}
\begin{subfigure}{0.3333\textwidth}
\includegraphics[width=\linewidth]{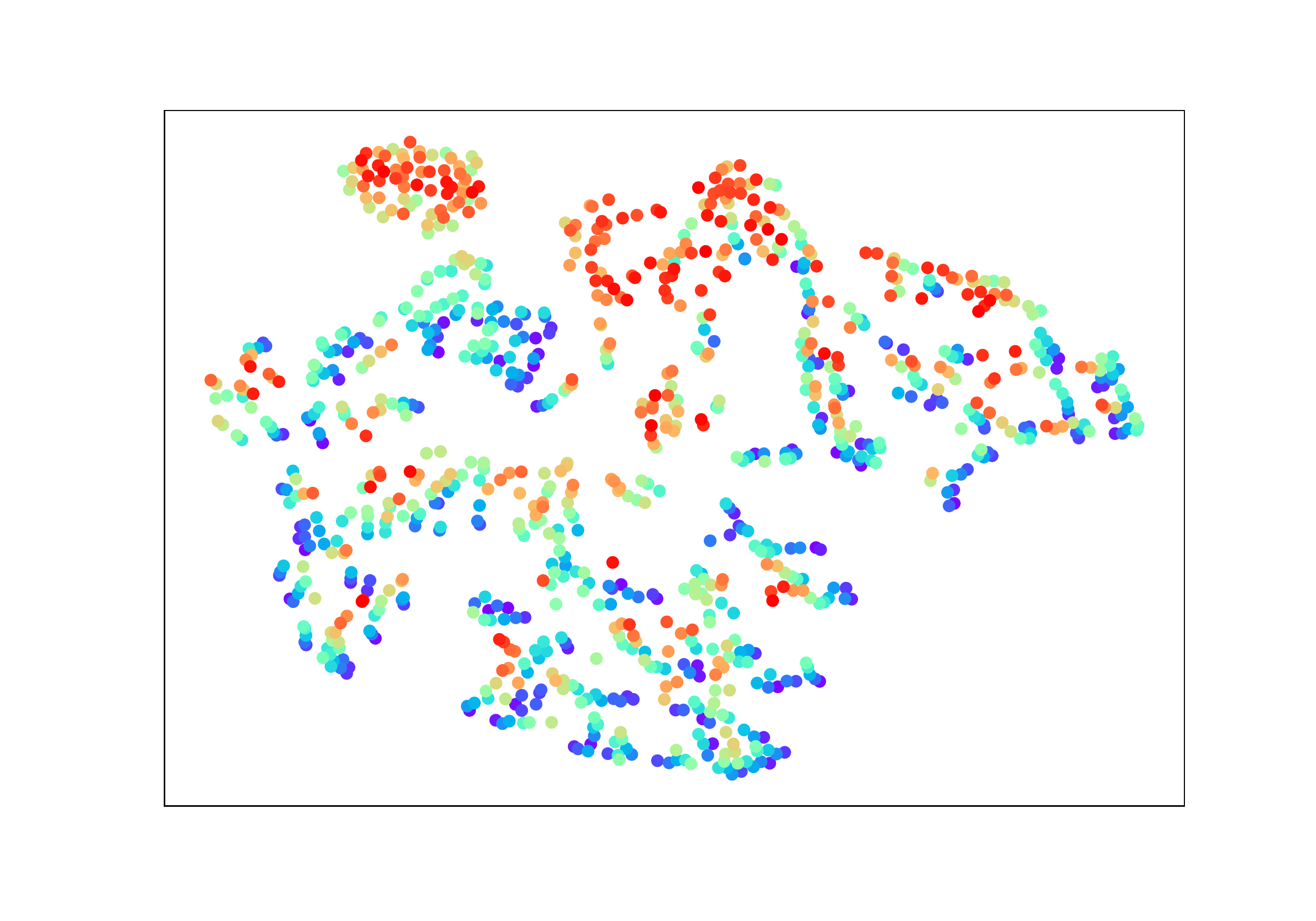} 
\caption{LARP elevation t-SNE} \label{fig:1picsa}
\end{subfigure} 
\hspace{-2em}
\begin{subfigure}{0.3333\textwidth}
\includegraphics[width=\linewidth]{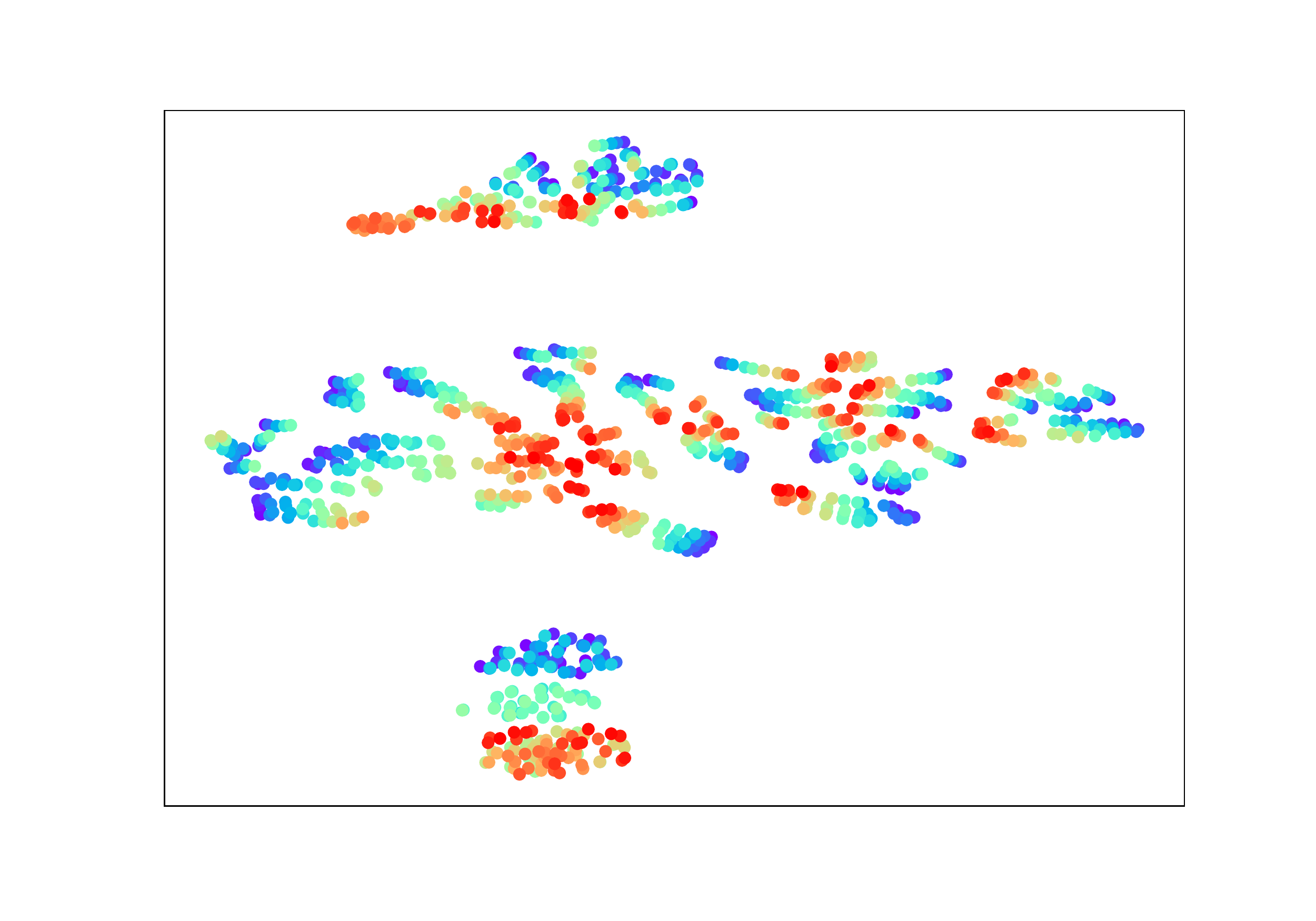}
\caption{CAE elevation t-SNE} \label{fig:2picsb}
\end{subfigure}
\hspace{-2em}
\begin{subfigure}{0.3333\textwidth}
\includegraphics[width=\linewidth]{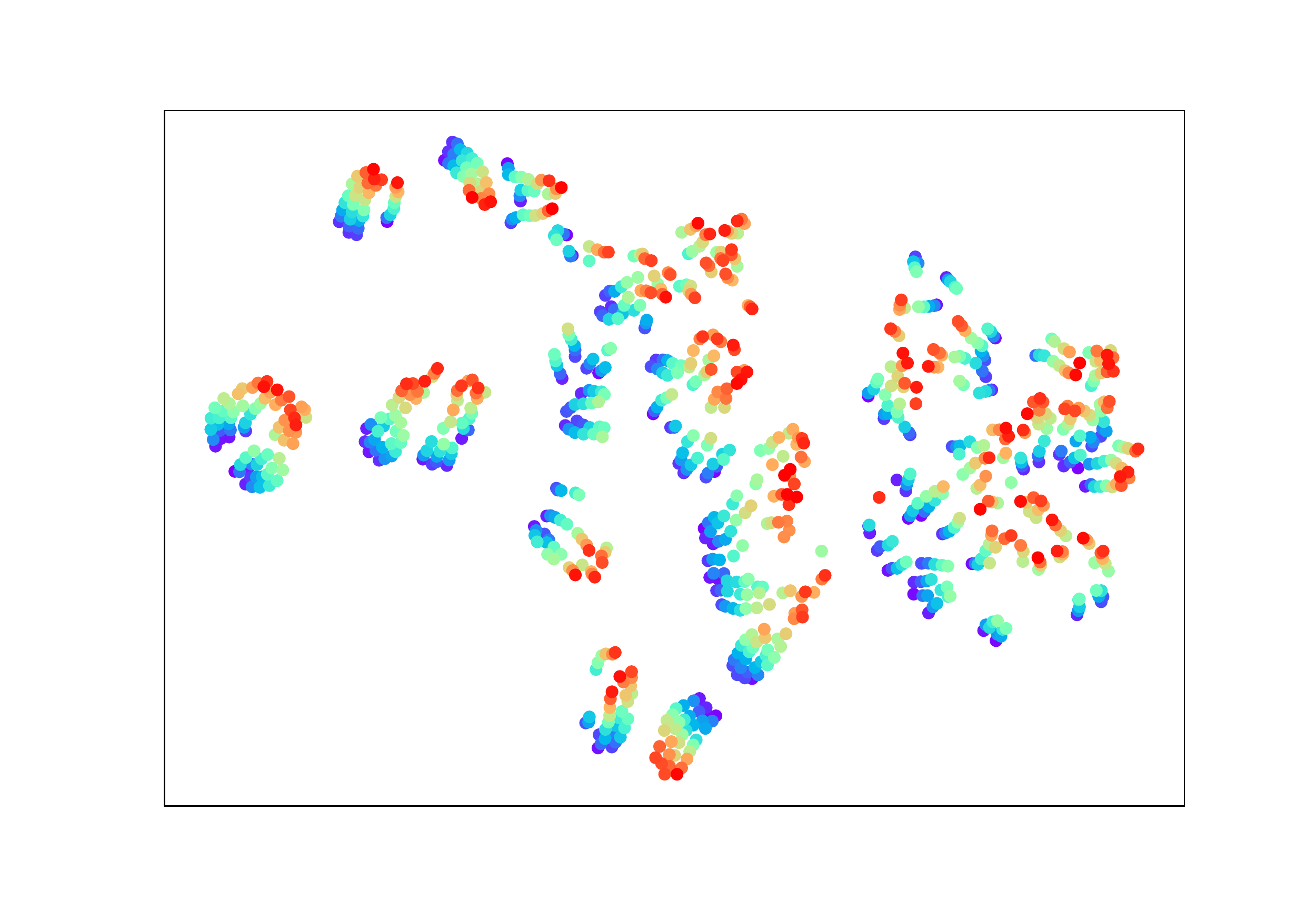}
\caption{VGG16 elevation t-SNE} \label{fig:3picsb}
\end{subfigure}
\smallskip
\begin{subfigure}{0.33\textwidth}
\includegraphics[width=\linewidth]{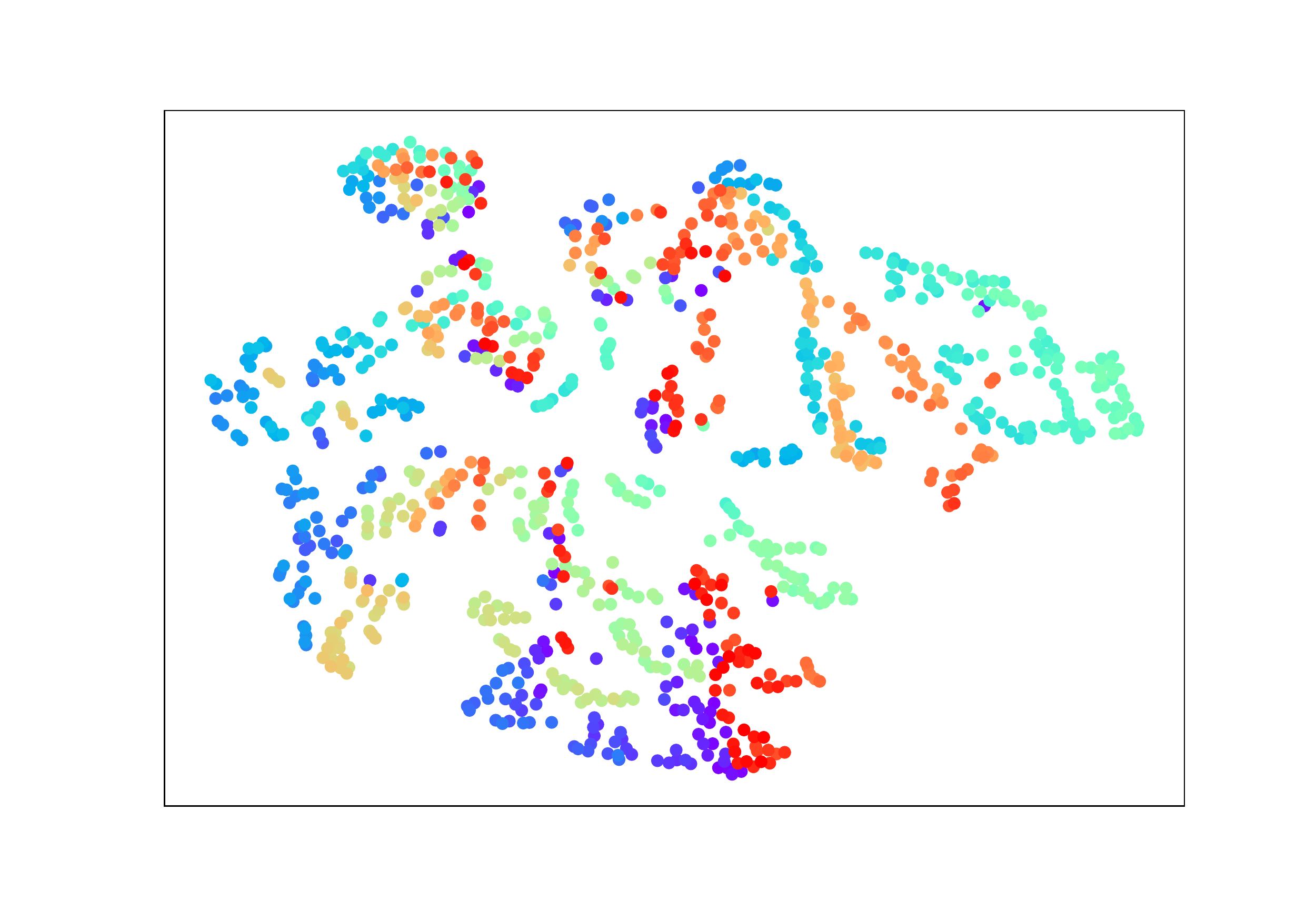} 
\caption{LARP azimuth t-SNE} \label{fig:4picsa}
\end{subfigure} 
\hspace{-2em}
\begin{subfigure}{0.3333\textwidth}
\includegraphics[width=\linewidth]{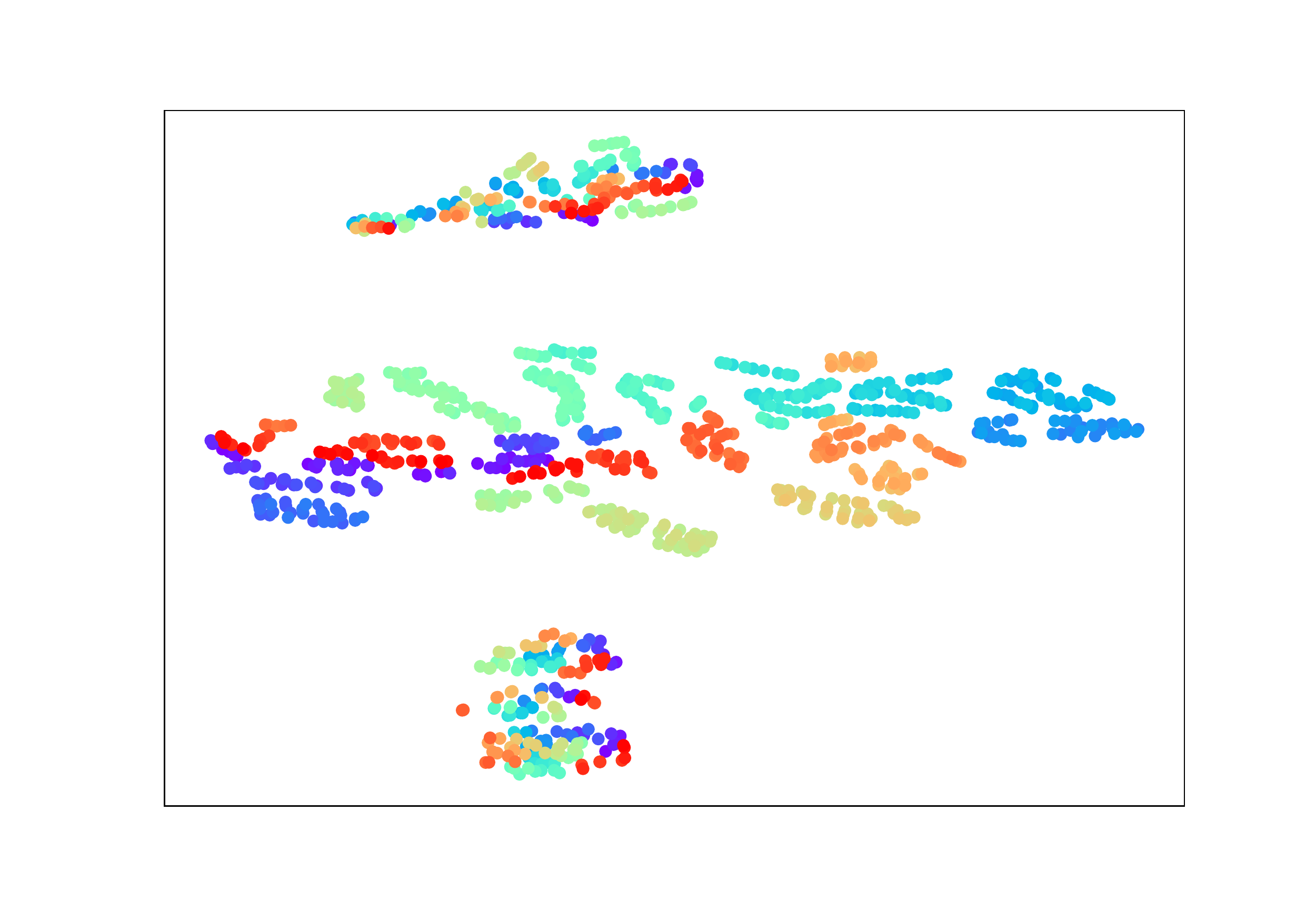}
\caption{CAE azimuth t-SNE} \label{fig:5picsb}
\end{subfigure}
\hspace{-2em}
\begin{subfigure}{0.3333\textwidth}
\includegraphics[width=\linewidth]{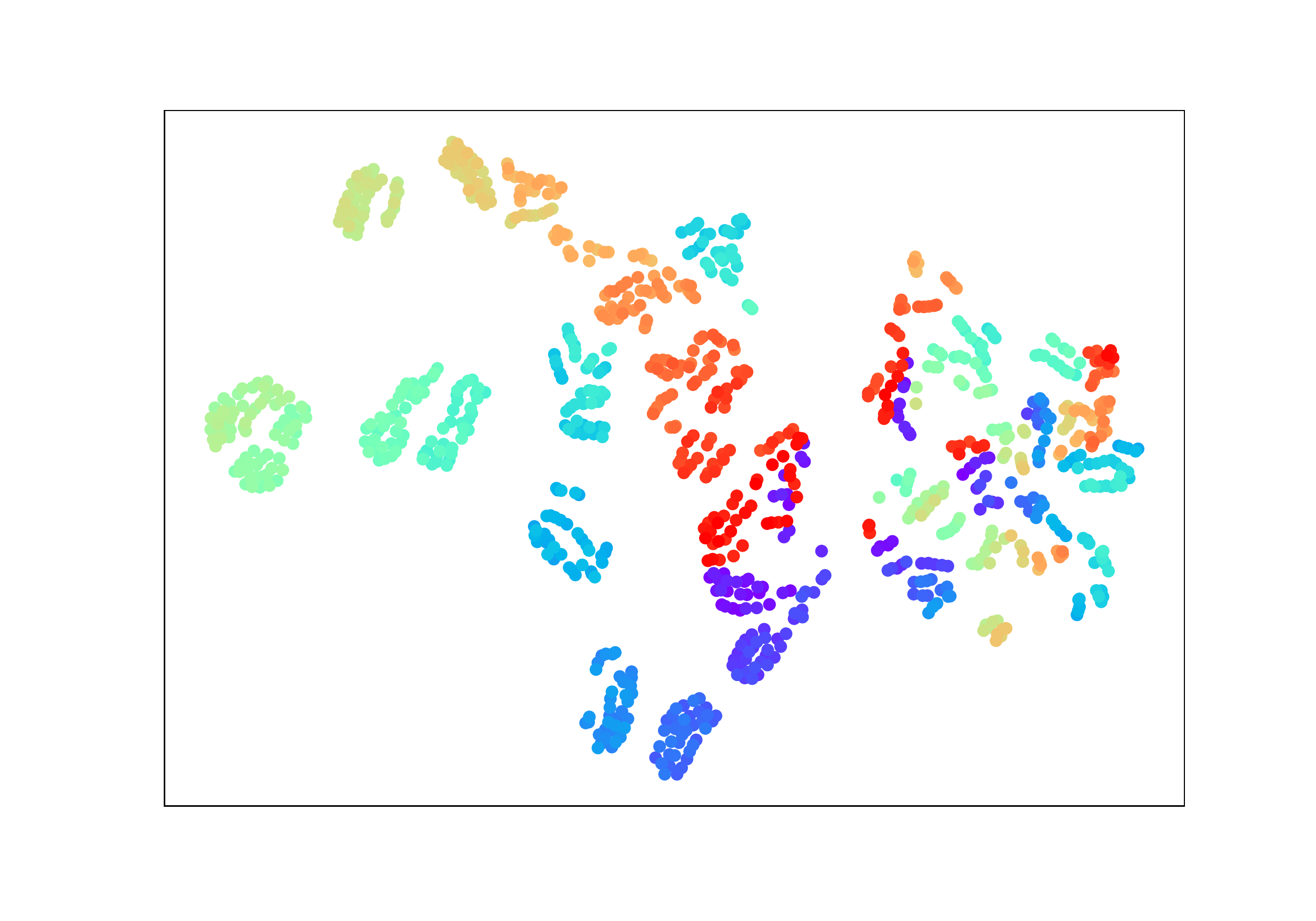}
\caption{VGG16 azimuth t-SNE} \label{fig:6picsb}
\end{subfigure}
\smallskip
\begin{subfigure}{0.33\textwidth}
\includegraphics[width=\linewidth]{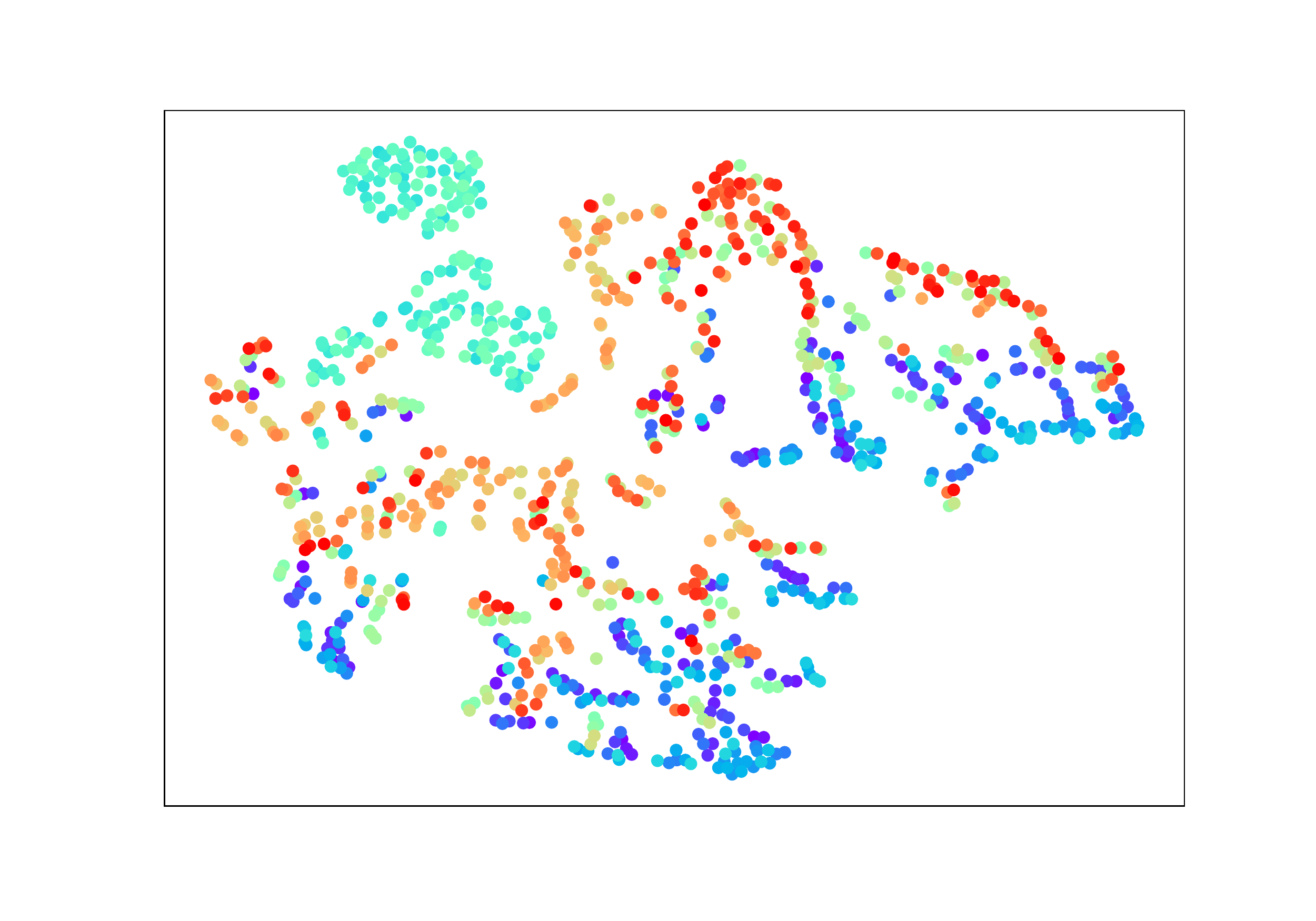}
\caption{LARP lighting t-SNE} \label{fig:7picsc}
\end{subfigure}
\hspace{-0.5em}
\begin{subfigure}{0.33\textwidth}
\includegraphics[width=\linewidth]{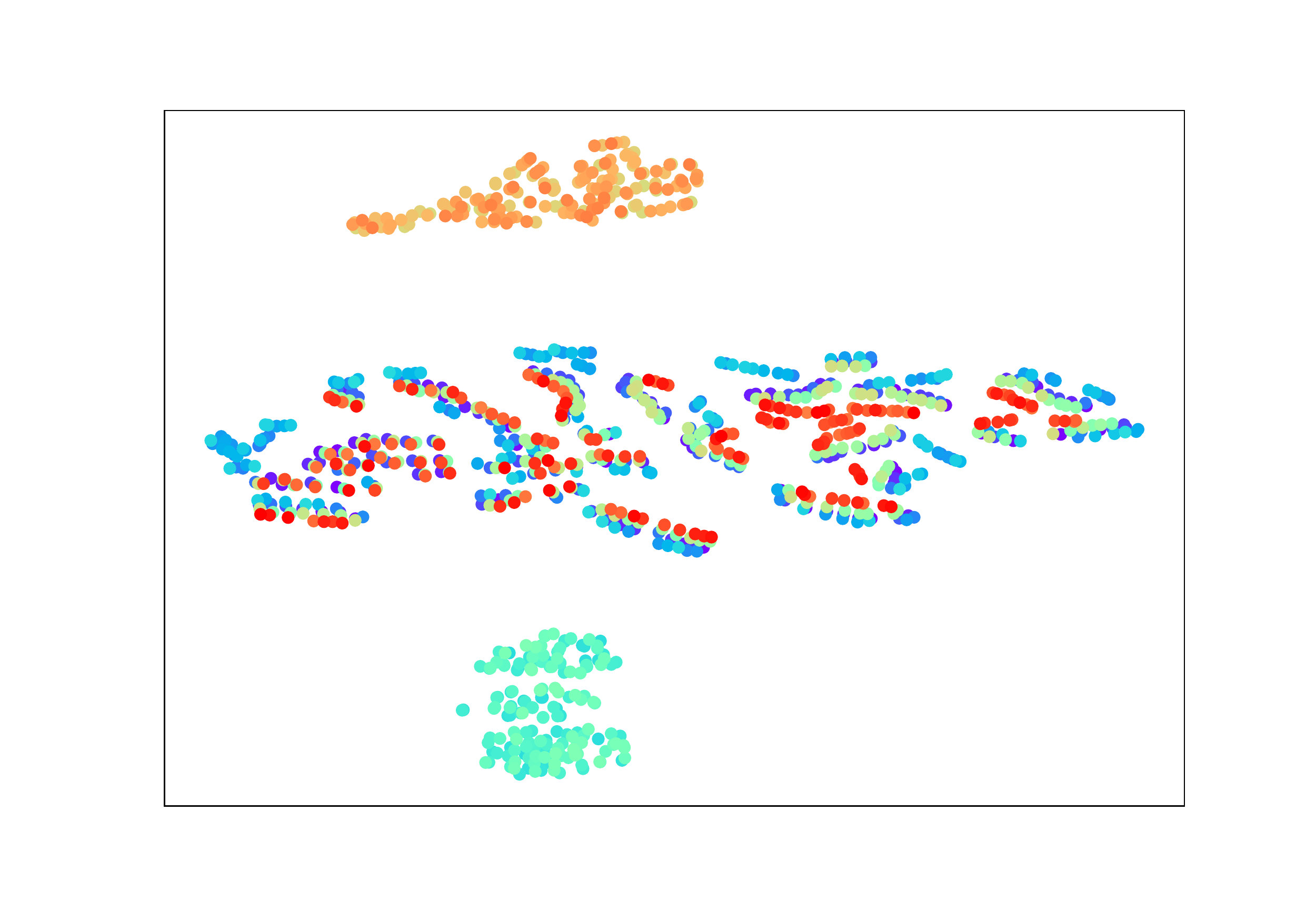}
\caption{CAE lighting t-SNE} \label{fig:8picsd}
\end{subfigure}
\hspace{-0.6em}
\begin{subfigure}{0.33\textwidth}
\includegraphics[width=\linewidth]{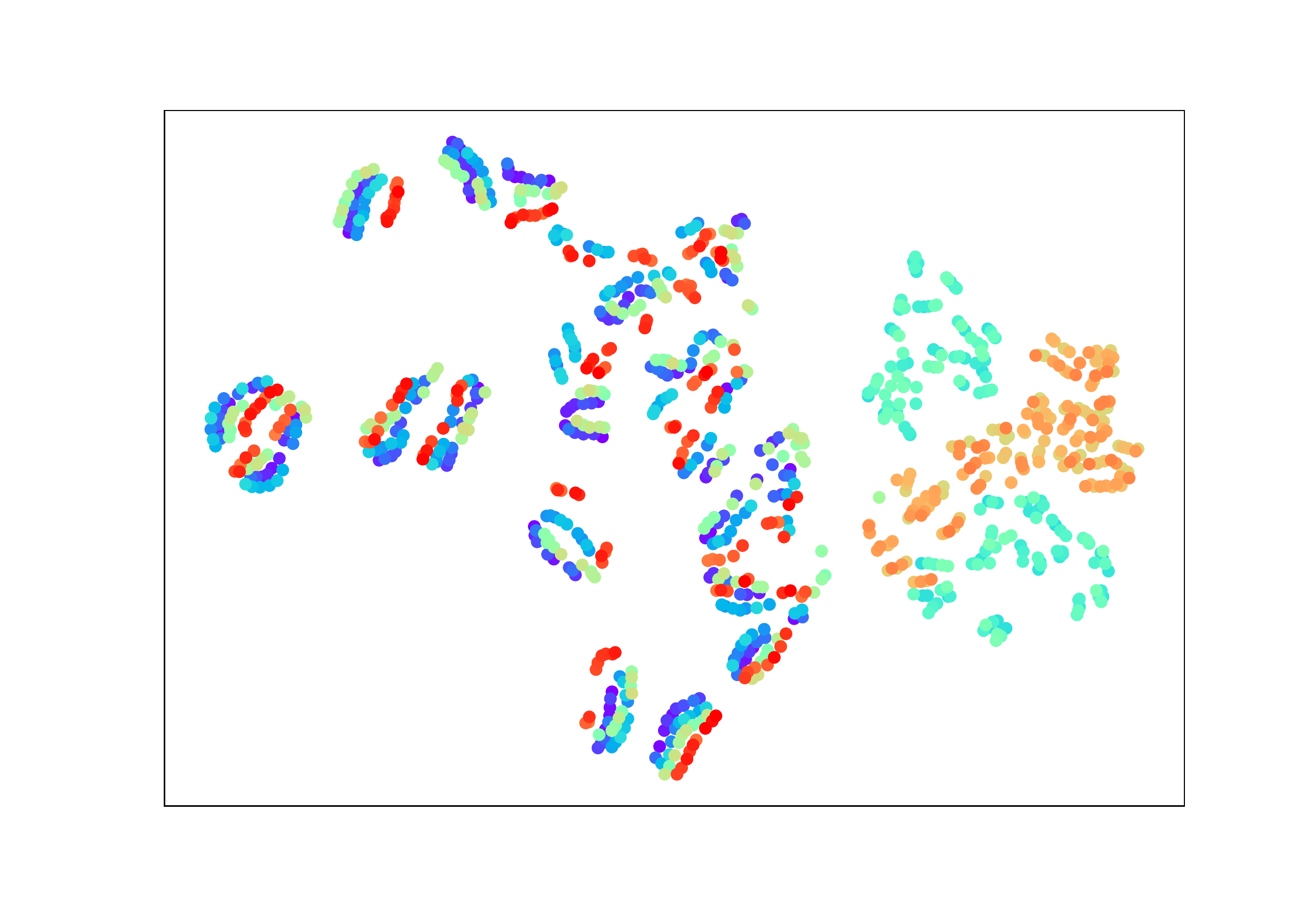}
\caption{VGG16 lighting t-SNE}
\label{fig:9picsb}
\end{subfigure}
\captionsetup{margin=0cm}
\caption[Result: LARP Latent space visualization]{ \textbf{Latent space visualization.} The clustering structure of a toy soldier embedding after dimensionality reduction with t-SNE. The top row colors the samples by elevation, the middle row by azimuth and the bottom one by lighting. The left column is the embedding given by our method using the contrastive loss term, the middle by a convolutional autoencoder (CAE), and the right by a pretrained VGG16 network.} \label{fig:4apics}
\end{figure}

We offer a visualization of the learned representation of the kneeling soldier toy using our method, a convolutional encoder, and VGG16  in Fig.~\ref{fig:4apics}. Each embedding was reduced to 2 dimensions using t-SNE. 

Every method structures the domain similarly. In the bottom row, we see that the largest clusters for all methods are the ones with the highest (teal dots) illumination settings, which is explained by the effect of the lighting on the pixel value intensities. Within these clusters, we see clustering based on the azimuth (middle row). Finally, within these clusters, there is a gradient structure based on elevation (top row). This is due to the elevation changing in smaller step-sizes, with 5 degree differences, than azimuth with 20 degree differences. 

\subsubsection{Quantitative results}

Our experiments include the pretrained VGG16 network because we believe that a flexible algorithm should rather be based on a generic multi-purpose-representation and not on a specific representation.  To our surprise, it is outperformed by our approach on every object, even as the viewpoint-matching task is done on objects that are different from the original car toys they are trained on.

We also try transfer learning a VAE and a convolutional encoder using the same network as our representation and a UMAP embedding, with 64 features each. We display the results of the same viewpoint-matching tasks as before using these different representations on the unseen army figures in Table~\ref{moretesttoys}. Convolutional encoders have the lowest performance out of all the methods, but the VAE representation has the most similar performance to our own representations.

\begin{table}\begin{center}
\caption[Result: LARP transfer learning performance]{ {\bf Transfer learning performance.} The methods are trained on the data set of different car images, and then their performance on dissimilar toys is tested. Each number is the mean success rate of viewpoint matching out of 1000 trials.} 
\label{moretesttoys}
 \begin{tabular}{ l r r r r | r   l} 
  Representation & Soldier & Soldier & Native American   & Cowboy  & 
 \\ [0.5ex] 
 &   (Kneeling)  &  (Standing)  & with Bow & with Rifle & Mean  \\ [0.5ex] 
 \hline 

   LARP (Contrastive)    &  \textbf{22.7} & 11.1 & 12.1 & \textbf{20.7} & \textbf{16.7}\\
   
    LARP (Sphering)   &  18.4 &  15.3&  \textbf{15.2} & 15.8& 16.2\\
    
    LARP (Decoder) &  14.3 &  \textbf{16.6} &  14.7 & 15.6 & 15.3\\ 
    
    VAE   &  14.4 &  13.1 &  13.6 & 13.8 & 13.7\\
    
    VGG16   &  12.3 &  12.9 &  11.8 & 13.6 & 12.7\\ 
    
UMAP   &  8.6 &  10.8 &  9.7 & 11.1 & 10.1\\ 
 
Conv. Encoder  &  4.9 &  14.3 & 11.5 & 6.4 & 9.3 \\ 
    
\end{tabular}\end{center}\end{table}

Our representation is better than the others for this experiment. But we see that the performance of our method drops significantly as we attempt generalizing to different objects with changed input statistics.

Note that even though the representation clustering (Fig.~\ref{fig:4apics}) was more clear-cut using convolutional encoders and VGG, compared to ours, they did not perform as well in the trials. We attribute this to the fact that the main loss term in our representation was the one of future state  predictability, which may not give as clear of a structure in two dimensions as a representation that is trained only for static image reconstruction.
\section{Conclusion}
\label{larpdiscussion}

In this chapter, we present Latent Representation Prediction (LARP) networks with applications to visual planning. We jointly learn a model to predict transitions in Markov decision processes with a representation trained to be maximally predictable. This allows us to accurately search the latent space defined by the representation using a heuristic graph traversal algorithm. We validate our method on a viewpoint-matching task derived from the NORB data set, and we find that a representation that is optimized jointly with the predictor network performs best in our experiments.

A common issue of unsupervised representation learning is one of trivial solutions: a constant representation which optimally solves the unsupervised optimization problem but brings across no information. To avoid the trivial solution, we constrain the training by introducing a sphering layer or a loss term that is either contrastive or reconstructive. Any of these approaches will do the job of preventing the representations from collapsing to constants, and none of them displays stronger performance than the others in our experiments.

Our LARP representation is competitive with pretrained representations for planning and compares favorably to other reinforcement learning (RL) methods. Our approach is a sound solution for learning a useful representation that is suitable for planning only from interactions. 

Furthermore, we find that our method has better data-efficiency during training than several reinforcement learning methods from the literature.
However, a disadvantage of our approach compared to standard RL methods is that the execution time of our method scales worse with the size of the state space, as a forward pass is calculated for each node during the latent space search, potentially resulting in a combinatorial explosion.

Our approach is adaptable to changes in the tasks. For example, our search would only be slightly hindered if some obstacles were placed in the environment or some states were forbidden to traverse through. Furthermore, our method is independent of specific rewards or discount rates, while standard RL methods are usually restricted in their optimization problems. Often, there is a choice between optimizing discounted or undiscounted expected returns. Simulation/rollout-based planning methods are not restricted in that sense: If reward trajectories can be predicted, one can optimize arbitrary functions of these and regularize behavior. For example, a risk-averse portfolio manager can prioritize smooth reward trajectories over volatile ones.

Future lines of work should investigate further the effect of the different constraints on the end-to-end learning of representations suited for a predictive forward model, as well as considering novel ones. The search algorithm can be improved and made faster, especially for higher-dimensional action spaces or continuous ones. Our network could also in principle be used to train an RL system, for instance, by encouraging it to produce similar outputs as ours and thereby combining data-efficiency with fast performance during inference time. 






\clearpage

\chapter{Reward prediction for representation learning and reward shaping}
\label{chap:rewpred}

The previous chapter introduced the LARP network and showed its usefulness in view-point matching experiments, in comparison to reinforcement learning (RL) methods, particularly the model-based ones. The network learns state features that are tailor-made for a prediction module, which is then used for graph-based search. The learning of the state representation and predictor is done in a self-supervised manner, independently of a reward signal. The method is data-efficient and is able to speed to learning for new tasks after pre-training on a similar one.


However, the method requires a forward pass of the network for every node that is considered during the goal search, which scales poorly with the size of the state space compared to methods that only need to process the current state. 
 Depending on the branching factor of the environment and the details of the latent representation graph search, this can result in a combinatorial explosion.

In this chapter\footnote{This chapter is adapted from \citep{hlynsson2021reward}.}, we approach the problem of learning state representations for RL from another angle: instead of predicting the results of actions, we learn a representation that is predictive of the local distance from the goal in single-goal environments. The representation is learned alongside a reward predictor that learns to estimate either a raw or a smoothed version of the true reward signal. 

We augment the training of out-of-the-box RL agents by shaping the reward using our reward predictor during policy learning. Using our representation for preprocessing high-dimensional observations, as well as using the predictor for reward shaping, is shown to significantly enhance Actor Critic using Kronecker-factored Trust Region and Proximal Policy Optimization in single-goal environments with visual inputs.

The remainder of the chapter has the following structure: We start with a chapter introduction in Section \ref{sec:rnintro}, followed by an explanation of required background knowledge for this chapter in Section \ref{sec:backrw}. We go into related work in Section \ref{sec:rnrw}, after which the details of our approach is outlined in Section \ref{chaptern_method}. The methodology of our experiments is explained in Section \ref{sec:exp}. We display and discuss the results of our experiments in Section \ref{sec:rwrd}. Finally, we close the chapter with concluding remarks in Section \ref{sec:disc}.

\section{Introduction}
\label{sec:rnintro}

Even though the dominance of humans is being tested by RL agents on numerous fronts, there are still great difficulties for the field to overcome. For instance, the data that is required for algorithms to reach human performance is on a far larger scale than that needed by humans. Furthermore, the general intelligence of humans remains unchallenged. Even though an RL agent has reached superhuman performance in one field, its performance is usually poor when it is tested in new areas.

The study of methods to overcome the problem of data-efficiency and transferability of RL agents in environments where the agent must reach a single goal is the focal point of this work. We consider a simple way of learning a state representation by predicting either a raw or a smoothed version of a sparse reward yielded by an environment. The two objectives, learning a state representation and predicting the reward, are directly connected as we train a deep neural network for the prediction, and the hidden layers of this network learn a reward-predictive state representation. 

The reward signal is created by collecting data from a relatively low number of initial episodes using a controller that acts randomly. The representation is then extracted from an intermediate layer of the prediction model and re-used as general preprocessing for RL agents, to  reduce the dimensionality of visual inputs. The agent processes inputs corresponding to its current state as well as the desired end state, which is analogous to mentally visualizing a goal before attempting to reach it. This general approach of relying on state representations, that are learned to predict the reward rather than maximizing it, has been motivated in the literature \citep{lehnert2020reward} and we show that our representation is well-suited for single-goal environments.  Our work adds to the recently growing body of knowledge related to deep unsupervised \citep{hlynsson2019learning} or self-supervised \citep{schuler2018gradient} representation learning.

We also investigate the effectiveness of augmenting the reward for RL agents, when the reward is sparse,  with a novel problem-agnostic reward shaping technique. The reward predictor, which is used to train our representation, is not only used as a part of an auxiliary loss function to learn a representation, but it is also used during training the RL system to encourage the agent to move closer to a goal location. Similar to advantage functions in the RL literature \citep{schulman2015high}, given the trained reward predictor, the agent receives an additional reward signal if it moves from states with a low predicted reward to states with a higher predicted reward. We find this reward augmentation to be beneficial for our test environment with the largest state-space.

\section{Background}
\label{sec:backrw}
\subsection{Reward shaping}

Sparse rewards in environments is a common problem for reinforcement learning agents. The agent's objective is to learn how to associate its inputs with actions that lead to high rewards, which can be a lengthy process if the agent only rarely experiences positive or negative rewards. 

Reward shaping \citep{mataric1994reward, ng1999policy, brys2015policy} is a popular method of modifying the reward function of an MDP to speed up learning. It is useful for environments with sparse rewards to augment the training of the agent, but skillful applications of reward shaping can in principle aid the optimization for any environment -- although the efficacy of the reward shaping is highly dependent on the details of the implementation \citep{clark2016faulty}. In the last few years,  reward shaping has been shown to be useful for complex video game environments, such as real-time strategy games \citep{efthymiadis2013using} and platformers \citep{brys2014multi} and it has also been combined with deep neural networks to improve agents in first-person shooter games \citep{lample2017playing}.

As an illustration, consider learning a policy for car racing. If the goal is to train an agent to drive optimally, then supplying it with a positive reward for reaching the finish line first is in theory sufficient. However, if it is punished for actions that are never beneficial, for instance crashing into walls, it prioritizes learning to avoid such situations, allowing it to explore more promising parts of the state space. 

Furthermore, just reaching the goal is insufficient if there is competition. To make sure that we have a winning racer, a small negative reward can be introduced at every time step to urge the agent to reach the finish line quickly. Note that the details of the reward shaping in this example requires domain knowledge from a designer who is familiar with the environment. It would be more generally useful if the reward shaping would be autonomously learned, just as the policy of the agent, as we propose to do in this work.

\subsection{Reward-predictive vs. reward-maximizing representations}
\label{predvsmax}
\begin{figure}[ht!]
  \begin{center}
        \includegraphics[width=0.95\linewidth]{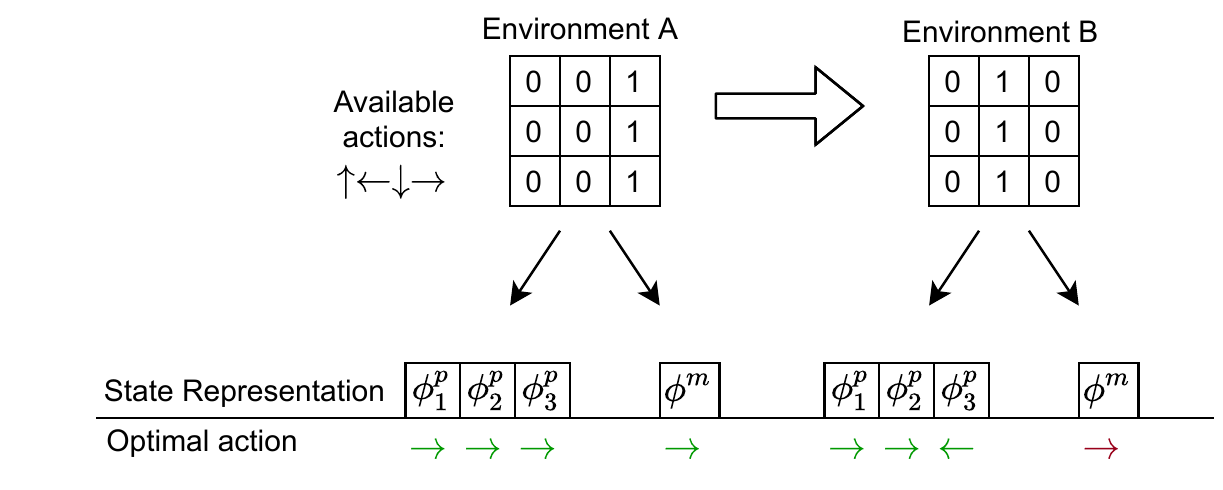}
  \caption[Illustration: Reward-maximizing vs. reward-predictive representations]{\textbf{Reward-maximizing vs. reward-predictive representations}. In this grid world example, the agent starts the episode at a random location and can move up, down, left, or right. The episode ends with a reward of 1 and terminates when the agent reaches the rightmost column. Both the reward-predictive representation and reward-maximizing representation $\phi^p$ and $\phi^m$, respectively, are useful for learning the optimal policy in Environment A. The reward-predictive representation $\phi^p$ collapses each column into a single state to predict the discounted future reward. The reward-maximizing representation $\phi^m$ makes no such distinction, as moving right is the optimal action in any state. It is a different story if the representations are transferred to Environment B, where reaching the middle column is now the goal. The representation $\phi^p$ can be reused, and the optimal policy is found if the agent now takes a step left in $\phi_3^p$. However, the representation $\phi^m$  is unable to discriminate between the different states and is useless for determining the optimal policy.}
  \label{fig:transferrep}
  \end{center}
\end{figure}

\cite{lehnert2020reward} make the distinction between \textit{reward-maximizing} representations are \textit{reward-predictive} representations. They argue how reward-maximizing representations can transfer poorly to new environments, while reward-predictive representations generalize successfully. Take the simple grid world navigation environments in Fig. \ref{fig:transferrep}, for example. The agent starts at a random tile in the grid and gets a reward of +1 by reaching the rightmost column in Environment A or by reaching the middle column in Environment B. 
 The state space in Environment A can be compressed from the $3 \times 3$ grid to a vector of length 3, $[\phi_1^p, \phi_2^p, \phi_3^p]$ of reward-predictive representations. To predict the discounted reward, it suffices to describe the agent's state with $\phi_j^p$ if it is in the $j$th row. 
 
 The reward-maximizing representation for Environment A is much simpler: the whole state space can be collapsed to a single element $\phi^m$, with the optimal policy of always moving to the right. 
 If these representations are kept, then the reward-predictive representation $\phi^p$ is informative enough for a RL agent to learn how to solve Environment B. The reward-maximizing representation $\phi^m$ has discarded too many details of the environment to be useful for solving this new environment.

\subsection{Successor features}
The \textit{successor representation} algorithm learns two functions: the expected reward $R_\pi^{\text{SF}}$ received after transitioning into a state $s$, as well as the matrix $M_\pi^{\text{SF}}$ of discounted expected future occupancy of each state,  assuming that the agent starts in a given state and follows a particular policy $\pi$.  Knowing the quantities $R_\pi^{\text{SF}}$ and $M_\pi^{\text{SF}}$ allows us to rewrite the value function:

 \begin{equation} \label{eq:sr0}
V_{\pi}(s) = \mathbb{E}_{s'}\left[ R_\pi^{\text{SF}}(s) M_\pi^{\text{SF}}(s, s')\right]\end{equation}

The motivation for this algorithm is that it combines the speed of model-free methods, by enabling fast computations of the value function, with the flexibility of model-based methods for environments with changing reward contingencies. 

This method is made for small, discrete environments, but it has been generalized for continuous environments with so-called feature-based successor representations, or \textit{successor features} (SFs) \citep{barreto2016successor}.  The SF algorithm similarly calculates the discounted expected representation of future states, given the agent takes the action $a$ in the state $s$ and follows a policy $\pi$:

\begin{equation} \label{eq:sf}
\psi_{\pi}(s, a) = \mathbb{E}_\pi \left[\sum_{t=0}^\infty \gamma^{t-1} \phi_{t+1} | s_t = s, a_t = a\right]\end{equation}




\noindent where $\phi$ is some state representation. Both the SF $\psi$ and the representation $\phi$ can be deep neural networks.

\section{Related work}
\label{sec:rnrw}

 \subsection{Reward-predictive representations } 
\cite{lehnert2020reward} compare successor features (SFs) to a nonparametric Bayesian predictor that is trained to learn transition and reward tables for the environment, either with a reward-maximizing or a reward-predictive loss function. \cite{lehnert2020successor} prove under what conditions successor features (SFs) are either \textit{reward-predictive} or \textit{reward-maximizing} (see distinction in Section \ref{predvsmax}). They also show that SFs work successfully for transfer learning between environments with changing reward functions and unchanged transition functions, but they generalize poorly between environments where the transition function changes. 
 
Our work is distinct from the reward-predictive methods that they compare, as our representation does not need to calculate expected future state occupancy, as is the case for SFs. Our method scales better for more complicated state-spaces because we do not tabulate the states, as they do with their Bayesian model, but learn arbitrary continuous features of high-dimensional input data. In addition to that, learning our reward predictor is not only a "surrogate" objective function, as we use it for reward shaping as well.

\subsection{Reward shaping}

The advantages of reward shaping are well understood in the literature \citep{mataric1994reward}. A recent trend in RL research is the study of methods that can learn the reward shaping function automatically, without the need of (often faulty) human intervention. 

\cite{marashi2012automatic} assume that the environment can be expressed as a graph and that this graph formulation is known. Under these strong assumptions, they perform graph analysis to extract a reward shaping function. More recently, \cite{zou2019reward} have proposed a meta-learning algorithm for potential-based automatic reward shaping. Our approach is different from previous work as we assume no knowledge about the environment and train a simple predictor to approximate (potentially smoothed) rewards, which is then used to construct a potential-based reward shaping function.






\subsection{Goal-conditioned reinforcement learning } 
\cite{kaelbling1993learning} studied  environments with multiple goals and small state-spaces.  In their problem setting, the agent must reach a known but dynamically changing goal in the fewest number of moves. The observation space is of a low enough dimension for dynamic programming to be satisfactory in their case. \cite{schaul2015universal} introduce the  Universal Value Function Approximators and tackle environments of larger dimensions by  learning a value function neural network approximator that accepts both the current state and a goal state as the inputs. In a similar vein, \cite{pathak2018zero} learn a policy that is given a current state and a goal state and outputs an action that bridges the gap between them. \cite{hlynsson2020latent} learn a predictable representation that is paired with a representation predictor and combine it with graph search to find a given goal location. In contrast to these approaches, we learn a reward-predictive representation in a self-supervised manner, which is used to preprocess raw inputs for RL policies.

\section{Approach}
\label{chaptern_method}

In this section, we explain our approach mathematically. Intuitively, we train a deep neural network to predict either a raw or a smoothed reward signal from a single-goal environment. The output of an intermediate layer in the network is then extracted as the representation -- for example, by simply removing the top layers of the network. The full reward predictor network is used for reward shaping by rewarding the agent for moving from lower predicted values toward higher predicted values of the network.

\subsection{Learning the representation}

Suppose that $f_\theta: \mathbb{R}^c \rightarrow [0, 1]$ is a differentiable function parameterized by $\theta$ and $c$ is a positive integer. We use $f_\theta$ to approximate the discounted return in a POMDP with a sparse reward: the agent receives a reward of 0 for each time step except when it reaches a goal location, at which point it receives a positive reward and the episode terminates.

Given an experience buffer $\mathcal{D} = $ $\{(s_t, a_t, r_t, s_{t+1})_i\}$, we create a new data set $\mathcal{D}^* = \{ \left(s_t, a_t, r^*_t, s_{t+1} \right)_i \}$. The new rewards are calculated according to the equation  

\begin{equation} \label{eq:rstar}
r^*_t = \gamma^m r_{t+m}\end{equation}

\noindent where $\gamma \in [0, 1]$ is a discount factor and $M>m>0$ is the difference between $t$ and the time step index of the final transition in that episode, for some maximum time horizon $M$. Throughout our experiments, we keep the value of the discount factor equal to $0.99$ and we train on $\mathcal{D} $ or $\mathcal{D}^*$.

\begin{figure}[ht]
\begin{center}
  \includegraphics[width=.92\linewidth]{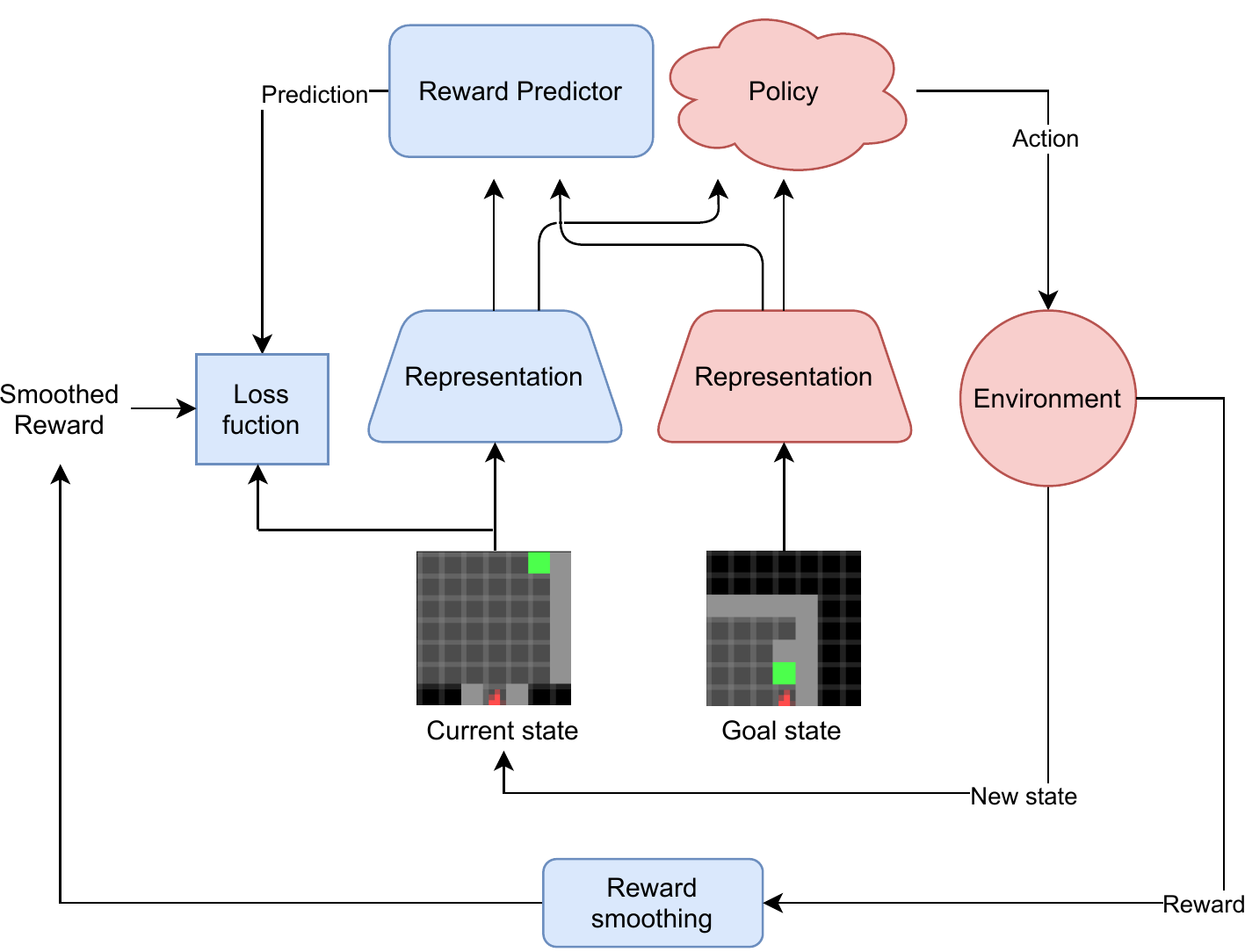}
  \caption[Illustration: Learning and using the representation]{\textbf{Learning and using the representation.} Our representation and reward predictor is trained with the elements highlighted in blue. The trained representation is then used for dimensionality reduction for an RL agent, that interacts with the environment, as indicated by the elements highlighted in red.  }
  \label{fig:boat5}
\end{center}
\end{figure}

Assume that our differentiable representation function $\phi: \mathbb{R}^d \rightarrow \mathbb{R}^c$ is parameterized by $\theta'$ and maps the $d$-dimensional raw observation of the POMDP to the $c$-dimensional feature vector. We train the representation for the discounted-reward prediction by minimizing the loss function

\begin{equation}
\mathcal{L}(f_\theta [ \phi_{\theta'} (s_{t+1})], r^*_t) =  \left( r^*_t - f_\theta [ \phi_{\theta'} (s_{t+1})] \right)^2
    \label{discounted_loss}
\end{equation}

\noindent with respect to the parameters $\theta$ of $f$ and the parameters $\theta'$ of $\phi$ over the whole data set $\mathcal{D}^*$. See Fig. \ref{fig:boat5} for a conceptual overview of our representation learning. 

\subsection{Reward shaping}

\cite{ng1999policy} define a reward shaping function $F$ as  \textit{potential-based} if there exists a function $f: \mathcal{S} \rightarrow \mathbb{R}$ such that for all states $s, s' \in \mathcal{S}$ the following equation holds:

\begin{equation}
F(s, a, s') = \gamma f(s') - f(s)  
    \label{potential_based}
\end{equation}

\noindent and $\gamma$ is the MDP's discount factor. They prove for single-goal environments that every optimal policy for the MDP $M = (\mathcal{S}, \mathcal{A}, \mathcal{P}, \mathcal{R}, \mathbb{P}(s_0), \gamma)$ is also optimal for its reward shaped counterpart $M' = (\mathcal{S}, \mathcal{A}, \mathcal{P}, \mathcal{R}+F, \mathbb{P}(s_0), \gamma)$, and vice versa. They also show, for a given state space $\mathcal{S}$ and action space $\mathcal{A}$, that if $F$ is not potential-based, then there exist a transition function $\mathcal{P}$ and a reward function $\mathcal{R}$ such that no optimal policy in $M'$ is optimal in $M$.

Deciding that the reward shaping function should be potential-based is just the first step in its design. Now assume that we have an environment where an agent is tasked with reaching a goal state $g$. That is, for a given distance function $d: \mathcal{S} \times \mathcal{S} \rightarrow \mathbb{R}^+$ the agent receives a reward of 1 if it is close enough to the goal location, $d(s, g) \leq \delta$, for some reward threshold $\delta \in \mathbb{R}^+$. Otherwise, it receives a reward of 0.

The distance $d$ between the agent's location new $s'$ and the goal location $g$ can be a useful value to calculate in the design of a reward shaping function

\begin{equation}
F(s, a, s') = \begin{cases} 1 & \text{if } d(s', g) \leq \delta \\
                      -d(s', g)                                    & \text{otherwise}      %
        \end{cases}    
\label{negdist}
\end{equation}
However, this depends on the environment, as the agent could get stuck in local optima before coming close to the goal, i.e. if it would have to move through a region with a large $d(s, g)$ before it can globally minimize it. This could for example be the case in a maze environment if $d$ is the Euclidean distance between the $(x, y)$ coordinates of the agent's location and the goal location and there is a wall between the agent and the goal.  \cite{trott2019keeping} propose to solve this by incorporating the potential local optima in the reward shaping function as so-called "anti-goals" $\Bar{g}$ to be avoided

\begin{equation}
F(s, a, s') = \begin{cases} 1 & \text{if } d(s', g) \leq \delta \\
                      \min [0, -d(s', g) + d(s, \Bar{g})]                               & \text{otherwise}      %
        \end{cases}    
\label{antigoals}
\end{equation}

These states can be hand-picked by domain experts. However, adding anti-goals like this could iteratively introduce even more local optima and a solution to the original problem is not guaranteed.

It is generally not true that the distance function $d$ and all the variables needed to calculate it, such as the coordinates of the agent and the goal in a maze, are available to the agent. Even if $d$ were computable, using it naively can bring about its own problems, as was alluded to above. 

We argue that instead of using $d$ in Equation~\ref{negdist}, it would be better to measure the distance between the agent and the goal in terms of how many actions the agent has to take until the goal is reached. This function is not assumed to be given, but it can be estimated as the agent is being trained on the environment, for instance by optimizing Equation~\ref{discounted_loss}.

Additionally, we would like our reward shaping function to be potential-based (Equation \ref{potential_based}) to reap the theoretical advantages. Thus, we propose a potential-based reward shaping function based on the discounted reward predictor


\begin{equation}
\begin{split}
F(s, a, s') & = \left( \gamma f_\theta ( \phi_{\theta'} [s']) - f_\theta ( \phi_{\theta'} [s]) \right)  (H - I) / H \\
 & = \gamma \left( f_\theta ( \phi_{\theta'} [s']) (H - I) / H \right) - f_\theta ( \phi_{\theta'} [s]) (H - I) / H \\
 & = \gamma f^*(s') - f^*(s)
\end{split}
\label{ourshaped}
\end{equation}

\noindent where $f^* =  f_\theta ( \phi_{\theta'} [s']) (H - I) / H $, $f_\theta$ is the reward predictor and $\phi_{\theta'}$ is our representation from the previous section. Note that both $f_\theta$ and $\phi_{\theta'}$ are assumed to be fully trained before the policy of the agent is trained, for example using data gathered by a random policy, but they can in principle also be updated as the policy is being learned. The factor $(H - I) / H$ scales down the intensity of the reward shaping where  $I \in \mathbb{N}^+$ is the number of episodes that the agent has experienced and $H \in \mathbb{N}^+$ is the maximum number of episodes where the agent is trained using reward shaping.  The strength of the reward shaping is the highest in the beginning to counteract potentially adverse effects of errors in the reward predictor. It is also more important to incentivize moving toward the general direction of the goal in the early stages of learning, after which the un-augmented reward signal of the environment is allowed to "speak for itself" and guide the learning of the agent toward the goal precisely.




\section{Methodology and implementation} \label{sec:exp}

\subsection{Environment} \label{subsec:env}

\begin{figure}
\centering
\begin{subfigure}[b]{0.8\textwidth}
   \includegraphics[width=1\linewidth]{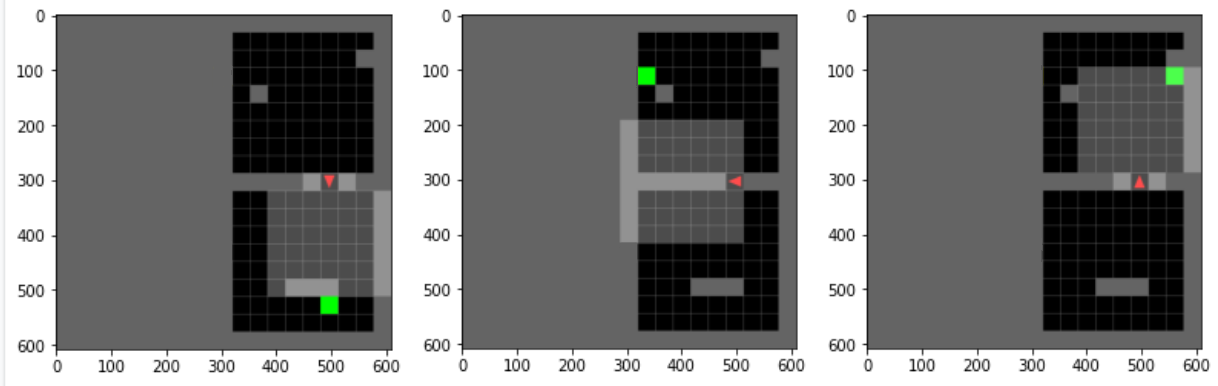}
   \caption{Full world states.}
   \label{fig:Nw1b} 
\end{subfigure}
\begin{subfigure}[b]{0.8 \textwidth}
   \includegraphics[width=1\linewidth]{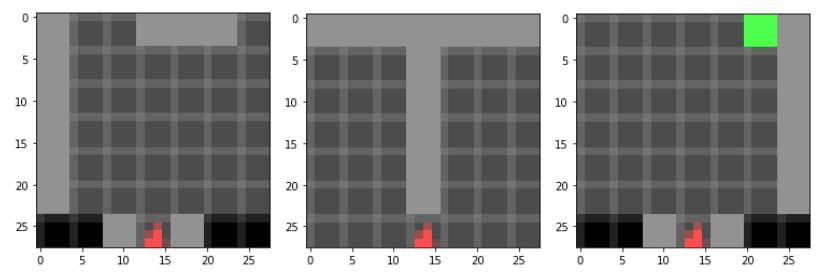}
   \caption{Agent's point of view.}
   \label{fig:Nw2obses}
\end{subfigure}
\begin{subfigure}[b]{0.8\textwidth}
   \includegraphics[width=1\linewidth]{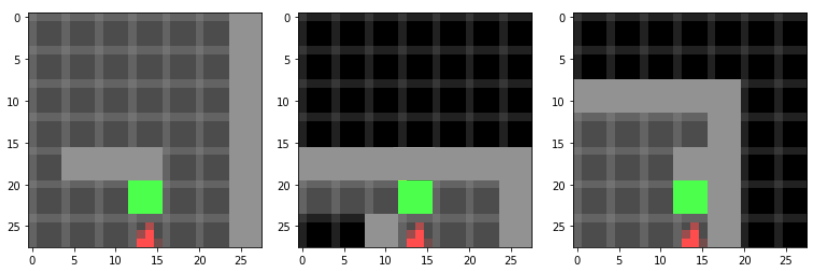}
   \caption{Goal observations.}
   \label{fig:Nw2targets}
\end{subfigure}
  \caption[Example: The two-room environment]{\textbf{Two-room environment.} The red agent must reach the green goal in as few steps as possible. The agent starts each episode between the two rooms, facing a random direction (up, down, left or right).  Each column corresponds to a snapshot of one episode. The light tiles correspond to what the agent sees, while the dark tiles are unseen by the agent. (a) Examples of the full state (b)~The observation from the agent's current state (c) A goal observation. This is the agent's point of view from a state that separates the agent from the goal by one action. }
  \label{tworooms}
  \end{figure}

The method is tested on three different gridworld environments based on the Minimalistic Gridworld Environment (MiniGrid) \citep{gym_minigrid}. Tiles can be empty, occupied by a wall or occupied by lava. The structure of the environments fit naturally into our POMDP tuple template (Eq. \ref{eq:pomdp}): 
\begin{itemize}
\setlength{\itemindent}{0.65cm}
    \item[$\mathcal{S}$:]The constituent states of $\mathcal{S}$ are determined by the agent's location and direction (facing north, west, south or east). See Fig. \ref{fig:Nw1b} for three different world states in one of our environments. 
    \item[$\mathcal{A}$:]The action space $\mathcal{A}$ consists of three actions: (1) turn left, (2) turn right and (3) move forward. 
    \item[$\mathcal{P}$:] The transition function is deterministic. The agent relocates to the tile it faces if it moves forward and the tile is empty, and nothing happens if the tile is occupied by a wall. The episode terminates if the tile is occupied by lava or the goal. The agent rotates in place if it turns left or right. 
    \item[$\mathcal{R}$:] Reaching the green tile goal gives a reward of $1 - 0.9 \cdot \frac{\# \textrm{steps taken} }{\# \textrm {max steps}} $, every other action gives 0 points. The environment automatically times out after $\# \textrm {max steps}=100$ steps.
        \setlength{\itemindent}{1.3cm}
    \item[$\mathbb{P}(s_0)$:] Differs between the three environments (see below).
    \setlength{\itemindent}{0.65cm}
    \item[$\Omega$:] All $7\times7$ subset of tiles, represented by $28\times28\times3$ arrays, from the point of view of an agent who cannot see through walls, see Fig. \ref{fig:Nw2obses}.
    \item[$\mathcal{O}$:] The point of view of the agent from its current viewpoint (Fig. \ref{fig:Nw2obses}) and a goal observation (Fig. \ref{fig:Nw2targets}).
    \item[$\gamma$:] the discount factor is $0.99$.
\end{itemize}

  We consider the following three environments:

  \subsubsection{Two-room environment}
  \label{tworoomimplementation}
  The world is a $8 \times 17$ grid of tiles, split into two rooms, where walls are placed at different locations to facilitate discrimination between the rooms from the agent's point of view. (Fig. \ref{tworooms}). The agent is placed between the two rooms, facing a random direction. The goal is at one of three possible locations. This is a modified version of the classical four-room environment layout \citep{sutton1999between}.

  \subsubsection{Lava gap environment}
  
\begin{figure*}[ht]
	\centering
	\begin{minipage}{.83\columnwidth}
		\centering
		\includegraphics[width=\textwidth]{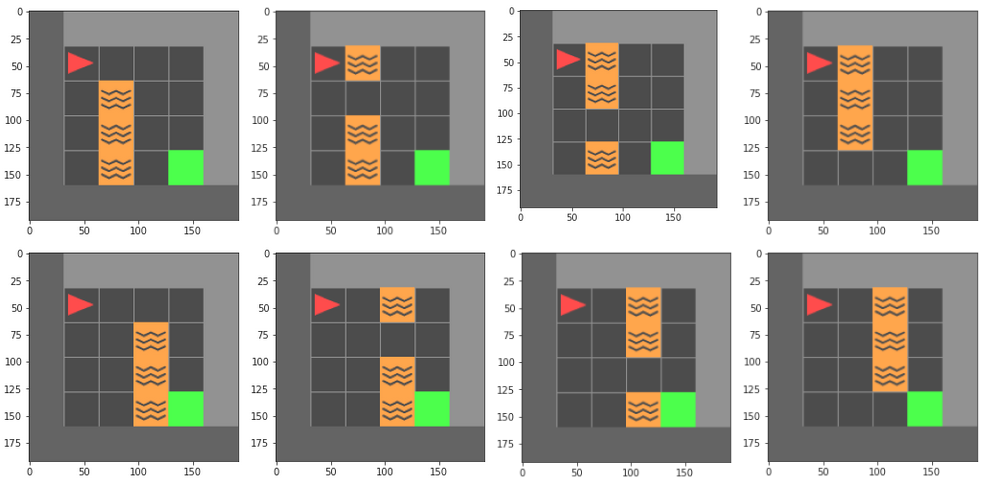}
	\end{minipage}
		\caption[Example: The lava gap environment]{ \textbf{The lava gap environment.} The red agent must reach the green goal in as few steps as possible while avoiding the orange lava tiles. Each episode is randomly selected to be one of the eight pictured configurations. If the agent tries to go through the orange lava tile, it experiences an immediate episode termination with no reward. Note that the wall tiles that are lighter than the others are presently in the agent's field of vision. }
		\label{fig:Ng-1}
\end{figure*}

  In this environment, the agent is in a $4 \times 4$ room with a column of lava either one or two spaces in front of the agent (Fig. \ref{fig:Ng-1}) with a gap in a random row. The agent always starts in the upper left corner and the goal is always in the lower right corner.

  \subsubsection{Four-room environment}
    \label{fourroomimplementation}
\begin{figure*}[ht]
	\centering
	\begin{minipage}{.67\columnwidth}
		\centering
		\includegraphics[width=\textwidth]{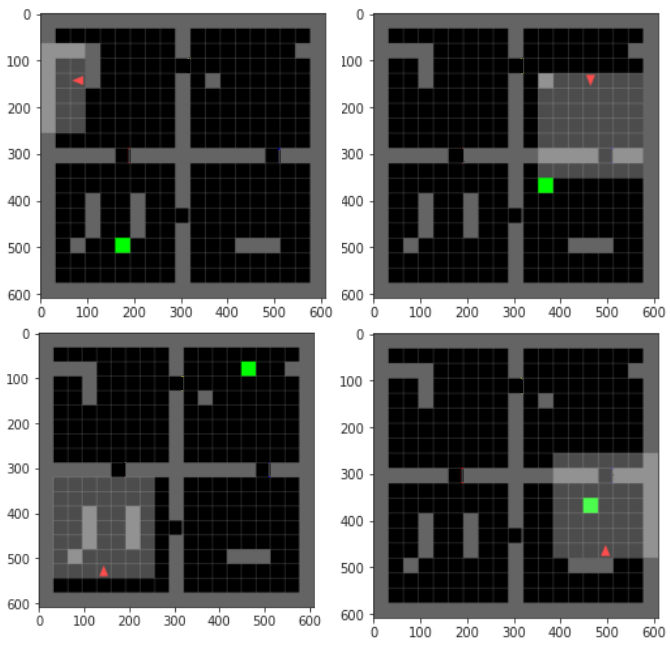}
	\end{minipage}
		\caption[Example: Four-room environment]{ \textbf{Four-room environment}. The agent (red triangle) must reach the goal (green square) in as few steps as possible, both are randomly placed in each episode. The $7\times7$ grid of highlighted tiles in front of the agent indicates its observation.}
		\label{fullillus}
\end{figure*}

An expansion to the two-room environment with two additional rooms (Fig. \ref{fullillus}). In this setup, both the agent and the goal location are placed at random locations within the $17\times17$ gridworld.
  
\subsection{Baselines} \label{subsec:cm}
 We combine our representations with two RL algorithms as implemented in Stable Baselines \citep{stable-baselines} using the default hyperparameters:

\begin{itemize}

\item (ACKTR) Actor Critic using Kronecker-Factored Trust Region \citep{wu2017scalable}, which combines actor-critic methods, trust-region optimization, and distributed Kronecker factorization to enhance data-efficiency. 
\item (PPO2) A version of the Proximal Policy Optimization algorithm \citep{schulman2017proximal}. It modifies the original algorithm by using clipped value functions and a normalized advantage.
\end{itemize}

For both algorithms, six variations are compared:

\begin{itemize}
    \item (Deep RL) The RL algorithm learns the representation from scratch on raw images
    \item (SF) The input is preprocessed using successor features
    \item (Ours 1r) The input is preprocessed using our representation, trained on raw reward predictions
    \item (Ours 1r + Shaping) The input is preprocessed using our representation and the reward is shaped, trained on raw reward predictions
   \item (Ours 64r) The input is preprocessed using our representation, trained on smoothed reward predictions
    \item (Ours 64r + Shaping) The input is preprocessed using our representation and the reward is shaped, trained on smoothed reward predictions

\end{itemize}

Care has been taken to ensure that each variation has the same architecture and the same number of parameters. 
\subsection{Model architectures} \label{subs:ma}

Every model is realized as a neural network using Keras \citep{chollet2015keras}. Below, the representation and policy networks are used for our method and the SF comparison, the reward prediction network is used only for our method and the deep RL network is used only for the deep RL comparison, where the RL algorithm also learns the representation. 

\textbf{The representation networks} are two convolutional networks (Table \ref{tab:repnet})  with a $28 \times 28 \times 3$ input, taking either the agent's current observation or the goal observation.

\begin{table}[ht]
    \centering \begin{tabular}{|l @{\hskip 0.2in} r @{\hskip 0.2in} r @{\hskip 0.2in} l @{\hskip 0.2in} r @{\hskip 0.2in} c|} 
     \hline\rule{0pt}{2.2ex}
     \textbf{Layer} & \textbf{Filters} & \textbf{Filter size} & \textbf{Stride}  & \textbf{Padding} & \textbf{Output shape} \\ [0.5ex] 
     \hline\rule{0pt}{2.2ex}Input tensor & - & -  &-&-&$28\!\times\!28\!\times\!3$  \\[1ex] 
       \hline \rule{0pt}{2.2ex}Convolution & 8 & $3\!\times\!3$  & 3 & None & $9\!\times\!9\!\times\!8$ \\[.5ex] 
      ReLU & - & -  & -&- & $4\!\times\!4\!\times\!8$  \\[.5ex] 
       \hline\rule{0pt}{2.2ex}2D max pooling & 8 & $2\!\times\!2$  & -& None & $4\!\times\!4\!\times\!8$  \\[.5ex] 
     \hline \rule{0pt}{2.2ex}Convolution     & 16 & $3\!\times\!3$  & 2&None & $1\!\times\!1\!\times\!16$   \\[.5ex] 
       ReLU & - & -  & -&-& $1\!\times\!1\!\times\!16$ \\[.5ex] 
       \hline\rule{0pt}{2.2ex}Flatten & - & -  & - &-& 16 \\[.5ex] 
       Dense & - & -  & - &-&16 \\[.5ex]

    \hline
    \end{tabular}
    \vspace{0.1cm}
    \caption[Result: Representation network]{\textbf{Representation network.}}
    \label{tab:repnet}
\end{table}

The first layer subsamples the input, keeping only every other column and row. This is followed by 8 filters of size $3 \times 3$ with a stride of 3. This is followed with a ReLU activation and a $2 \times 2 $ max pooling layer with a stride value of 2. The pooling layer's output is passed to a layer with 16 convolutional filters of size $3 \times 3$ and a stride of 2 and a ReLU activation function. The output is then flattened and passed to a dense layer with 16 units and a linear activation, defining the dimension of the representation. No zero padding is applied in the convolutional layers or the pooling layer.

\textbf{The policy networks} are three-layer fully-connected networks (Table \ref{tab:polnet1}) accepting the concatenated output of the representation network for the agent's current point of view and the goal observation as an input. The first two layers have 64 units and a ReLU activation, and the last layer has 3 units and a linear activation function. The three units represent the three actions left, right, and forward in a one hot encoding. Winner takes all is used to decide on
the action.

\begin{table}[ht]
    \centering \begin{tabular}{|l @{\hskip 0.3in} r @{\hskip 0.3in} c|}
     \hline\rule{0pt}{2.2ex}
     \textbf{Layer} & \textbf{Units} &  \textbf{Output shape}   \\ [0.5ex] 
     \hline\rule{0pt}{2.2ex}Input tensor &-&32  \\[1ex] 
     \hline \rule{0pt}{2.2ex}Dense & 64 & 64 \\[.5ex] 
      \rule{0pt}{2.2ex}ReLU & - & 64 \\[.5ex] 

\hline Dense & 64 &3 \\[.5ex] 
ReLU & -  & 3 \\[.5ex] 

\hline Dense & 3 & 3 \\[.5ex] 
    \hline
    \end{tabular}
    \vspace{0.1cm}
    \caption[Result: Policy network]{\textbf{Policy network.}}
    \label{tab:polnet1}
\end{table}

\textbf{Our reward prediction network} is a three-layer fully-connected network (Table \ref{tab:polnet3}) with the same input as the policy network: the concatenated representation of the agent's current view and the goal observation. The first two layers have 256 units and a ReLU activation, but the last layer has 1 unit and a logistic activation function. 

\begin{table}[ht]
    \centering \begin{tabular}{|l @{\hskip 0.3in} r @{\hskip 0.3in} c|} 
     \hline\rule{0pt}{2.2ex}
     \textbf{Layer} & \textbf{Units} &  \textbf{Output shape}  \\ [0.5ex] 
     \hline\rule{0pt}{2.2ex}Input tensor &-& 32  \\[1ex] 
     \hline \rule{0pt}{2.2ex}Dense & 256 &64 \\[.5ex] 
      \rule{0pt}{2.2ex}ReLU & - &64 \\[.5ex] 
\hline Dense & 256 & 3 \\[.5ex] 
ReLU & - & 3 \\[.5ex] 
\hline Dense & 1 & 3 \\[.5ex] 
Logistic & - & 3 \\[.5ex] 
    \hline
    \end{tabular}
    \vspace{0.1cm}
    \caption[Result: Reward prediction network]{\textbf{Reward prediction network.}}
    \label{tab:polnet3}
\end{table}

\textbf{The deep RL network} stacks the representation network and the policy network on top of each other. The representation network accepts the input and outputs the low-dimensional representation to the policy network that outputs the action scores.

\subsection{Training the representation and predictor networks} \label{subsec:policy} We collect a data set of $10$ thousand transitions by following a random policy in the two-room environment. For this data collection, each episode has a $50\%$ chance to have the goal location in the bottom room or on the left side of the top room (see the left and middle pictures in Fig. \ref{fig:Nw1b}). The reward predictor and the representation are trained in this manner for all experiments, including the lava gap and the four-room environment. Thus, we use a representation and reward predictor that have never seen lava.  For the experiments with smoothed rewards, the sparse reward associated with the observations in the data set is augmented by associating a new reward to the $64$ states leading to observations with a positive reward, according to Equation \ref{eq:rstar}, with a discount factor of $0.99$. Additionally, after the reward has been (potentially) smoothed in this way, observations associated with a positive reward are oversampled $10$ times to balance the data set, regardless of whether the reward has been augmented or not.




\section{Results and discussion}
\label{sec:rwrd}
In the experiments, we compare RL agents that learn their representations from scratch (Deep RL) to agents that preprocess their inputs with different representations. We compare our representation, trained on raw reward predictions -- with (Ours 1r) or without reward shaping (Ours 1r + Shaping) -- to our method trained on smoothed reward prediction, also with (Ours 64r) or without reward shaping (Ours 64r + Shaping). We use "64r" to denote that our method was trained with reward shaping and "1r" to denote that our method was trained without augmentation. As a baseline, we compare our representation to a reward-predictive representation from the literature, Successor Features (SFs). 
\subsection{Two-room environment}



\begin{figure}[h]
\centering
\begin{subfigure}[b]{0.75 \textwidth}
   \includegraphics[width=1\linewidth]{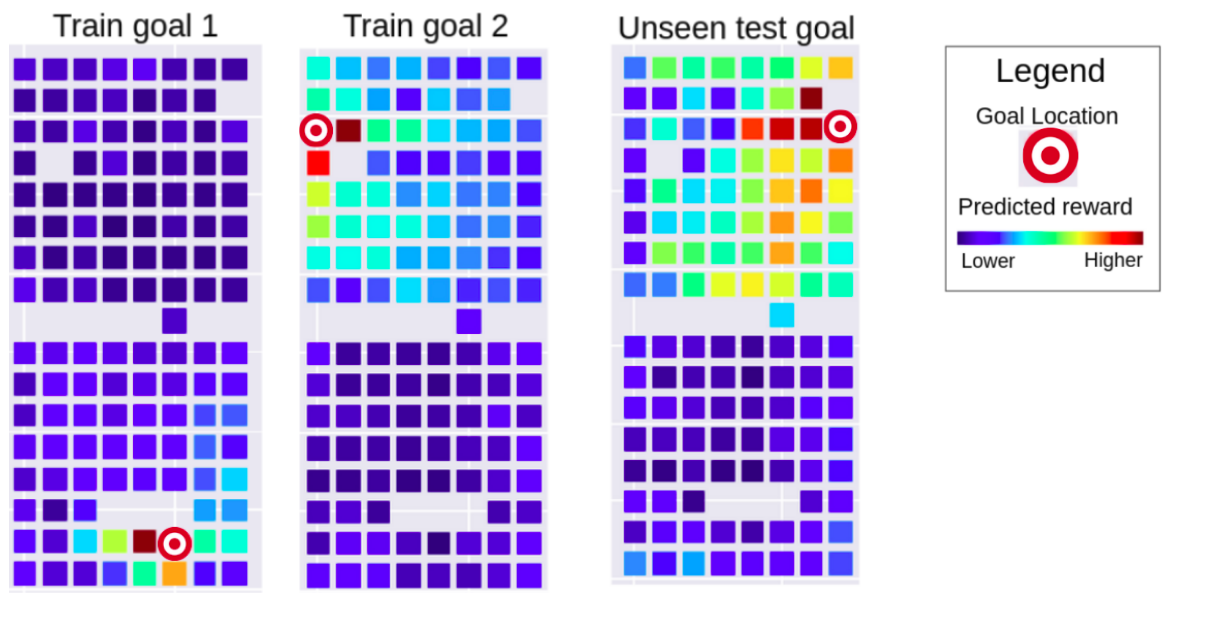}
   \caption{Raw reward prediction.}
   \label{fig:1r}
\end{subfigure}
\begin{subfigure}[b]{0.75\textwidth}
   \includegraphics[width=1\linewidth]{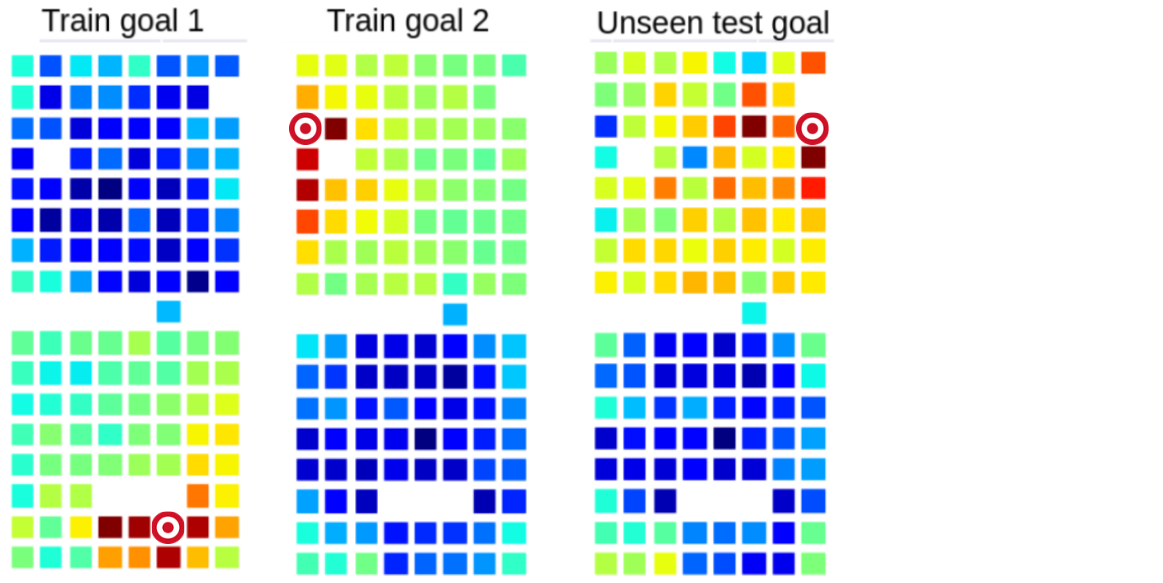}
   \caption{Smoothed reward prediction.}
   \label{fig:64r}
\end{subfigure}
  \caption[Result: Predicted rewards, two-room environment]{\textbf{Predicted rewards, two-room environment.} The predictor is trained on the setups shown on the left and in the middle, and tested for the setup on the right. The color becomes warmer for states where the reward is predicted to be higher.}
  \label{fig:rewardheatmap}
  \end{figure}

We start by visualizing the outputs of our reward predictor in the rooms, depending on the goal location, in Fig. \ref{fig:rewardheatmap}. Each square indicates the average predicted reward for transitioning to the corresponding tile in the room. 

The predicted reward spikes in a narrow region around the two goal locations
that were used to train the raw reward predictor (Fig. \ref{fig:1r}), but the area of states with high predicted rewards is
wider around the test goal. This difference is due to overfitting on the specific training paths that
were more frequently taken toward the respective goals, but this does not harm the generalization
capabilities of the network. The peakyness of the predictions disappears when the predictor is trained on the smoothed rewards (Fig. \ref{fig:64r}). However, higher predicted rewards in the corner of the other room appear. Both scenarios, raw and smoothed reward prediction, show promise for the application of reward shaping under our training scheme, as the agent would benefit from finding neighborhoods with higher values of predicted reward until it reaches the goal, instead of having to rely solely on a sparse reward that is only given when the agent lands exactly on the goal state.

\begin{figure*}[ht]
	\centering
	\begin{minipage}{.99\columnwidth}
		\centering
		\includegraphics[width=\textwidth]{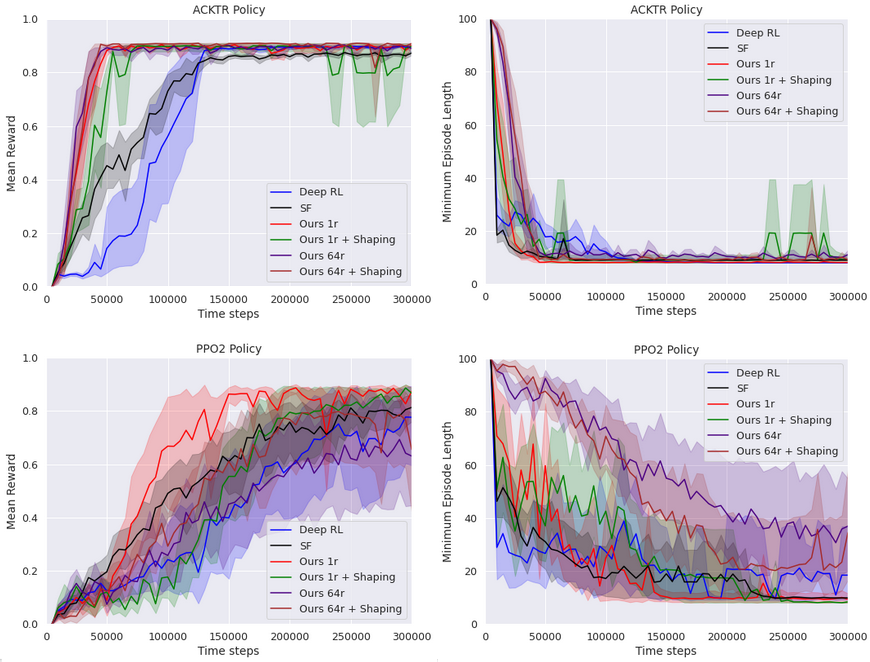}
	\end{minipage}
		\caption[Result: Two-room environment]{ \textbf{Two-room environment}. In these experiments, there are only two rooms and the agent must reach a goal that is always at the same location. The agent can traverse between the rooms and starts each episode between them, facing a random direction. The left side shows the mean of every agent's mean reward and the right side shows the mean of every agent's minimum episode length.}
		\label{allsingle}
\end{figure*}

In Figure \ref{allsingle}, we illustrate the variance of the mean reward (left side) and the variance of the optimal performance (right side) of the different methods, as a function of the time steps taken for training. We average over 10 runs and in each run we perform 10 test rollouts, so each point is the aggregate of 100 episodes in total.\footnote{Note that the standard deviation of the mean reward of all episodes, from all runs put together, is approximately $26\%$ higher than the standard deviation of the mean of the means or the mean of the mins.} The error bands indicate two standard deviations. This methodology of generating the plots also applies to Fig. \ref{fig:fromscratch}, Fig. \ref{graph:8k} and Fig. \ref{allfull2}.

The learning curves of both ACKTR and PPO2 get close to the highest achievable mean reward of $1$ the fastest using our representations. There no significant benefit from using smoothed reward shaping for ACKTR, and the raw reward shaping is in fact harmful in this case. For PPO2, the agent using our representation that is trained on raw reward predictions learns the fastest. Regular deep RL, where the representations are learned from scratch, is clearly outperformed by the variants that use reward-predictive representations. We believe that this is because RL agents can generally benefit from the input being preprocessed, as the computational overhead for learning the policy is reduced. This effect is enhanced when the preprocessing is good, which is the case for our reward-predictive representation: it abstracts away unnecessary information as it is trained to output features that indicate the distance between the agent and the goal, when the goal is in view.

The difference in aggregated mean rewards vs. aggregated minimum episode lengths can be explained due to systematically different behaviors. For example, an agent might have a weak long-term strategy of checking the different rooms, giving it poor average mean rewards, but a strong short-term tactic of taking the direct course to the goal when it sees it, giving it a good average minimum episode length. 

\newpage

\subsection{Lava gap environment}

  \subsubsection{Learning from scratch}

\begin{figure*}[ht]
	\centering
	\begin{minipage}{.37\columnwidth}
		\centering
		\includegraphics[width=\textwidth]{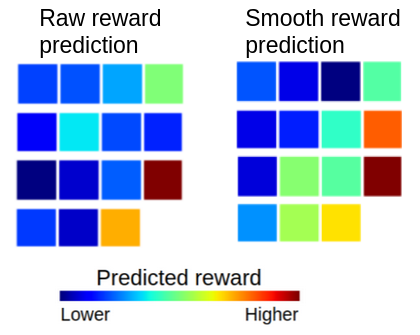}
	\end{minipage}
		\caption[Result: Predicted rewards, lava gap]{ \textbf{Predicted rewards, lava gap}. Average predicted reward per state in the lava gap environment. }
		\label{fig:gapheatmap}
\end{figure*}

\begin{figure*}[ht]
	\centering
	\begin{minipage}{.99\columnwidth}
		\centering
		\includegraphics[width=\textwidth]{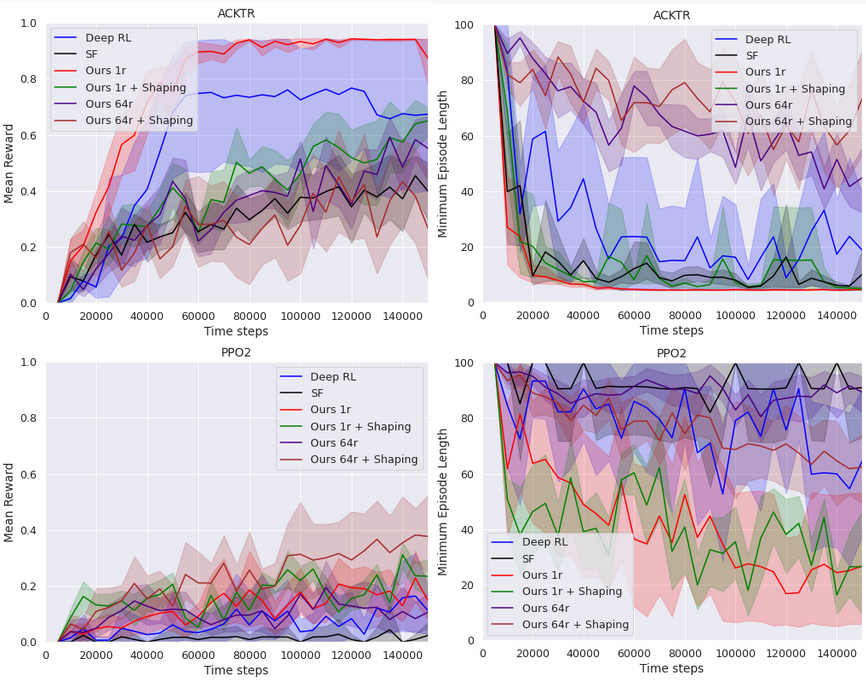}
	\end{minipage}
		\caption[Result: Lava gap experiment]{ \textbf{Lava gap experiment.} All policies are randomly initialized and learn to solve the lava gap environment from scratch. The representations in all methods except for Deep RL are learned on the training goals in the two-room environment (see Fig. \ref{fig:rewardheatmap}). The left side shows the mean of every agent's mean reward, and the right side shows the mean of every agent's minimum episode length.  }
		\label{fig:fromscratch}
\end{figure*}

The heatmaps of average predicted rewards are visualized in Fig. \ref{fig:gapheatmap}. The reward predictor was trained on the two-room environment. The tiles closest to the goal have the highest values, with a particularly smooth gradient toward the goal for the smoothed-reward predictor, which demonstrates that there is potential gain from transferring the prediction-based reward shaping between similar environments. The learning performance of the different methods can be seen in Fig. \ref{fig:fromscratch}. The decidedly fastest learning can be observed when the actor-critic method is combined with our representation, trained on raw reward predictions and without reward shaping. Regular deep RL is the second-best, but with a very large variance on the performance. Our reward shaping variations and the SFs are very close in performance, albeit significantly worse than the other two. The poor performance of reward shaping can be explained by the fact that there are very few states, which makes the reward shaping unnecessary in such a simple environment.  All the methods look more similar when PPO2 optimization is applied, with respect to the mean rewards, but our variant that is trained on smooth reward prediction and uses reward shaping reaches the highest average performance in the last iterations.

\subsubsection{Transfer learning}
 
\begin{figure*}[ht]
	\centering
	\begin{minipage}{.99\columnwidth}
		\centering
		\includegraphics[width=\textwidth]{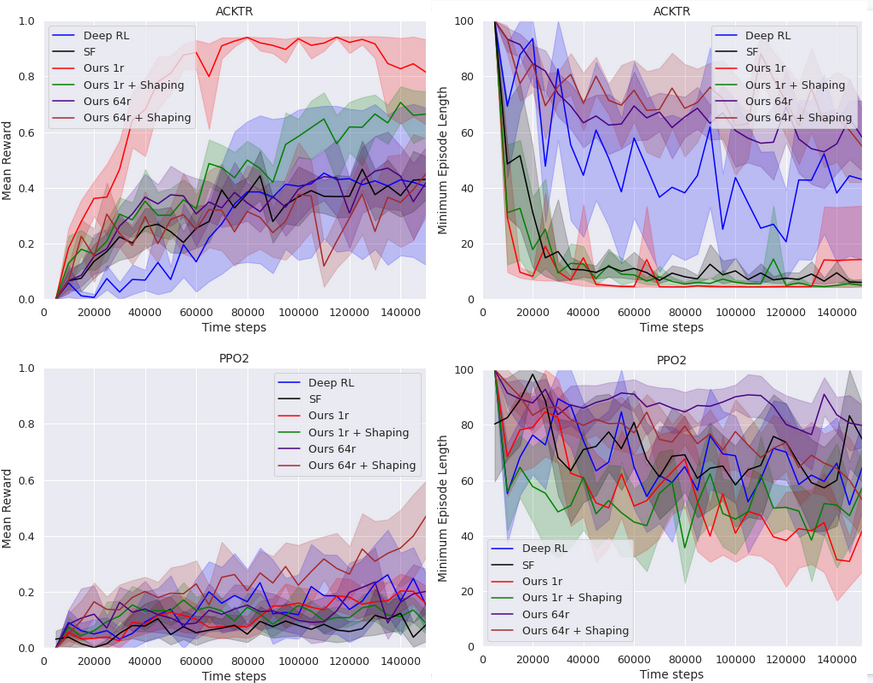}
	\end{minipage}
		\caption[Result: Re-learning experiment]{ \textbf{Re-learning experiment} The different methods are trained for eight thousand training steps on the two-room environment before being trained on the lava gap environment. The curves show the mean reward on the lava gap environment. The left side shows the mean of every agent's mean reward, and the right side shows the mean of every agent's minimum episode length.}
		\label{graph:8k}
\end{figure*}

 To investigate how the methods compare for adapting to new environments, we trained the policies for 8000  steps on the two-room environment before learning to solve the lava gap environment, see Fig. \ref{graph:8k}. Our method, without reward shaping, facilitates the fastest learning for ACKTR in this case. Deep RL is the most severely affected by this change, which is probably due to the method learning a reward-maximizing representation in one environment that does not transfer well to another environment. Every PPO2 variation looks bad for this scenario, but the smooth-reward prediction representation with reward shaping has the highest mean reward and our raw-reward prediction representation has the lowest average minimum episode length. 

\begin{figure}[ht]
\centering
\begin{subfigure}[b]{0.47\textwidth}
   \includegraphics[width=1\linewidth]{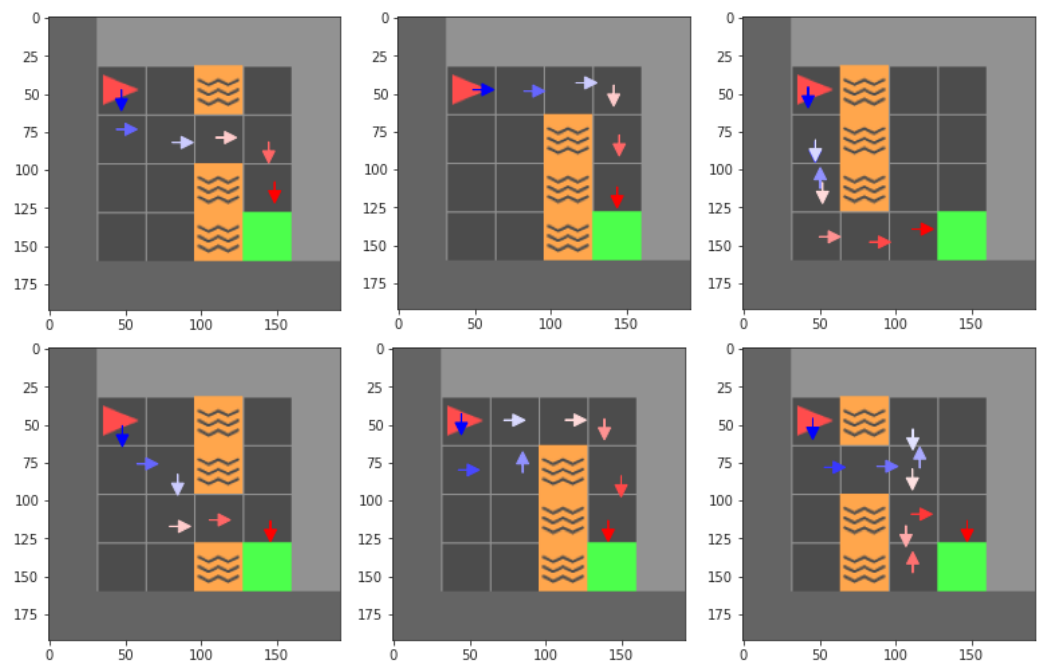}
   \caption{Six successful episodes.}
   \label{fig:Ng1} 
\end{subfigure}
\begin{subfigure}[b]{0.47\textwidth}
   \includegraphics[width=1\linewidth]{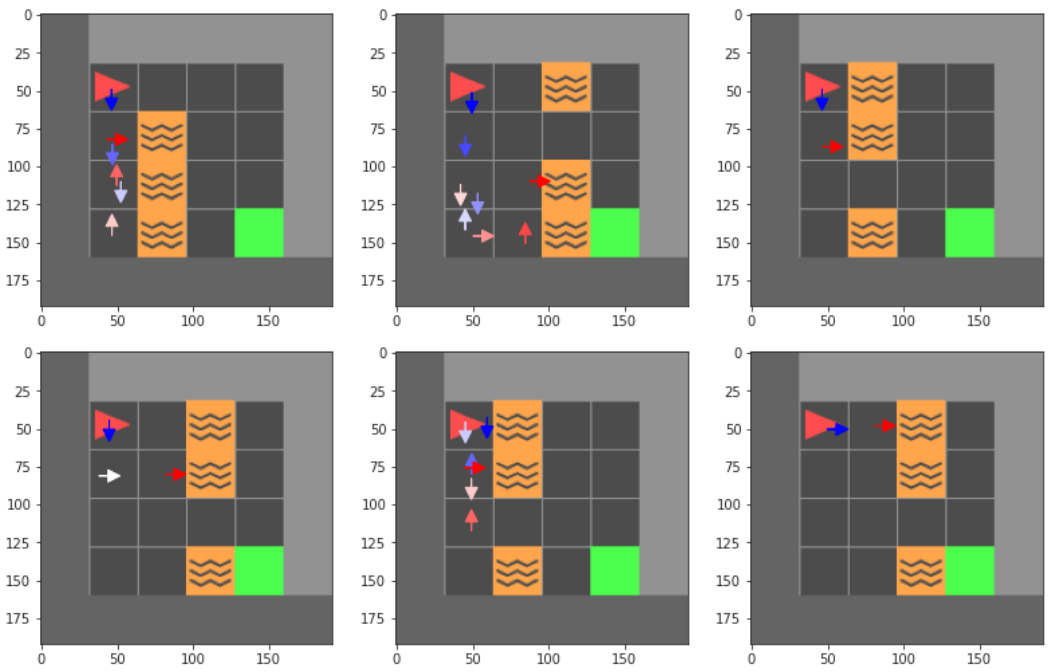}
   \caption{Six failed episodes.}
   \label{fig:Ng2}
\end{subfigure}
\caption[Result: Lava gap trajectories]{\textbf{Lava gap trajectories.} A visualization of six successful and six failed trajectories by an agent that was trained up to approximately $75\%$ success rate using our representation. The color gradient goes from the first actions taken in the episode in blue to the last actions in red. Note that rotations in-place are not visualized, but only transitions between tiles. }
\label{lavatraj}
\end{figure}

We visualize trajectories of an agent that is trained on our representation (Ours 1r) as it traverses the lava gaps environment (Fig. \ref{lavatraj}). For inspection of cases where it fails, we choose an agent that has been trained for $50$ thousand time steps only and has around $0.75$ mean reward.

\subsection{Four-room environment}

\begin{figure*}[ht]
	\centering
	\begin{minipage}{.97\columnwidth}
		\centering
		\includegraphics[width=\textwidth]{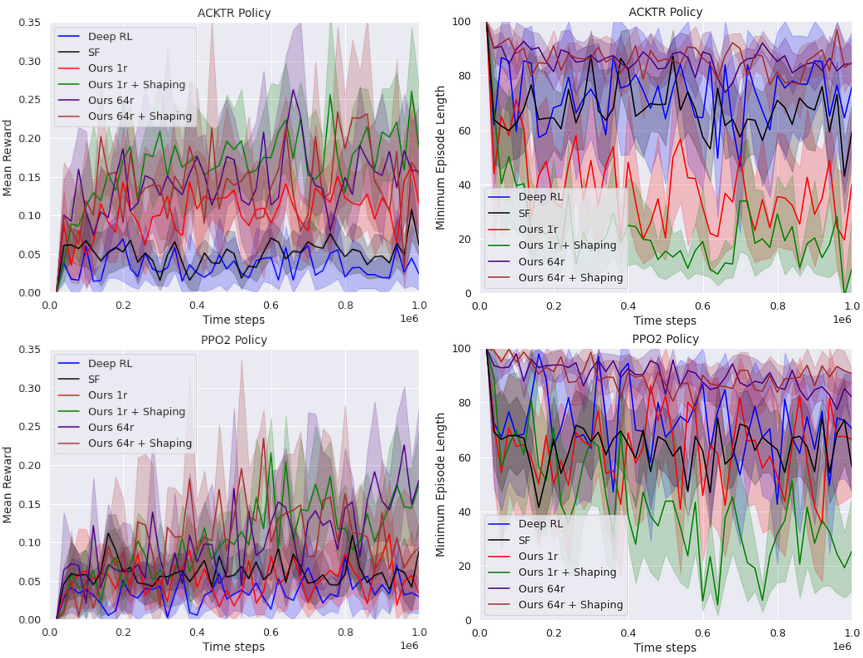}
	\end{minipage}
		\caption[Result: Full four-room environment]{ \textbf{Full four-room environment}. The agent and goal are placed at random locations at the start of each episode. The left side shows the mean of every agent's mean reward, and the right side shows the mean of every agent's minimum episode length.}
		\label{allfull2}
\end{figure*}

In our final comparison, we add two additional rooms to the two-room environment and randomize both the goal location and the starting position of the agent, with the results shown in Fig. \ref{allfull2}.  Looking at the minimum episode lengths, for the ACKTR learner, our raw-reward prediction representation with reward shaping performs best and the one without reward shaping comes in second.  There is little discernible difference between the performance of SFs and Deep RL, but they both perform significantly worse than our methods. The scale of the mean reward is a great deal lower than in the previous experiments, since the average distance between the starting tile of the agent and the goal is much larger than in the previous two environments. 

For this scenario, all the methods look similarly bad for the PPO2 policy, except for our raw-reward representations, with reward shaping, which has the lowest minimum episode length. The big advantage of reward shaping in this environment compared to the two-room environment can be explained by the increased complexity, making the reward shaping more helpful in guiding the agent's search. In the previous experiments, the agent and goal locations start at fixed locations, allowing the agents to solve it by rote memorization.  The reward shaping function calculated by the raw-reward predictor fares significantly better in this situation. We hypothesize that this is due to the smoothed-reward predictor distracting the agent by pushing it to corners, as the visualization in Fig. \ref{fig:64r} would suggest. 

The reward shaping given by the raw-reward predictor is more discriminative, as we see in Fig. \ref{fig:1r}. The agent receives a positive reward as soon as the goal reaches its point of view, which is any location up to six tiles in front of it and no further than 3 tiles away from it to the left or to the right. This allows the reward shaping function to guide the agent directly to the goal, assuming that they are in the same room and that there is no wall obstructing the agent's field of vision.

\begin{figure}[h]
\centering
\begin{subfigure}[b]{0.75\textwidth}
   \includegraphics[width=1\linewidth]{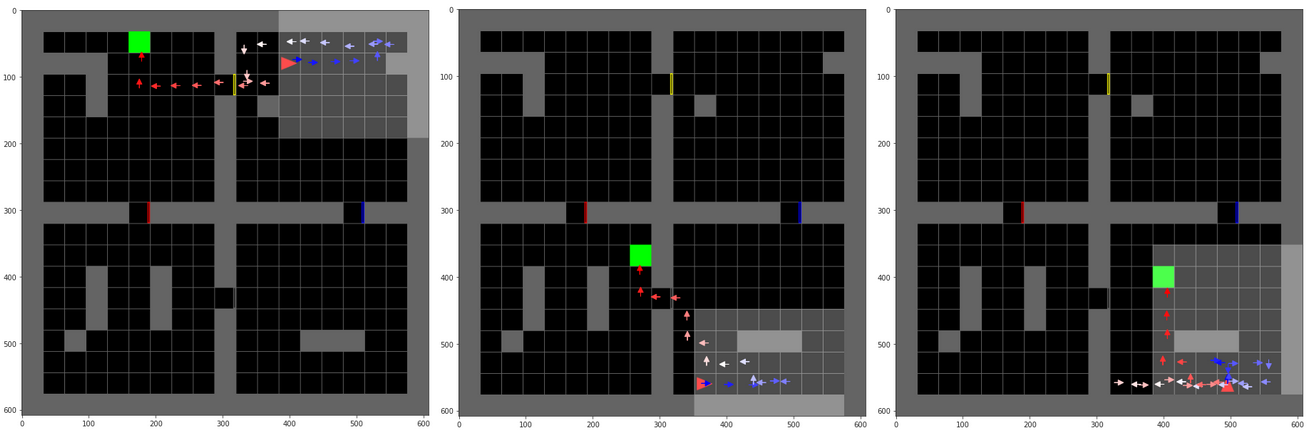}
   \caption{Three successful trajectories}
   \label{fig:Ng3} 
\end{subfigure}
\begin{subfigure}[b]{0.75\textwidth}
   \includegraphics[width=1\linewidth]{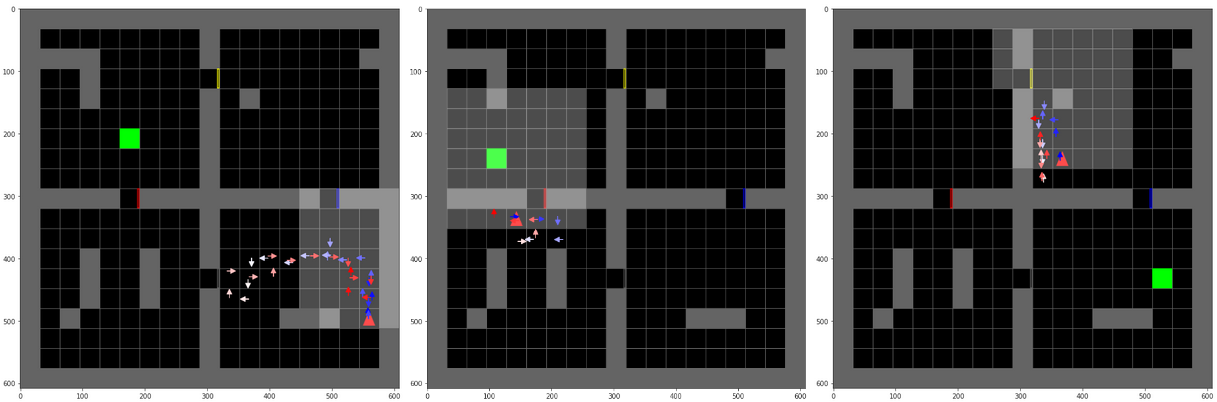}
   \caption{Three failed trajectories}
   \label{fig:Ng4}
\end{subfigure}
\caption[Result: Four-room trajectories]{\textbf{Four-room trajectories.} The agent was trained on our representations with Actor Critic using Kronecker-Factored Trust Region for a million time steps. }
\label{viz4rooms}
\end{figure}

Three successful and three failed trajectories of an ACKTR agent that has been trained, using our representation (Ours 1r) for a million time steps are visualized in Fig. \ref{viz4rooms}.  We can see undesirable behavior in both the successful and the failed trajectories, that the agent wastes effort re-visiting tiles it has already been to.

\section{Conclusion} \label{sec:disc}

 Processing high-dimensional inputs for reinforcement learning (RL) agents remains a difficult problem, especially if the agent must rely on a sparse reward signal to guide its representation learning. In this work, we put forward a method to help alleviate this problem with a method of learning representations that preprocesses visual inputs for RL methods. Our contributions are (i)~a reward-predictive representation that is trained  simultaneously with a reward predictor and (ii)~a reward shaping technique using this trained predictor. The predictor learns to approximate either the raw reward signal or a smoothed version of it, and it is used for reward shaping by encouraging the agent to transition to states with higher predicted rewards.
 
 We used a view of the goal as a second input for the methods in our experiments, but this is in principle not necessary, as moving toward the green tile as it becomes visible is sufficient. Removing the goal input might encourage the agents to learn policies that scan all the rooms faster until the goal reaches its field of vision.
 
 We have shown the usefulness of our representation and our reward shaping scheme in a series of gridworld experiments, where the agent receives a high-dimensional observation of its goal as an input along with an observation of its immediate surroundings. Preprocessing the input using this representation speeds up the training of two out-of-the-box RL methods, Actor Critic using Kronecker-Factored Trust Region and Proximal Policy Optimization, compared to having these methods learn the representations from scratch. In our most complicated experiment, combining our representation with our reward shaping technique is shown to perform significantly better than the vanilla RL methods, which hints at its potential for success, especially in more complex RL scenarios.

\clearpage

\chapter{Comparison of our three methods}
We compare in this chapter the algorithms that we developed over the course of the PhD work: the gradient-based ICA representation (GrICA) from Chapter \ref{chap:grica}, the latent representation prediction network (LARP) from Chapter \ref{chap:larp} and the reward-prediction representation from Chapter \ref{chap:rewpred} (RewPred). We use the representations calculated by these methods to preprocess inputs for deep reinforcement learning (Deep RL) agents on four different tasks: a cart-pole balancing environment, the two-room and four-room goal-finding environments from the previous chapter and an obstacle avoidance task. The implementation of our methods is the same as described in the experimental section of their corresponding chapter, unless specified otherwise. 

In Section \ref{compmethods} we introduce the new visual cart-pole environment and discuss the network architectures. In Section \ref{compresults} we describe and discuss the results of the experiments. 

\section{Introduction}

As each method was introduced over the course of the PhD work, it was only compared to other similar methods in the literature. We now take the opportunity to compare them against each other in different environments. The cart-pole environment and the obstacle avoidance environment have the commonality that they both require only reactive short-term planning to maximize the reward, but the goal-finding environments require more long-term planning to find and reach the goal. 

Every one of our representations is used for preprocessing the visual observations of each environment for RL algorithms. The RL agents are the same ones used in the previous chapter: ACKTR and PPO2, with the default parameters from the Stable Baselines package. As we were not able to solve the cart-pole environment using ACKTR, only PPO2 is used for that task. We compare the method from Chapter \ref{chap:larp}, LARP, in this manner, even though it is not trained for preprocessing inputs for model-free agents but is developed to be paired with a predictor for predictive state representation rollouts
\section{Methods}
\label{compmethods}

\subsection{Visual cart-pole}

We start by comparing our methods as visual preprocessing for a PPO2 policy on a variant of the classical cart-pole balancing task. A pole is attached to a cart that moves along a frictionless track. The goal is to keep the pole upright for as long as possible, but the agent must push the cart to the left or to the right at every time step -- choosing to remain in place is not an option. The original environment supplies four scalar variables to the agent: the cart position and velocity and the pole angle and angular velocity. The agent receives a positive reward of $+1$ for each transition. 

\begin{figure}[h]
    \centering
    \includegraphics[width=0.98\linewidth]{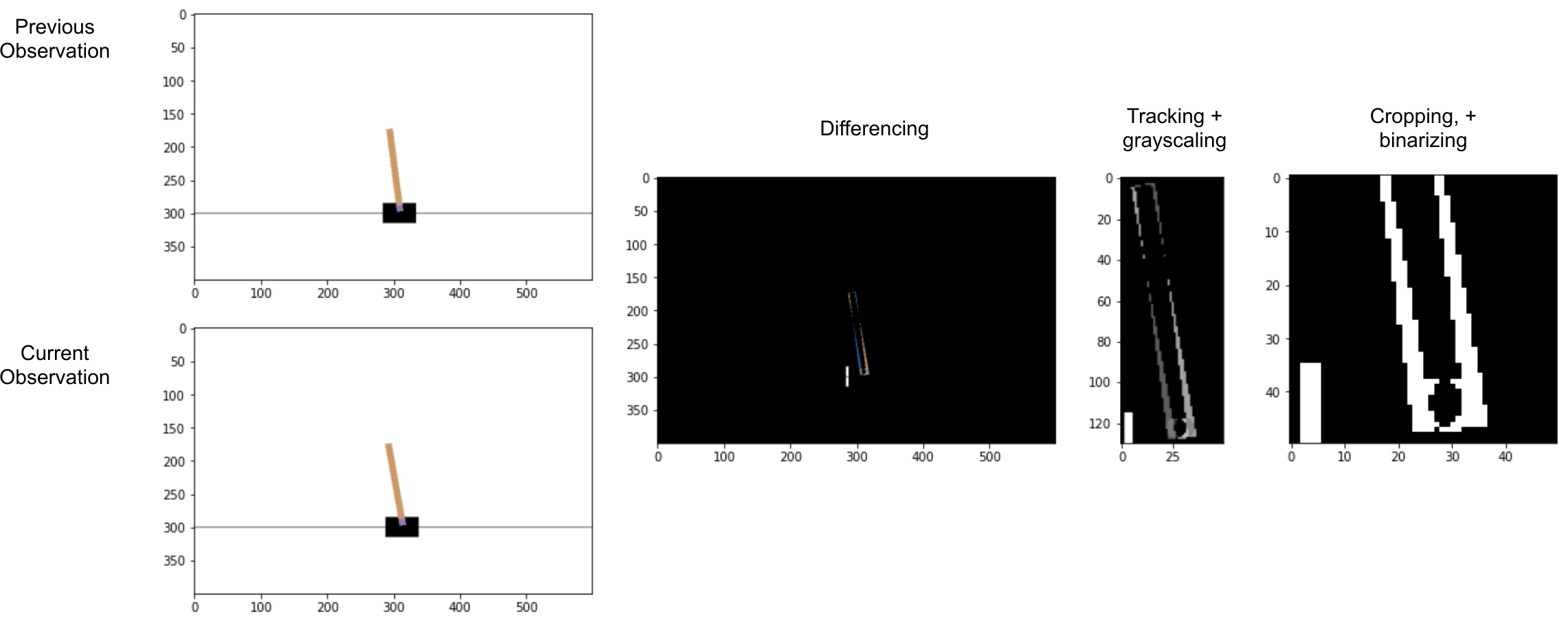}
    \caption[Example: Visual cart-pole]{\textbf{cart-pole processing pipeline.} We replace the original 4-dimensional state space of the OpenAI gym cart-pole environment with the output of its rendering function. The previous frame is subtracted from the current frame, the non-zero pixel values are centered, cropped and the final image is converted to binary.}
    \label{fig:cart-poleillustratio21}
\end{figure}

Instead of using these scalars, we adapt this task to our paradigm by using a visualization of the environment, see Fig. \ref{fig:cart-poleillustratio21}. To begin with, we extract the rendering of the environment that is meant for debugging and visualization purposes. As we can approximate the cart position and velocity and the pole angle and angular velocity using two adjacent frames, we use the current and previous observations. To simplify the input for the agent, we take the difference of the two frames, crop the image around the cart and binarize it. The only information that is lost in this process, compared to the original environment, is the position of the cart, which is not so important for solving the task.  In addition to giving the agent a positive reward of $+1$ for every time step it keeps the pole balanced, we also supply it with a negative reward of $-1$ when the pole falls down. 


Our methods are trained on 5000 transitions that are collected from a random policy. Even though this is not a sparse-reward environment, we calculated new rewards for the RewPred network according to Equation \ref{eq:rstar} with a discount factor of $\gamma = 0.9$ and a maximum horizon of $M=6$. This has the effect that every time step is associated with a reward of $+1$, except for the six transitions leading up to the end of the episode, which have the rewards $-(0.9^5), -(0.9^4), -(0.9^3), -(0.9^2), -(0.9)$ and $-(1.0)$.

\subsection{Room environments}

We compare the performance of PPO2 and ACKTR policies as they use our methods for preprocessing on the two-room and four-room goal tasks from the previous chapter, see Section \ref{tworoomimplementation}. The data gathering for the representations used for preprocessing for the RL agents is unchanged from the previous chapter.

\subsection{Obstacle avoidance}

\begin{figure}[h]
    \centering
    \includegraphics[width=0.98\linewidth]{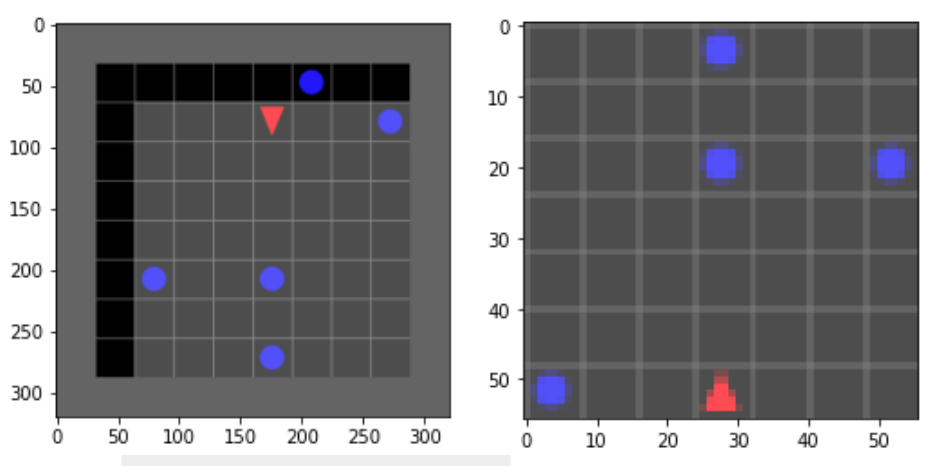}
    \caption[Example: Obstacle avoidance environment]{\textbf{Obstacle avoidance environment.} The agent is rewarded for maximizing the distance between itself and the closest circle. The left figure displays the full world state and the right figure displays the corresponding observation that the agent receives. }
    \label{fig:obstacleavoidance_illustration}
\end{figure}

In this gridworld environment, the agent is placed in a room filled with circular objects that move randomly in each step (Fig. \ref{fig:obstacleavoidance_illustration}).

The agent receives visual inputs of dimension  $56\times56\times3$ and must stay as far away from the objects as possible, as it receives a reward that is equal to the Euclidean distance between itself and the closest circle. The episode ends with a reward of $0$ after $100$ time steps or when the agent collides with a circle, which gives a reward of $-1$. The discount factor for the environment is 0.9. 

\subsection{Architectures}
All representations and policies have the same architecture as the in the previous chapter, except that the output of the representation has been lowered to a 16-dimensional vector for lower computational times, see Table \ref{tab:repnet} and Table \ref{tab:polnet1}. The RL models have the same architecture except that the representation and policy modules are trained end-to-end. The prediction module of the LARP network is the same as in Chapter \ref{chap:larp}, see Table \ref{table:predictornet}. The mutual information neural estimator network used for training GrICA is the same as in Chapter \ref{chap:grica}, see Section \ref{icamethod}.

\section{Results and discussion}
\label{compresults}

\subsection{Visual cart-pole}

\begin{figure}[htb]
    \centering
    \includegraphics[width=0.69 \linewidth]{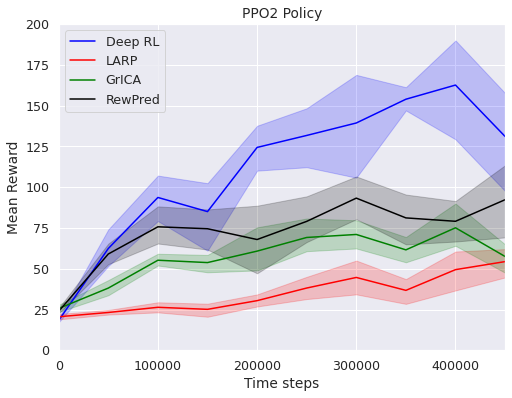}
    \caption[Results: Visual cart-pole]{\textbf{cart-pole results.} Each point is the aggregate of 5 different policies that are trained from scratch and tested for 20 episodes, each. }
    \label{fig:cart-poleillustration21}
\end{figure}

The results from the visual cart-pole experiment can be seen in Fig. \ref{fig:cart-poleillustration21}. The deep RL agent achieves the highest mean reward by a significant margin, probably because the environment requires the agent to take quick actions in response to the fast-changing environment.

The learning curve is the worst when the agent is trained on the LARP representations, which is to be expected as it is learned to be paired with a representation predictor and graph search, which it is not used for this scenario.

The RewPred and GrICA learning curves look similarly good, with the RewPred representation achieving a slightly higher average reward. Both representations should theoretically be useful here: learning a representation that is predictive of how close the pole is from falling down (RewPred) ought to guide the agent's actions away from states where the pole is about to fall down, and recovering the three statistically independent latent variables generating the environment (GrICA), which would be a perfect compression of the environment. In this scenario, however, they do not beat the representation learning of the Deep RL algorithm.

\subsection{Room goal-finding}
The results for training on the smooth-reward prediction RewPred\footnote{Denoted "Ours 64r" in the previous chapter.}  and Deep RL representations are kept in the plots from the previous chapter, and we have added the results for the GrICA and the LARP representations.
\subsubsection{Two-room environment}

\begin{figure}[h]
    \centering
    \includegraphics[width=0.98\linewidth]{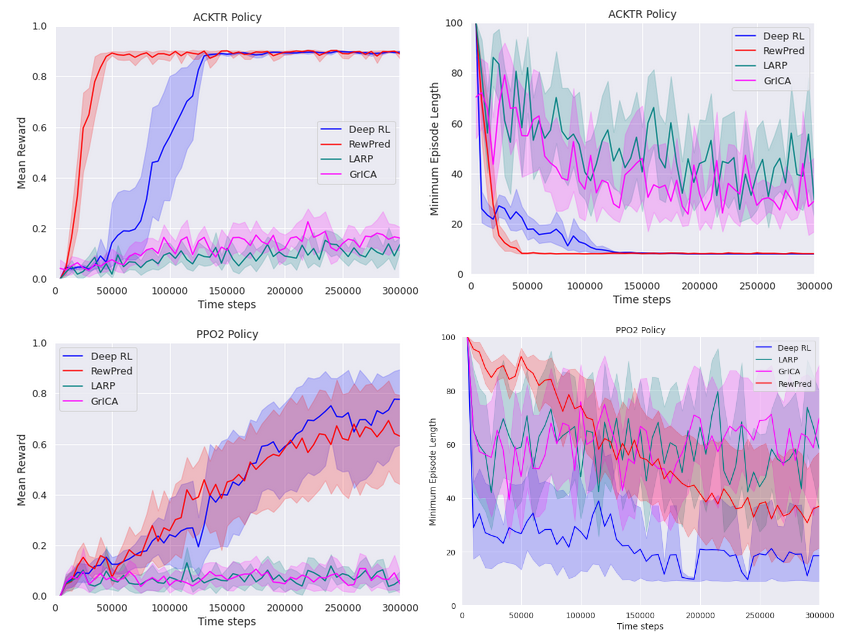}
    \caption[Results: Gridworld comparison, single goal location]{\textbf{Gridworld comparison: two-room goal-finding results}. The agent starts in between two rooms and get a reward from reaching a static goal location that is one of two rooms. Each point in the learning curve is aggregated from 10 initializations of policies that run 10 test episodes each.  }
    \label{fig:tworoomcomp}
\end{figure}

The results for the goal-finding task, when the agent starts facing a random direction between two rooms and must locate a static goal, can be seen in Fig. \ref{fig:tworoomcomp}. We display here the version of the reward-predictive RewPred representation from the previous chapter that is trained on smoothed rewards.

For the ACKTR policy,  both the LARP and the GrICA learning curves\footnote{See the previous chapter for the RewPred results} are considerably lower than those of the other two, with the GrICA representation again looking slightly better out of the two, both when the mean reward and the minimum episode lengths are considered. This is probably also due to the fact that the advantage of the LARP representation disappears when the predictor that it is trained with is discarded and the representation is being used in a model-free setting. For the PPO2 policy, neither the policies that train on the LARP nor the GrICA representations show signs of learning to solve the environment.

\subsubsection{Four-room environment}

\begin{figure}[h]
    \centering
    \includegraphics[width=0.98\linewidth]{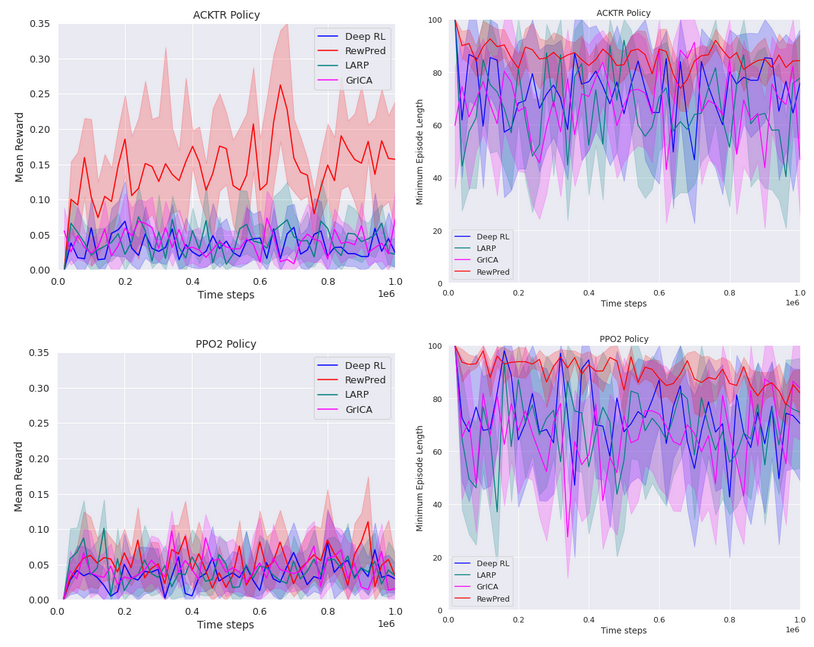}
    \caption[Results: Gridworld comparison, four-room goal-finding]{\textbf{Four-room goal-finding results.} The agent starts at a random location get a reward from reaching a dynamic goal location that is in one of four rooms. Each point in the learning curve is aggregated from 10 initializations of policies that run 10 test episodes each.}
    \label{fig:fourroomcomp}
\end{figure}

In this experiment, no combination of RL agent and representation shows progress toward solving the environment, except for RewPred paired with ACKTR. Both the LARP and GrICA variants barely display learning in the previous experiment, but all signs of progress disappear when the environment is sufficiently complex. The training results can be seen in Fig. \ref{fig:fourroomcomp}.

\subsection{Obstacle avoidance}

\begin{figure}[h]
    \centering
    \includegraphics[width=0.98\linewidth]{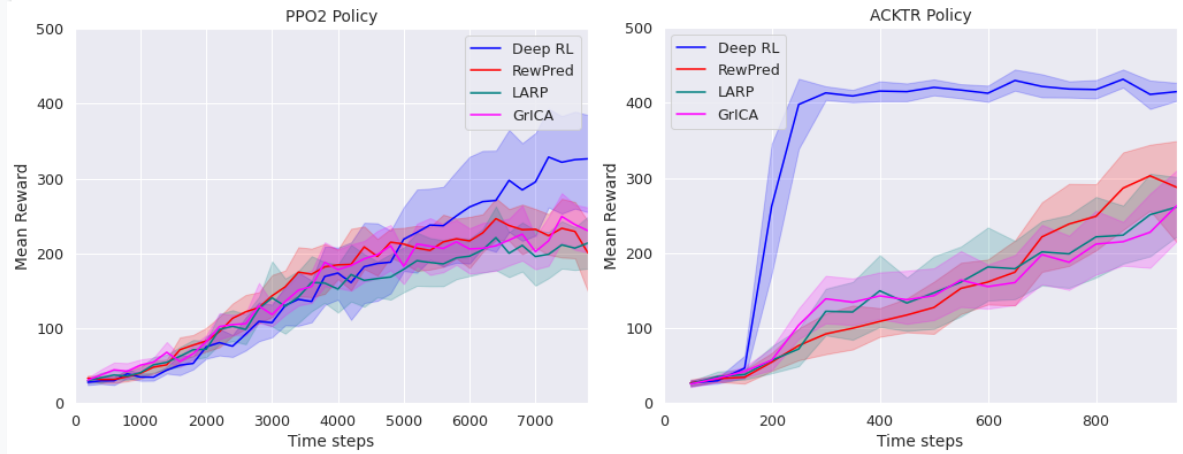}
    \caption[Results: Obstacle avoidance]{\textbf{Obstacle avoidance results.} The agent gets a reward for keeping its distance from objects that move randomly. Each point in the learning curve is aggregated from 10 initializations of policies that run 20 test episodes each.}
    \label{fig:obstacleavoidance}
\end{figure}

The algorithm's learning curves for the obstacle avoidance experiment can be seen in Fig. \ref{fig:obstacleavoidance}. For both PPO2 and ACKTR, we observe that preprocessing with our methods is not beneficial as the environment is solved the fastest using deep RL, and by a significant margin for ACKTR. This shows that our method is outperformed by regular deep RL for environments where short-term, reactive policies are needed -- in contrast to the goal-finding tasks, where the agent needs to execute a long-term plan.

\section{Conclusion}

In this chapter, we investigated how the three methods, that we have developed in this PhD work, compare when they are used for preprocessing visual inputs for RL agents. The RL agents are tested in four different environments, two of which require short-term decision-making and the other two require long-term decision-making for success.  Our representation that was developed in Chapter \ref{chap:grica}, GrICA, does not facilitate learning for any of the environments. The same holds true for the Chapter \ref{chap:larp} representation, LARP, although that one was not developed for preprocessing inputs to RL agents but rather to be used in conjunction with a prediction function for planning in a latent representation space. Our reward-predictive representation from Chapter \ref{chap:rewpred} is shown to speed up learning for a deep RL agent on the long-term planning tasks.

\chapter{Summary and conclusion}



Just as the complexity of the tasks that deep reinforcement learning (RL) is being applied to increases, so does the apparent  number of problems the field has. The goal of this research was to discover useful representations that can be used to alleviate two of modern deep RL's problems, namely, those of training instability and data inefficiency. Over the course of the PhD work, we developed three representation learning methods and investigated their suitability for fulfilling this goal:

\begin{itemize}
    \setlength{\itemindent}{1.55cm}

    \item[GrICA:] A gradient-based ICA method for learning statistically independent features (Chapter \ref{chap:grica}). We estimate the mutual information between one output component of an encoder and all the others using a neural network. In a push-pull fashion, the mutual information estimator and the encoder are trained, resulting in a system that outputs statistically independent features of the input. The output of GrICA is compare to the output of FastICA, an established ICA problem-solving method in the literature, for noisy blind signal separation.
    \setlength{\itemindent}{1.45cm}

    \item[LARP:] This system, which we call "Latent Representation Predictor" (Chapter \ref{chap:larp}), learns a transition model of the environment. We evaluate our system by combining it with graph search to manipulate toy objects to match a given viewpoint. Our algorithm learns a state representation jointly with a one-step lookahead predictor. We discuss and compare three different constraints that can be placed on the system to prevent the solution from collapsing to a constant function. Our approach outperforms deep RL in a low-data regime on the viewpoint-matching task.
    \setlength{\itemindent}{1.93cm}

    \item[RewPred:] A reward-predictive representation, that is learned along with a jointly learned reward predictor (Chapter \ref{chap:rewpred}). The reward predictor is employed for reward shaping: The agent is rewarded for moving from states of low predicted rewards to states of higher predicted rewards. The method is tested in several grid world environments where the agent must reach a goal. The learning of deep RL agents is sped up when their inputs are preprocessed by the RewPred representation in our experiments and the reward shaping is helpful when the environment is sufficiently complex.
\end{itemize}
 Not every unsupervised learning technique yields a state representation that is useful in the context of RL, as seen by the results across the board using our GrICA algorithm. However, based on our results, we have found that the training of deep RL methods can be augmented by learning appropriate representations in a self-supervised manner in environments where the agent must carry out long-term planning to reach a single goal state.

An important limitation of this work is the high cost of computational resources required for carrying out RL research. To illustrate, reproducing DeepMind's 2017 Go paper \citep{silver2017mastering} is estimated to cost 35 million dollars using Google's cloud computing service \citep{dan2018how} -- a budget currently unavailable to PhD students.

Reproducing their results only involves maximizing the performance of a single model, which does not take into account the work behind finding the final model. RL methods have many potential settings to tune, and running hyperparameter searches is costly. For this reason, we concentrated only on using the default parameters of RL model implementations as offered by the Stable Baselines package.

To make matters worse, the performance of RL techniques can be highly sensitive to the hyperparameter settings, and bad luck can cause the researcher to dismiss a method if unfortunate values are initially chosen.  A larger computational budget allows researchers to try more combinations of parameters when new methods are tested, lessening the risk of promising approaches being prematurely dismissed due to bad luck.

The methods were also tested for preprocessing the visual inputs of deep RL algorithms in environments where there is a set of state that the agent must \textit{avoid}, namely, preventing a pole from falling off a cart or preventing the agent from colliding with randomly moving objects. Preprocessing the inputs using our methods is not beneficial in these cases, potentially indicating that allowing deep RL algorithms to learn the state representations is preferable for tasks that require immediate reactions to quickly changing environments, particularly where inaction leads to an undesirable outcome.

Although our gradient-based ICA was not found advantageous in our goal of augmenting deep RL training, it is successful in recovering independent, noisy sources just as well as FastICA. A promising avenue of research is to take advantage of the flexibility offered by our method. Our algorithm can be paired with any differentiable function, such as convolutional neural networks (convnets), to tackle more difficult problems, for example nonlinear ICA. Even though general nonlinear ICA problems are ill-posed, regularization can be applied to make them well-posed. This can in principle be done using our method via the design of the neural network.

One aspect of representation learning that has become popular in recent years, but was not investigated in this PhD work, is the application of memory modules in neural networks. The role of memory can easily be integrated into our methods by designing our function approximators as artificial recurrent neural networks, for example by combining long short-term memory networks with convnets.


Although the problem of learning representations in the context of RL remains unsolved, the implication of our work is that taking advantage of the rich unsupervised source of supervision that is hidden in RL environments can lead to data-efficient, stable algorithms that are resilient to changes in the environment.

\begin{figure}[h]
    \centering
    \includegraphics[width=0.5\linewidth]{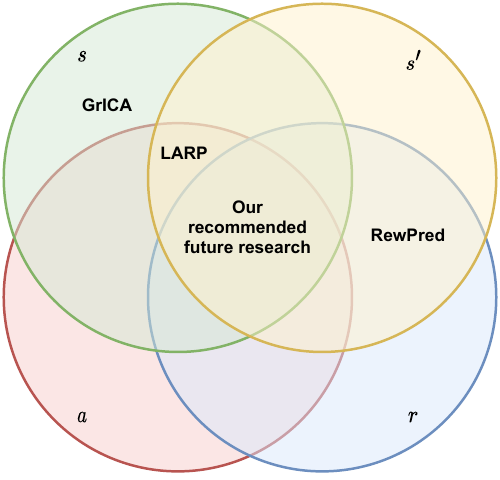}
    \caption[Illustration: Supervision source Venn diagram]{\textbf{Supervision source Venn diagram.} An illustration of the components of the RL transition tuples $(s, a, r, s') = (\text{state},\text{ action},\text{ reward},\text{ next state})$ that are used by our methods: GrICA (Chapter \ref{chap:grica}), LARP (Chapter \ref{chap:larp}), RewPred (Chapter \ref{chap:rewpred}). These tuples consist of all the information involved in a single transition in an RL environment. The intersection of every component corresponds to the theoretical extension of our work, combining elements of LARP and RewPred.}
    \label{fig:venn}
\end{figure}

A straightforward way of extending our work would be to take advantage of the full data that is given in every transition in an RL environment: the current state $s$, the action $a$ taken in the state $s$, the resulting next state $s'$ and the reward $r$ given to the agent by the environment. It would be an interesting future line of work  to do this by simply combining the loss function of LARP, which trains on $(s, a, s')$ tuples, and RewPred, which trains on $(r, s')$ tuples. This new algorithm would train the representation to be simultaneously predictable for a representation predictor, but also to be useful for a reward predictor. See Fig. \ref{fig:venn} for a Venn diagram that summarizes the sources of supervision that our proposed extension would use in addition to the sources of supervision that are used by our methods.

This dissertation adds to a rapidly growing body of knowledge that sits at the intersection of deep learning, reinforcement learning and representation learning. We contribute three novel methods of learning state representations to the literature and experimentally evaluate their effectiveness in the context of reinforcement learning. A long road to artificial intelligence that matches our general and efficient problem-solving capability still lies ahead of us.


\clearpage

\pdfbookmark[0]{Bibliography}{bibliography}
\bibliography{main}

\begin{thebibliography}{}

\bibitem[Abadi et~al., 2016]{abadi2016tensorflow}
Abadi, M., Barham, P., Chen, J., Chen, Z., Davis, A., Dean, J., Devin, M.,
  Ghemawat, S., Irving, G., Isard, M., et~al. (2016).
\newblock Tensorflow: A system for large-scale machine learning.
\newblock In {\em 12th $\{$USENIX$\}$ Symposium on Operating Systems Design and
  Implementation ($\{$OSDI$\}$ 16)}, pages 265--283.

\bibitem[Alom et~al., 2019]{alom2019state}
Alom, M.~Z., Taha, T.~M., Yakopcic, C., Westberg, S., Sidike, P., Nasrin,
  M.~S., Hasan, M., Van~Essen, B.~C., Awwal, A.~A., and Asari, V.~K. (2019).
\newblock A state-of-the-art survey on deep learning theory and architectures.
\newblock {\em Electronics}, 8(3):292.

\bibitem[Amari et~al., 1996]{amari1996new}
Amari, S.-i., Cichocki, A., and Yang, H.~H. (1996).
\newblock A new learning algorithm for blind signal separation.
\newblock In {\em Advances in neural information processing systems}, pages
  757--763.

\bibitem[Amodei and Clark, 2016]{clark2016faulty}
Amodei, D. and Clark, J. (2016).
\newblock Faulty reward functions in the wild.
\newblock \url{https://blog.openai.com/faulty-reward-functions/, 2016}.

\bibitem[Arora et~al., 2018]{arora2018analysis}
Arora, S., Hu, W., and Kothari, P.~K. (2018).
\newblock An analysis of the t-sne algorithm for data visualization.
\newblock In {\em Conference on Learning Theory}, pages 1455--1462. PMLR.

\bibitem[Baldi and Hornik, 1989]{baldi1989neural}
Baldi, P. and Hornik, K. (1989).
\newblock Neural networks and principal component analysis: Learning from
  examples without local minima.
\newblock {\em Neural networks}, 2(1):53--58.

\bibitem[Barreto et~al., 2016]{barreto2016successor}
Barreto, A., Dabney, W., Munos, R., Hunt, J.~J., Schaul, T., Van~Hasselt, H.,
  and Silver, D. (2016).
\newblock Successor features for transfer in reinforcement learning.
\newblock {\em arXiv preprint arXiv:1606.05312}.

\bibitem[Belghazi et~al., 2018]{belghazi2018mine}
Belghazi, M.~I., Baratin, A., Rajeswar, S., Ozair, S., Bengio, Y., Courville,
  A., and Hjelm, R.~D. (2018).
\newblock {MINE: Mutual Information Neural Estimation}.
\newblock {\em arXiv preprint arXiv:1801.04062}.

\bibitem[Belkin and Niyogi, 2003]{belkin2003laplacian}
Belkin, M. and Niyogi, P. (2003).
\newblock Laplacian eigenmaps for dimensionality reduction and data
  representation.
\newblock {\em Neural computation}, 15(6):1373--1396.

\bibitem[Bell and Sejnowski, 1995]{bell1995non}
Bell, A.~J. and Sejnowski, T.~J. (1995).
\newblock A non-linear information maximisation algorithm that performs blind
  separation.
\newblock In {\em Advances in neural information processing systems}, pages
  467--474.

\bibitem[Bellman et~al., 1957]{bellman1957dynamic}
Bellman, R., Corporation, R., and Collection, K. M.~R. (1957).
\newblock {\em Dynamic Programming}.
\newblock Rand Corporation research study. Princeton University Press.

\bibitem[Bengio et~al., 2013]{bengio2013representation}
Bengio, Y., Courville, A., and Vincent, P. (2013).
\newblock Representation learning: A review and new perspectives.
\newblock {\em IEEE transactions on pattern analysis and machine intelligence},
  35(8):1798--1828.

\bibitem[Blaschke and Wiskott, 2004]{blaschke2004cubica}
Blaschke, T. and Wiskott, L. (2004).
\newblock Cu{BICA}: Independent component analysis by simultaneous third-and
  fourth-order cumulant diagonalization.
\newblock {\em IEEE Transactions on Signal Processing}, 52(5):1250--1256.

\bibitem[Blaschke et~al., 2007]{blaschke2007independent}
Blaschke, T., Zito, T., and Wiskott, L. (2007).
\newblock Independent slow feature analysis and nonlinear blind source
  separation.
\newblock {\em Neural computation}, 19(4):994--1021.

\bibitem[Brakel and Bengio, 2017]{brakel2017learning}
Brakel, P. and Bengio, Y. (2017).
\newblock Learning independent features with adversarial nets for non-linear
  {ICA}.
\newblock {\em arXiv preprint arXiv:1710.05050}.

\bibitem[Brys et~al., 2015]{brys2015policy}
Brys, T., Harutyunyan, A., Taylor, M.~E., and Now{\'e}, A. (2015).
\newblock Policy transfer using reward shaping.
\newblock In {\em Proceedings of the 2015 International Conference on
  Autonomous Agents and Multiagent Systems}, pages 181--188.

\bibitem[Brys et~al., 2014]{brys2014multi}
Brys, T., Harutyunyan, A., Vrancx, P., Taylor, M.~E., Kudenko, D., and
  Now{\'e}, A. (2014).
\newblock Multi-objectivization of reinforcement learning problems by reward
  shaping.
\newblock In {\em 2014 international joint conference on neural networks
  (IJCNN)}, pages 2315--2322. IEEE.

\bibitem[Bucilua et~al., 2006]{bucilua2006model}
Bucilua, C., Caruana, R., and Niculescu-Mizil, A. (2006).
\newblock Model compression.
\newblock In {\em Proceedings of the 12th ACM SIGKDD international conference
  on Knowledge discovery and data mining}, pages 535--541.

\bibitem[Buckman et~al., 2018]{buckman2018sample}
Buckman, J., Hafner, D., Tucker, G., Brevdo, E., and Lee, H. (2018).
\newblock Sample-efficient reinforcement learning with stochastic ensemble
  value expansion.
\newblock In {\em Advances in Neural Information Processing Systems}, pages
  8224--8234.

\bibitem[Cardoso, 1997]{cardoso1997infomax}
Cardoso, J.-F. (1997).
\newblock {InfoMax} and maximum likelihood for blind source separation.
\newblock {\em IEEE Signal processing letters}, 4(4):112--114.

\bibitem[Chen et~al., 2015]{chen2015mxnet}
Chen, T., Li, M., Li, Y., Lin, M., Wang, N., Wang, M., Xiao, T., Xu, B., Zhang,
  C., and Zhang, Z. (2015).
\newblock Mxnet: A flexible and efficient machine learning library for
  heterogeneous distributed systems.
\newblock {\em arXiv preprint arXiv:1512.01274}.

\bibitem[Chen et~al., 2016]{chen2016infogan}
Chen, X., Duan, Y., Houthooft, R., Schulman, J., Sutskever, I., and Abbeel, P.
  (2016).
\newblock Infogan: Interpretable representation learning by information
  maximizing generative adversarial nets.
\newblock In {\em Advances in neural information processing systems}, pages
  2172--2180.

\bibitem[Chevalier-Boisvert et~al., 2018]{gym_minigrid}
Chevalier-Boisvert, M., Willems, L., and Pal, S. (2018).
\newblock Minimalistic gridworld environment for openai gym.
\newblock \url{https://github.com/maximecb/gym-minigrid}.

\bibitem[Chollet et~al., 2015]{chollet2015keras}
Chollet, F. et~al. (2015).
\newblock Keras.
\newblock https://keras.io.

\bibitem[Chua et~al., 2018]{chua2018deep}
Chua, K., Calandra, R., McAllister, R., and Levine, S. (2018).
\newblock Deep reinforcement learning in a handful of trials using
  probabilistic dynamics models.
\newblock In {\em Advances in Neural Information Processing Systems}, pages
  4754--4765.

\bibitem[Clark, 2015]{clark20152015}
Clark, J. (2015).
\newblock Why 2015 was a breakthrough year in artificial intelligence.
\newblock {\em Bloomberg Technology [online journal]}, 18(1).

\bibitem[Corneil et~al., 2018]{corneil2018efficient}
Corneil, D., Gerstner, W., and Brea, J. (2018).
\newblock Efficient model-based deep reinforcement learning with variational
  state tabulation.
\newblock {\em arXiv preprint arXiv:1802.04325}.

\bibitem[Cuccu et~al., 2018]{cuccu2018playing}
Cuccu, G., Togelius, J., and Cudr{\'e}-Mauroux, P. (2018).
\newblock Playing atari with six neurons.
\newblock {\em arXiv preprint arXiv:1806.01363}.

\bibitem[Curry, 1944]{curry1944method}
Curry, H.~B. (1944).
\newblock The method of steepest descent for non-linear minimization problems.
\newblock {\em Quarterly of Applied Mathematics}, 2(3):258--261.

\bibitem[Cybenko, 1989]{cybenko1989approximation}
Cybenko, G. (1989).
\newblock Approximation by superpositions of a sigmoidal function.
\newblock {\em Mathematics of control, signals and systems}, 2(4):303--314.

\bibitem[Darmois, 1953]{darmois1953analyse}
Darmois, G. (1953).
\newblock Analyse g{\'e}n{\'e}rale des liaisons stochastiques: etude
  particuli{\`e}re de l'analyse factorielle lin{\'e}aire.
\newblock {\em Revue de l'Institut international de statistique}, pages 2--8.

\bibitem[Deng et~al., 2009]{deng2009imagenet}
Deng, J., Dong, W., Socher, R., Li, L.-J., Li, K., and Fei-Fei, L. (2009).
\newblock Imagenet: A large-scale hierarchical image database.
\newblock In {\em 2009 IEEE conference on computer vision and pattern
  recognition}, pages 248--255. Ieee.

\bibitem[Denton et~al., 2017]{denton2017unsupervised}
Denton, E.~L. et~al. (2017).
\newblock Unsupervised learning of disentangled representations from video.
\newblock In {\em Advances in neural information processing systems}, pages
  4414--4423.

\bibitem[Dhariwal et~al., 2017]{baselines}
Dhariwal, P., Hesse, C., Klimov, O., Nichol, A., Plappert, M., Radford, A.,
  Schulman, J., Sidor, S., Wu, Y., and Zhokhov, P. (2017).
\newblock Openai baselines.
\newblock \url{https://github.com/openai/baselines}.

\bibitem[Donsker and Varadhan, 1975]{donsker1975asymptotic}
Donsker, M.~D. and Varadhan, S.~S. (1975).
\newblock Asymptotic evaluation of certain markov process expectations for
  large time, i.
\newblock {\em Communications on Pure and Applied Mathematics}, 28(1):1--47.

\bibitem[Dozat, 2016]{dozat2016incorporating}
Dozat, T. (2016).
\newblock Incorporating {N}esterov momentum into {ADAM}.

\bibitem[Du et~al., 2019]{du2019good}
Du, S.~S., Kakade, S.~M., Wang, R., and Yang, L.~F. (2019).
\newblock Is a good representation sufficient for sample efficient
  reinforcement learning?
\newblock {\em arXiv preprint arXiv:1910.03016}.

\bibitem[Efthymiadis and Kudenko, 2013]{efthymiadis2013using}
Efthymiadis, K. and Kudenko, D. (2013).
\newblock Using plan-based reward shaping to learn strategies in starcraft:
  Broodwar.
\newblock In {\em 2013 IEEE Conference on Computational Inteligence in Games
  (CIG)}, pages 1--8. IEEE.

\bibitem[Escalante-B and Wiskott, 2013]{escalante2013solve}
Escalante-B, A.~N. and Wiskott, L. (2013).
\newblock How to solve classification and regression problems on
  high-dimensional data with a supervised extension of slow feature analysis.
\newblock {\em The Journal of Machine Learning Research}, 14(1):3683--3719.

\bibitem[Gal et~al., 2016]{gal2016improving}
Gal, Y., McAllister, R., and Rasmussen, C.~E. (2016).
\newblock Improving pilco with bayesian neural network dynamics models.
\newblock In {\em Data-Efficient Machine Learning workshop, ICML}, volume~4,
  page~34.

\bibitem[Garcia et~al., 1989]{garcia1989model}
Garcia, C.~E., Prett, D.~M., and Morari, M. (1989).
\newblock Model predictive control: theory and practice—a survey.
\newblock {\em Automatica}, 25(3):335--348.

\bibitem[Gelada et~al., 2019]{gelada2019deepmdp}
Gelada, C., Kumar, S., Buckman, J., Nachum, O., and Bellemare, M.~G. (2019).
\newblock Deepmdp: Learning continuous latent space models for representation
  learning.
\newblock {\em arXiv preprint arXiv:1906.02736}.

\bibitem[Gidaris et~al., 2018]{gidaris2018unsupervised}
Gidaris, S., Singh, P., and Komodakis, N. (2018).
\newblock Unsupervised representation learning by predicting image rotations.
\newblock {\em arXiv preprint arXiv:1803.07728}.

\bibitem[Gogna and Majumdar, 2016]{gogna2016semi}
Gogna, A. and Majumdar, A. (2016).
\newblock Semi supervised autoencoder.
\newblock In {\em International Conference on Neural Information Processing},
  pages 82--89. Springer.

\bibitem[Goodfellow et~al., 2016]{Goodfellow-et-al-2016}
Goodfellow, I., Bengio, Y., and Courville, A. (2016).
\newblock {\em Deep Learning}.
\newblock MIT Press.
\newblock \url{http://www.deeplearningbook.org}.

\bibitem[Goodfellow et~al., 2014]{goodfellow2014generative}
Goodfellow, I., Pouget-Abadie, J., Mirza, M., Xu, B., Warde-Farley, D., Ozair,
  S., Courville, A., and Bengio, Y. (2014).
\newblock Generative adversarial nets.
\newblock In {\em Advances in neural information processing systems}, pages
  2672--2680.

\bibitem[Goroshin et~al., 2015]{goroshin2015learning}
Goroshin, R., Mathieu, M.~F., and LeCun, Y. (2015).
\newblock Learning to linearize under uncertainty.
\newblock In {\em Advances in Neural Information Processing Systems}, pages
  1234--1242.

\bibitem[Ha and Schmidhuber, 2018]{ha2018world}
Ha, D. and Schmidhuber, J. (2018).
\newblock World models.
\newblock {\em arXiv preprint arXiv:1803.10122}.

\bibitem[Hadsell et~al., 2006]{hadsell2006dimensionality}
Hadsell, R., Chopra, S., and LeCun, Y. (2006).
\newblock Dimensionality reduction by learning an invariant mapping.
\newblock In {\em 2006 IEEE Computer Society Conference on Computer Vision and
  Pattern Recognition (CVPR'06)}, volume~2, pages 1735--1742. IEEE.

\bibitem[Hafner et~al., 2019]{hafner2019learning}
Hafner, D., Lillicrap, T., Fischer, I., Villegas, R., Ha, D., Lee, H., and
  Davidson, J. (2019).
\newblock Learning latent dynamics for planning from pixels.
\newblock In {\em International Conference on Machine Learning}, pages
  2555--2565. PMLR.

\bibitem[Hamilton et~al., 2014]{hamilton2014efficient}
Hamilton, W., Fard, M.~M., and Pineau, J. (2014).
\newblock Efficient learning and planning with compressed predictive states.
\newblock {\em The Journal of Machine Learning Research}, 15(1):3395--3439.

\bibitem[Henaff et~al., 2017]{henaff2017model}
Henaff, M., Whitney, W.~F., and LeCun, Y. (2017).
\newblock Model-based planning with discrete and continuous actions.
\newblock {\em arXiv preprint arXiv:1705.07177}.

\bibitem[Hill et~al., 2018]{stable-baselines}
Hill, A., Raffin, A., Ernestus, M., Gleave, A., Kanervisto, A., Traore, R.,
  Dhariwal, P., Hesse, C., Klimov, O., Nichol, A., Plappert, M., Radford, A.,
  Schulman, J., Sidor, S., and Wu, Y. (2018).
\newblock Stable baselines.
\newblock \url{https://github.com/hill-a/stable-baselines}.

\bibitem[Hinton et~al., 2015]{hinton2015distilling}
Hinton, G., Vinyals, O., and Dean, J. (2015).
\newblock Distilling the knowledge in a neural network.
\newblock {\em arXiv preprint arXiv:1503.02531}.

\bibitem[Hjelm et~al., 2018]{hjelm2018learning}
Hjelm, R.~D., Fedorov, A., Lavoie-Marchildon, S., Grewal, K., Trischler, A.,
  and Bengio, Y. (2018).
\newblock Learning deep representations by mutual information estimation and
  maximization.
\newblock {\em arXiv preprint arXiv:R1808.06670}.

\bibitem[Hlynsson et~al., 2019]{hlynsson2019measuring}
Hlynsson, H., Escalante-B., A., and Wiskott, L. (2019).
\newblock Measuring the data efficiency of deep learning methods.
\newblock In {\em Proceedings of the 8th International Conference on Pattern
  Recognition Applications and Methods}. SCITEPRESS - Science and Technology
  Publications.

\bibitem[Hlynsson et~al., 2020]{hlynsson2020latent}
Hlynsson, H.~D., Sch{\"u}ler, M., Schiewer, R., Glasmachers, T., and Wiskott,
  L. (2020).
\newblock Latent representation prediction networks.
\newblock {\em arXiv preprint arXiv:2009.09439}.

\bibitem[Hlynsson and Wiskott, 2019]{hlynsson2019learning}
Hlynsson, H.~D. and Wiskott, L. (2019).
\newblock Learning gradient-based {ICA} by neurally estimating mutual
  information.
\newblock In {\em Joint German/Austrian Conference on Artificial Intelligence
  (K{\"u}nstliche Intelligenz)}, pages 182--187. Springer.

\bibitem[Hlynsson and Wiskott, 2021]{hlynsson2021reward}
Hlynsson, H.~D. and Wiskott, L. (2021).
\newblock Reward prediction for representation learning and reward shaping.

\bibitem[Huang, 2018]{dan2018how}
Huang, D. (2018).
\newblock How much did alphago zero cost?
\newblock {\em Internet: https://www.yuzeh.com/data/agz-cost.html}.

\bibitem[Hyv{\"a}rinen and Oja, 2000]{hyvarinen2000independent}
Hyv{\"a}rinen, A. and Oja, E. (2000).
\newblock Independent component analysis: algorithms and applications.
\newblock {\em Neural networks}, 13(4-5):411--430.

\bibitem[Hyv{\"a}rinen and Pajunen, 1999]{hyvarinen1999nonlinear}
Hyv{\"a}rinen, A. and Pajunen, P. (1999).
\newblock Nonlinear independent component analysis: Existence and uniqueness
  results.
\newblock {\em Neural Networks}, 12(3):429--439.

\bibitem[Ilin and Honkela, 2004]{ilin2004post}
Ilin, A. and Honkela, A. (2004).
\newblock Post-nonlinear independent component analysis by variational
  {B}ayesian learning.
\newblock In {\em International Conference on Independent Component Analysis
  and Signal Separation}, pages 766--773. Springer.

\bibitem[Irpan, 2018]{irpan2018deep}
Irpan, A. (2018).
\newblock Deep reinforcement learning doesn’t work yet.
\newblock {\em Internet: https://www. alexirpan. com/2018/02/14/rl-hard. html}.

\bibitem[Jutten and Karhunen, 2003]{jutten2003advances}
Jutten, C. and Karhunen, J. (2003).
\newblock Advances in nonlinear blind source separation.
\newblock In {\em Proc. of the 4th Int. Symp. on Independent Component Analysis
  and Blind Signal Separation (ICA2003)}, pages 245--256.

\bibitem[Kaelbling, 1993]{kaelbling1993learning}
Kaelbling, L.~P. (1993).
\newblock Learning to achieve goals.
\newblock In {\em IJCAI}, pages 1094--1099. Citeseer.

\bibitem[Khan et~al., 2018]{khan2018guide}
Khan, S., Rahmani, H., Shah, S. A.~A., and Bennamoun, M. (2018).
\newblock A guide to convolutional neural networks for computer vision.
\newblock {\em Synthesis Lectures on Computer Vision}, 8(1):1--207.

\bibitem[Kingma and Ba, 2014]{kingma2014adam}
Kingma, D.~P. and Ba, J. (2014).
\newblock Adam: A method for stochastic optimization.
\newblock {\em arXiv preprint arXiv:1412.6980}.

\bibitem[Kingma and Welling, 2013]{kingma2013auto}
Kingma, D.~P. and Welling, M. (2013).
\newblock Auto-encoding variational bayes.
\newblock {\em arXiv preprint arXiv:1312.6114}.

\bibitem[Kiran et~al., 2021]{kiran2021deep}
Kiran, B.~R., Sobh, I., Talpaert, V., Mannion, P., Al~Sallab, A.~A., Yogamani,
  S., and P{\'e}rez, P. (2021).
\newblock Deep reinforcement learning for autonomous driving: A survey.
\newblock {\em IEEE Transactions on Intelligent Transportation Systems}.

\bibitem[Krizhevsky et~al., 2012]{krizhevsky2012imagenet}
Krizhevsky, A., Sutskever, I., and Hinton, G.~E. (2012).
\newblock Imagenet classification with deep convolutional neural networks.
\newblock {\em Advances in neural information processing systems},
  25:1097--1105.

\bibitem[Kurutach et~al., 2018]{kurutach2018learning}
Kurutach, T., Tamar, A., Yang, G., Russell, S.~J., and Abbeel, P. (2018).
\newblock Learning plannable representations with causal infogan.
\newblock In {\em Advances in Neural Information Processing Systems}, pages
  8733--8744.

\bibitem[Lample and Chaplot, 2017]{lample2017playing}
Lample, G. and Chaplot, D.~S. (2017).
\newblock Playing fps games with deep reinforcement learning.
\newblock {\em Proceedings of the AAAI Conference on Artificial Intelligence},
  31(1).

\bibitem[LeCun et~al., 2004]{lecun2004learning}
LeCun, Y., Huang, F.~J., and Bottou, L. (2004).
\newblock Learning methods for generic object recognition with invariance to
  pose and lighting.
\newblock In {\em Proceedings of the 2004 IEEE Computer Society Conference on
  Computer Vision and Pattern Recognition, 2004. CVPR 2004.}, volume~2, pages
  II--104. IEEE.

\bibitem[LeCun et~al., 2012]{lecun2012efficient}
LeCun, Y.~A., Bottou, L., Orr, G.~B., and M{\"u}ller, K.-R. (2012).
\newblock Efficient backprop.
\newblock In {\em Neural networks: Tricks of the trade}, pages 9--48. Springer.

\bibitem[Lehnert and Littman, 2020]{lehnert2020successor}
Lehnert, L. and Littman, M.~L. (2020).
\newblock Successor features combine elements of model-free and model-based
  reinforcement learning.
\newblock {\em Journal of Machine Learning Research}, 21(196):1--53.

\bibitem[Lehnert et~al., 2020]{lehnert2020reward}
Lehnert, L., Littman, M.~L., and Frank, M.~J. (2020).
\newblock Reward-predictive representations generalize across tasks in
  reinforcement learning.
\newblock {\em PLoS computational biology}, 16(10):e1008317.

\bibitem[Li et~al., 2019]{li2019deep}
Li, Y., Huang, C., Ding, L., Li, Z., Pan, Y., and Gao, X. (2019).
\newblock Deep learning in bioinformatics: Introduction, application, and
  perspective in the big data era.
\newblock {\em Methods}, 166:4--21.

\bibitem[Linnainmaa, 1970]{linnainmaa1970representation}
Linnainmaa, S. (1970).
\newblock The representation of the cumulative rounding error of an algorithm
  as a taylor expansion of the local rounding errors.
\newblock {\em Master's Thesis (in Finnish), Univ. Helsinki}, pages 6--7.

\bibitem[Liu et~al., 2020]{liu2020hallucinative}
Liu, K., Kurutach, T., Tung, C., Abbeel, P., and Tamar, A. (2020).
\newblock Hallucinative topological memory for zero-shot visual planning.
\newblock {\em arXiv preprint arXiv:2002.12336}.

\bibitem[Maaten and Hinton, 2008]{maaten2008visualizing}
Maaten, L. v.~d. and Hinton, G. (2008).
\newblock Visualizing data using t-sne.
\newblock {\em Journal of machine learning research}, 9(Nov):2579--2605.

\bibitem[Marashi et~al., 2012]{marashi2012automatic}
Marashi, M., Khalilian, A., and Shiri, M.~E. (2012).
\newblock Automatic reward shaping in reinforcement learning using graph
  analysis.
\newblock In {\em 2012 2nd International eConference on Computer and Knowledge
  Engineering (ICCKE)}, pages 111--116. IEEE.

\bibitem[Masci et~al., 2011]{masci2011stacked}
Masci, J., Meier, U., Cire{\c{s}}an, D., and Schmidhuber, J. (2011).
\newblock Stacked convolutional auto-encoders for hierarchical feature
  extraction.
\newblock In {\em International conference on artificial neural networks},
  pages 52--59. Springer.

\bibitem[Mataric, 1994]{mataric1994reward}
Mataric, M.~J. (1994).
\newblock Reward functions for accelerated learning.
\newblock In {\em Machine learning proceedings 1994}, pages 181--189. Elsevier.

\bibitem[McCorduck, 1979]{mccorduck1979machines}
McCorduck, P. (1979).
\newblock Machines who think.

\bibitem[McInnes et~al., 2018]{mcinnes2018umap}
McInnes, L., Healy, J., and Melville, J. (2018).
\newblock Umap: Uniform manifold approximation and projection for dimension
  reduction.
\newblock {\em arXiv preprint arXiv:1802.03426}.

\bibitem[Mnih et~al., 2013]{mnih2013playing}
Mnih, V., Kavukcuoglu, K., Silver, D., Graves, A., Antonoglou, I., Wierstra,
  D., and Riedmiller, M. (2013).
\newblock Playing atari with deep reinforcement learning.
\newblock {\em arXiv preprint arXiv:1312.5602}.

\bibitem[Mnih et~al., 2015]{mnih2015human}
Mnih, V., Kavukcuoglu, K., Silver, D., Rusu, A.~A., Veness, J., Bellemare,
  M.~G., Graves, A., Riedmiller, M., Fidjeland, A.~K., Ostrovski, G., et~al.
  (2015).
\newblock Human-level control through deep reinforcement learning.
\newblock {\em Nature}, 518(7540):529.

\bibitem[Ng et~al., 1999]{ng1999policy}
Ng, A.~Y., Harada, D., and Russell, S. (1999).
\newblock Policy invariance under reward transformations: Theory and
  application to reward shaping.
\newblock In {\em Icml}, volume~99, pages 278--287.

\bibitem[Nwankpa et~al., 2018]{nwankpa2018activation}
Nwankpa, C., Ijomah, W., Gachagan, A., and Marshall, S. (2018).
\newblock Activation functions: Comparison of trends in practice and research
  for deep learning.
\newblock {\em arXiv preprint arXiv:1811.03378}.

\bibitem[Oh et~al., 2015]{oh2015action}
Oh, J., Guo, X., Lee, H., Lewis, R.~L., and Singh, S. (2015).
\newblock Action-conditional video prediction using deep networks in atari
  games.
\newblock In {\em Advances in neural information processing systems}, pages
  2863--2871.

\bibitem[Paszke et~al., 2019]{NEURIPS2019_9015}
Paszke, A., Gross, S., Massa, F., Lerer, A., Bradbury, J., Chanan, G., Killeen,
  T., Lin, Z., Gimelshein, N., Antiga, L., Desmaison, A., Kopf, A., Yang, E.,
  DeVito, Z., Raison, M., Tejani, A., Chilamkurthy, S., Steiner, B., Fang, L.,
  Bai, J., and Chintala, S. (2019).
\newblock Pytorch: An imperative style, high-performance deep learning library.
\newblock In Wallach, H., Larochelle, H., Beygelzimer, A., d\textquotesingle
  Alch\'{e}-Buc, F., Fox, E., and Garnett, R., editors, {\em Advances in Neural
  Information Processing Systems 32}, pages 8024--8035. Curran Associates, Inc.

\bibitem[Pathak et~al., 2018]{pathak2018zero}
Pathak, D., Mahmoudieh, P., Luo, G., Agrawal, P., Chen, D., Shentu, Y.,
  Shelhamer, E., Malik, J., Efros, A.~A., and Darrell, T. (2018).
\newblock Zero-shot visual imitation.
\newblock In {\em Proceedings of the IEEE Conference on Computer Vision and
  Pattern Recognition Workshops}, pages 2050--2053.

\bibitem[Pearson, 1901]{pearson1901liii}
Pearson, K. (1901).
\newblock Liii. on lines and planes of closest fit to systems of points in
  space.
\newblock {\em The London, Edinburgh, and Dublin Philosophical Magazine and
  Journal of Science}, 2(11):559--572.

\bibitem[Pedregosa et~al., 2011]{scikit-learn}
Pedregosa, F., Varoquaux, G., Gramfort, A., Michel, V., Thirion, B., Grisel,
  O., Blondel, M., Prettenhofer, P., Weiss, R., Dubourg, V., Vanderplas, J.,
  Passos, A., Cournapeau, D., Brucher, M., Perrot, M., and Duchesnay, E.
  (2011).
\newblock Scikit-learn: Machine learning in {P}ython.
\newblock {\em Journal of Machine Learning Research}, 12:2825--2830.

\bibitem[Rhodios, 2008]{rhodios2008argonautika}
Rhodios, A. (2008).
\newblock {\em The Argonautika}, volume~25.
\newblock Univ of California Press.

\bibitem[Richards, 2005]{richards2005robust}
Richards, A.~G. (2005).
\newblock {\em Robust constrained model predictive control}.
\newblock PhD thesis, Massachusetts Institute of Technology.

\bibitem[Richthofer and Wiskott, 2015]{richthofer2015predictable}
Richthofer, S. and Wiskott, L. (2015).
\newblock Predictable feature analysis.
\newblock In {\em 2015 IEEE 14th International Conference on Machine Learning
  and Applications (ICMLA)}, pages 190--196. IEEE.

\bibitem[Rosenblatt, 1958]{rosenblatt1958perceptron}
Rosenblatt, F. (1958).
\newblock The perceptron: a probabilistic model for information storage and
  organization in the brain.
\newblock {\em Psychological review}, 65(6):386.

\bibitem[Saphal et~al., 2020]{saphal2020seerl}
Saphal, R., Ravindran, B., Mudigere, D., Avancha, S., and Kaul, B. (2020).
\newblock Seerl: Sample efficient ensemble reinforcement learning.
\newblock {\em arXiv preprint arXiv:2001.05209}.

\bibitem[Savinov et~al., 2018]{savinov2018semi}
Savinov, N., Dosovitskiy, A., and Koltun, V. (2018).
\newblock Semi-parametric topological memory for navigation.
\newblock {\em arXiv preprint arXiv:1803.00653}.

\bibitem[Schaul et~al., 2015]{schaul2015universal}
Schaul, T., Horgan, D., Gregor, K., and Silver, D. (2015).
\newblock Universal value function approximators.
\newblock In {\em International conference on machine learning}, pages
  1312--1320.

\bibitem[Schmidhuber, 1992]{schmidhuber1992learning}
Schmidhuber, J. (1992).
\newblock Learning factorial codes by predictability minimization.
\newblock {\em Neural Computation}, 4(6):863--879.

\bibitem[Schmidhuber, 2018]{schmidhuberunsupervised}
Schmidhuber, J. (2018).
\newblock Unsupervised neural networks fight in a minimax game.
\newblock
  \url{https://people.idsia.ch/~juergen/unsupervised-neural-nets-fight-minimax-game.html}.

\bibitem[Schroff et~al., 2015]{schroff2015facenet}
Schroff, F., Kalenichenko, D., and Philbin, J. (2015).
\newblock Facenet: A unified embedding for face recognition and clustering.
\newblock In {\em Proceedings of the IEEE conference on computer vision and
  pattern recognition}, pages 815--823.

\bibitem[Sch{\"u}ler et~al., 2019]{schuler2018gradient}
Sch{\"u}ler, M., Hlynsson, H.~D., and Wiskott, L. (2019).
\newblock Gradient-based training of slow feature analysis by differentiable
  approximate whitening.
\newblock In {\em Asian Conference on Machine Learning}, pages 316--331. PMLR.

\bibitem[Schulman et~al., 2015]{schulman2015high}
Schulman, J., Moritz, P., Levine, S., Jordan, M., and Abbeel, P. (2015).
\newblock High-dimensional continuous control using generalized advantage
  estimation.
\newblock {\em arXiv preprint arXiv:1506.02438}.

\bibitem[Schulman et~al., 2017]{schulman2017proximal}
Schulman, J., Wolski, F., Dhariwal, P., Radford, A., and Klimov, O. (2017).
\newblock Proximal policy optimization algorithms.
\newblock {\em arXiv preprint arXiv:1707.06347}.

\bibitem[Scikit-learn, 2019]{sklearn}
Scikit-learn (2019).
\newblock Blind source separation using {FastICA}.
\newblock
  \url{https://scikit-learn.org/stable/auto\_examples/decomposition/plot\_ica\_blind\_source\_separation.html}.
\newblock Accessed: 2019-02-24.

\bibitem[Sejnowski, 2018]{sejnowski2018deep}
Sejnowski, T.~J. (2018).
\newblock {\em The deep learning revolution}.
\newblock Mit Press.

\bibitem[Sharif~Razavian et~al., 2014]{sharif2014cnn}
Sharif~Razavian, A., Azizpour, H., Sullivan, J., and Carlsson, S. (2014).
\newblock Cnn features off-the-shelf: an astounding baseline for recognition.
\newblock In {\em Proceedings of the IEEE conference on computer vision and
  pattern recognition workshops}, pages 806--813.

\bibitem[Shlens, 2014]{shlens2014tutorial}
Shlens, J. (2014).
\newblock A tutorial on principal component analysis.
\newblock {\em arXiv preprint arXiv:1404.1100}.

\bibitem[Silver et~al., 2017]{silver2017mastering}
Silver, D., Schrittwieser, J., Simonyan, K., Antonoglou, I., Huang, A., Guez,
  A., Hubert, T., Baker, L., Lai, M., Bolton, A., et~al. (2017).
\newblock Mastering the game of go without human knowledge.
\newblock {\em nature}, 550(7676):354--359.

\bibitem[Simonyan and Zisserman, 2014]{simonyan2014very}
Simonyan, K. and Zisserman, A. (2014).
\newblock Very deep convolutional networks for large-scale image recognition.
\newblock {\em arXiv preprint arXiv:1409.1556}.

\bibitem[Sprekeler et~al., 2014]{sprekeler2014extension}
Sprekeler, H., Zito, T., and Wiskott, L. (2014).
\newblock An extension of slow feature analysis for nonlinear blind source
  separation.
\newblock {\em The Journal of Machine Learning Research}, 15(1):921--947.

\bibitem[Srinivas et~al., 2018]{srinivas2018universal}
Srinivas, A., Jabri, A., Abbeel, P., Levine, S., and Finn, C. (2018).
\newblock Universal planning networks.
\newblock {\em arXiv preprint arXiv:1804.00645}.

\bibitem[Sun et~al., 2014]{sun2014deep}
Sun, Y., Chen, Y., Wang, X., and Tang, X. (2014).
\newblock Deep learning face representation by joint
  identification-verification.
\newblock In {\em Advances in neural information processing systems}, pages
  1988--1996.

\bibitem[Sutton, 1988]{sutton1988learning}
Sutton, R.~S. (1988).
\newblock Learning to predict by the methods of temporal differences.
\newblock {\em Machine learning}, 3(1):9--44.

\bibitem[Sutton and Barto, 2018]{sutton2018reinforcement}
Sutton, R.~S. and Barto, A.~G. (2018).
\newblock {\em Reinforcement learning: An introduction}.
\newblock MIT press.

\bibitem[Sutton et~al., 1999]{sutton1999between}
Sutton, R.~S., Precup, D., and Singh, S. (1999).
\newblock Between mdps and semi-mdps: A framework for temporal abstraction in
  reinforcement learning.
\newblock {\em Artificial intelligence}, 112(1-2):181--211.

\bibitem[Taleb and Jutten, 1999]{taleb1999source}
Taleb, A. and Jutten, C. (1999).
\newblock Source separation in post-nonlinear mixtures.
\newblock {\em IEEE Transactions on signal Processing}, 47(10):2807--2820.

\bibitem[Tamar et~al., 2016]{tamar2016value}
Tamar, A., Wu, Y., Thomas, G., Levine, S., and Abbeel, P. (2016).
\newblock Value iteration networks.
\newblock In {\em Advances in Neural Information Processing Systems}, pages
  2154--2162.

\bibitem[Tieleman and Hinton, 2012]{tieleman2012lecture}
Tieleman, T. and Hinton, G. (2012).
\newblock Lecture 6.5-rmsprop: Divide the gradient by a running average of its
  recent magnitude.
\newblock {\em COURSERA: Neural networks for machine learning}, 4(2):26--31.

\bibitem[Tobler, 1970]{tobler1970computer}
Tobler, W.~R. (1970).
\newblock A computer movie simulating urban growth in the detroit region.
\newblock {\em Economic geography}, 46(sup1):234--240.

\bibitem[Trott et~al., 2019]{trott2019keeping}
Trott, A., Zheng, S., Xiong, C., and Socher, R. (2019).
\newblock Keeping your distance: Solving sparse reward tasks using
  self-balancing shaped rewards.
\newblock {\em arXiv preprint arXiv:1911.01417}.

\bibitem[Veli{\v{c}}kovi{\'c} et~al., 2018]{velivckovic2018deep}
Veli{\v{c}}kovi{\'c}, P., Fedus, W., Hamilton, W.~L., Li{\`o}, P., Bengio, Y.,
  and Hjelm, R.~D. (2018).
\newblock Deep graph {InfoMax}.
\newblock {\em arXiv preprint arXiv:1809.10341}.

\bibitem[Vincent et~al., 2008]{vincent2008extracting}
Vincent, P., Larochelle, H., Bengio, Y., and Manzagol, P.-A. (2008).
\newblock Extracting and composing robust features with denoising autoencoders.
\newblock In {\em Proceedings of the 25th international conference on Machine
  learning}, pages 1096--1103.

\bibitem[Vincent et~al., 2010]{vincent2010stacked}
Vincent, P., Larochelle, H., Lajoie, I., Bengio, Y., Manzagol, P.-A., and
  Bottou, L. (2010).
\newblock Stacked denoising autoencoders: Learning useful representations in a
  deep network with a local denoising criterion.
\newblock {\em Journal of machine learning research}, 11(12).

\bibitem[Vondrick et~al., 2016]{vondrick2016anticipating}
Vondrick, C., Pirsiavash, H., and Torralba, A. (2016).
\newblock Anticipating visual representations from unlabeled video.
\newblock In {\em Proceedings of the IEEE Conference on Computer Vision and
  Pattern Recognition}, pages 98--106.

\bibitem[Wang et~al., 2019]{wang2019learning}
Wang, A., Kurutach, T., Liu, K., Abbeel, P., and Tamar, A. (2019).
\newblock Learning robotic manipulation through visual planning and acting.
\newblock {\em arXiv preprint arXiv:1905.04411}.

\bibitem[Wang et~al., 2018]{wang2018toybox}
Wang, X., Ma, T., Ainooson, J., Cha, S., Wang, X., Molla, A., and Kunda, M.
  (2018).
\newblock The toybox dataset of egocentric visual object transformations.
\newblock {\em arXiv preprint arXiv:1806.06034}.

\bibitem[Watkins and Dayan, 1992]{watkins1992q}
Watkins, C.~J. and Dayan, P. (1992).
\newblock Q-learning.
\newblock {\em Machine learning}, 8(3-4):279--292.

\bibitem[Winston, 2010]{patrick2010ai}
Winston, P. (2010).
\newblock 6.034 artificial intelligence, fall 2010.

\bibitem[Wiskott and Sejnowski, 2002]{wiskott2002slow}
Wiskott, L. and Sejnowski, T.~J. (2002).
\newblock Slow feature analysis: Unsupervised learning of invariances.
\newblock {\em Neural computation}, 14(4):715--770.

\bibitem[Wolf et~al., 2020]{wolf2020transformers}
Wolf, T., Chaumond, J., Debut, L., Sanh, V., Delangue, C., Moi, A., Cistac, P.,
  Funtowicz, M., Davison, J., Shleifer, S., et~al. (2020).
\newblock Transformers: State-of-the-art natural language processing.
\newblock In {\em Proceedings of the 2020 Conference on Empirical Methods in
  Natural Language Processing: System Demonstrations}, pages 38--45.

\bibitem[Wu et~al., 2017]{wu2017scalable}
Wu, Y., Mansimov, E., Grosse, R.~B., Liao, S., and Ba, J. (2017).
\newblock Scalable trust-region method for deep reinforcement learning using
  kronecker-factored approximation.
\newblock In {\em Advances in neural information processing systems}, pages
  5279--5288.

\bibitem[Xu et~al., 2019]{xu2019regression}
Xu, D., Mart{\'\i}n-Mart{\'\i}n, R., Huang, D.-A., Zhu, Y., Savarese, S., and
  Fei-Fei, L.~F. (2019).
\newblock Regression planning networks.
\newblock In {\em Advances in Neural Information Processing Systems}, pages
  1319--1329.

\bibitem[Yao et~al., 2019]{yao2019negative}
Yao, H., Zhu, D.-l., Jiang, B., and Yu, P. (2019).
\newblock Negative log likelihood ratio loss for deep neural network
  classification.
\newblock In {\em Proceedings of the Future Technologies Conference}, pages
  276--282. Springer.

\bibitem[Zeiler and Fergus, 2014]{zeiler2014visualizing}
Zeiler, M.~D. and Fergus, R. (2014).
\newblock Visualizing and understanding convolutional networks.
\newblock In {\em European conference on computer vision}, pages 818--833.
  Springer.

\bibitem[Zhang et~al., 2016]{zhang2016colorful}
Zhang, R., Isola, P., and Efros, A.~A. (2016).
\newblock Colorful image colorization.
\newblock In {\em European conference on computer vision}, pages 649--666.
  Springer.

\bibitem[Zhang et~al., 2019]{zhang2019deep}
Zhang, S., Yao, L., Sun, A., and Tay, Y. (2019).
\newblock Deep learning based recommender system: A survey and new
  perspectives.
\newblock {\em ACM Computing Surveys (CSUR)}, 52(1):1--38.

\bibitem[Ziehe et~al., 2003]{ziehe2003blind}
Ziehe, A., Kawanabe, M., Harmeling, S., and M{ü}ller, K.-R. (2003).
\newblock Blind separation of post-nonlinear mixtures using linearizing
  transformations and temporal decorrelation.
\newblock {\em Journal of Machine Learning Research}, 4(Dec):1319--1338.

\bibitem[Zou et~al., 2019]{zou2019reward}
Zou, H., Ren, T., Yan, D., Su, H., and Zhu, J. (2019).
\newblock Reward shaping via meta-learning.
\newblock {\em arXiv preprint arXiv:1901.09330}.

\end{thebibliography}
\appendix

\chapter{}

\vspace*{-1.6in}
\hspace*{-1.2in}
\includegraphics[scale=0.97]{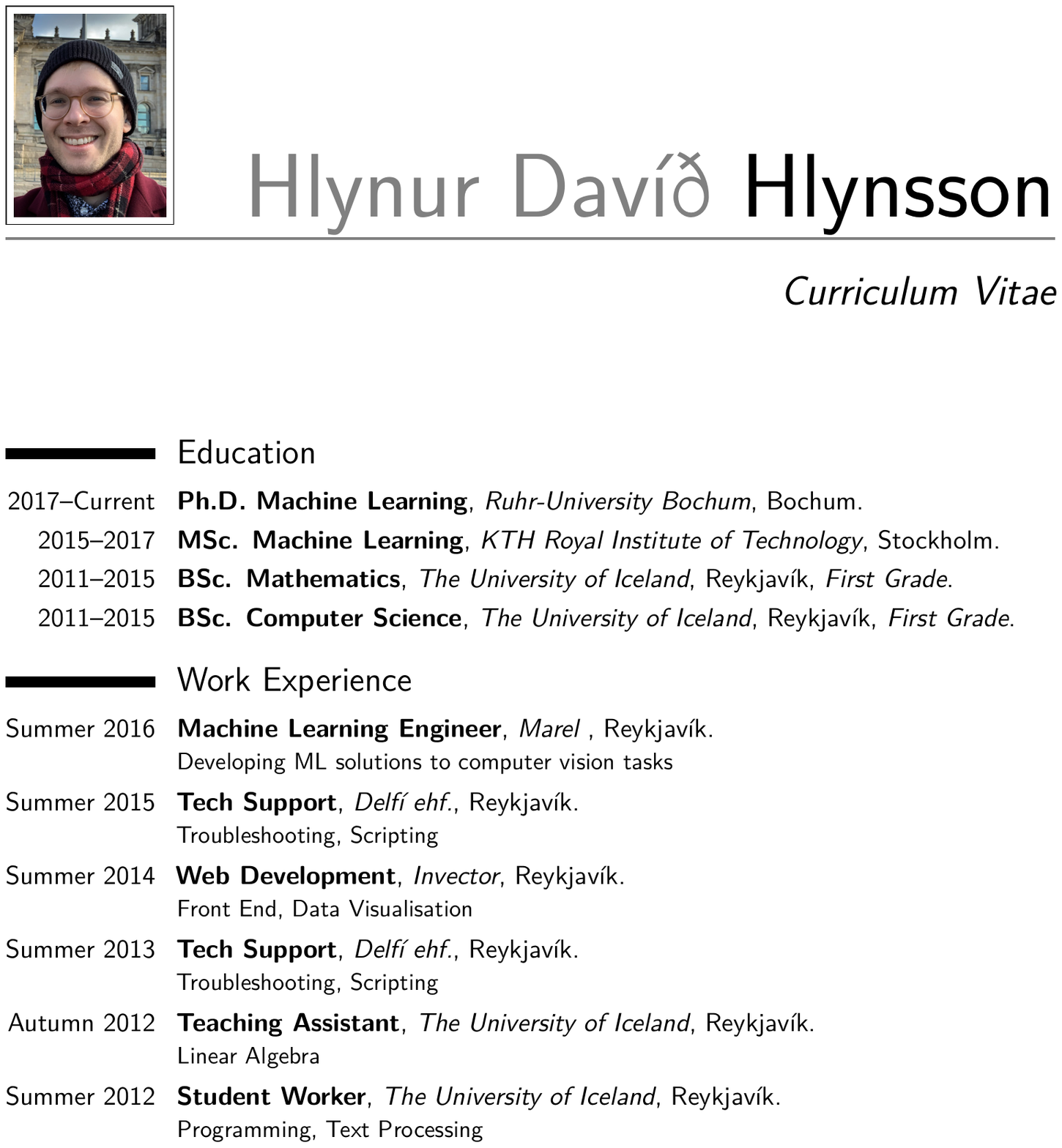}
\newpage

\vspace*{-0.5in}
\hspace*{-1.2in}
\includegraphics[scale=0.97]{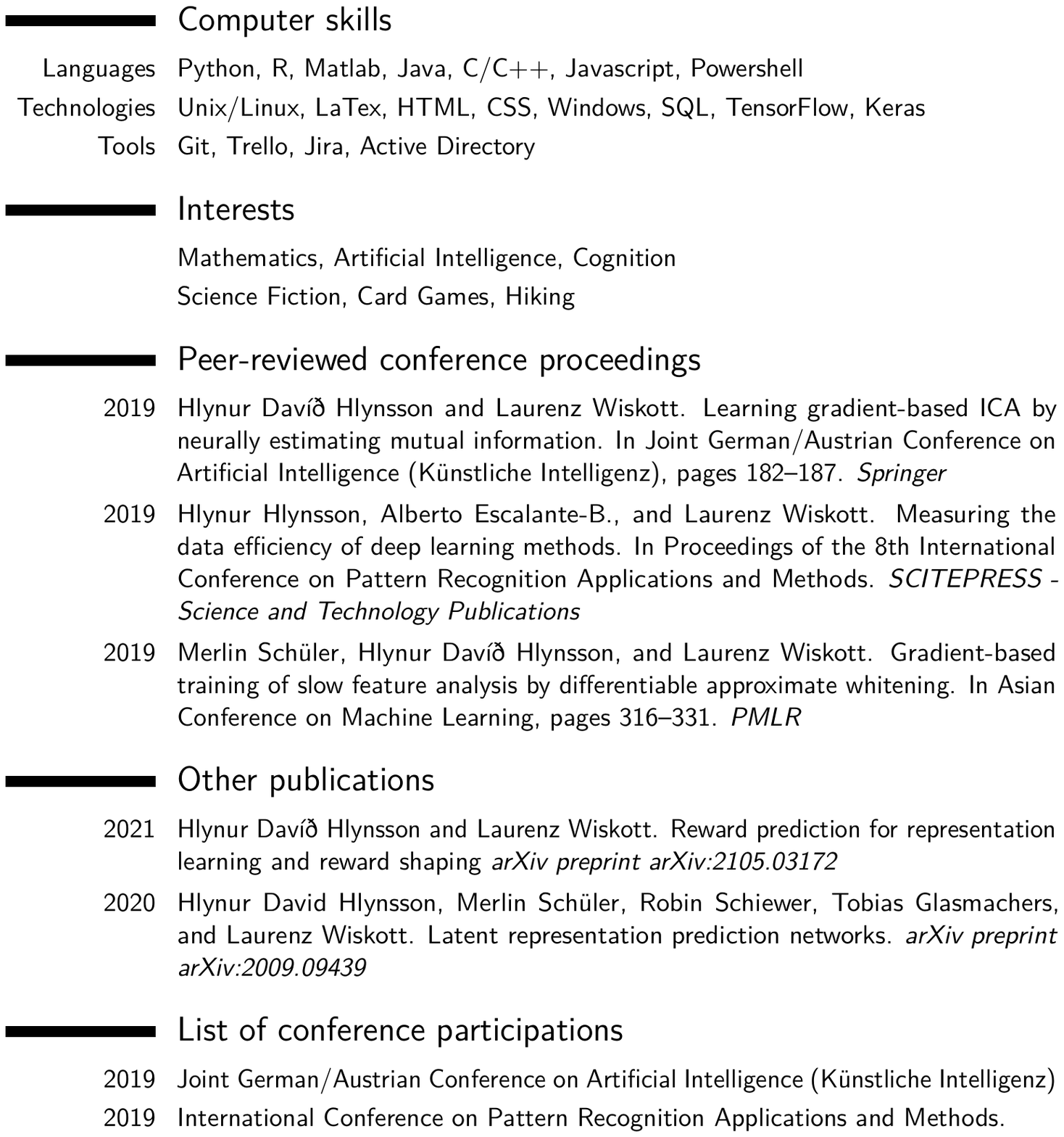}

\renewcommand{\nomgroup}[1]{%
\ifthenelse{\equal{#1}{C}}{\item[\textbf{Roman Symbols}]}{%
    \ifthenelse{\equal{#1}{V}}{\item[\textbf{Greek Symbols}]}{%
        \ifthenelse{\equal{#1}{S}}{\item[\textbf{Abbreviations and Acronyms}]}{}
        }
    }
}

\pdfbookmark[1]{Nomenclature}{nomenclature}

\nomenclature[C]{$\mathcal{S}$}{State Space}
\nomenclature[C]{$\mathcal{A}$}{Action Space}
\nomenclature[C]{$\mathbb{R}$}{Real Numbers}
\nomenclature[S]{RL}{Reinforcement Learning}
\nomenclature[S]{MDP}{Markov Decision Process}
\nomenclature[S]{POMDP}{Partially Observable MDP}
\nomenclature[S]{MLP}{Multilayer Perceptron}
\nomenclature[S]{ANN}{Artificial Neural Network}
\nomenclature[S]{DL}{Deep Learning}
\nomenclature[S]{PCA}{Principal Component Analysis}
\nomenclature[S]{ICA}{Independent Component Analysis}
\nomenclature[S]{SF}{Successor Features}
\nomenclature[S]{GrICA}{Our Gradient-based ICA Algorithm (from Chapter 4)}
\nomenclature[S]{LARP}{Latent Representation Prediction (from chapter 5)}
\nomenclature[S]{RewPred}{Our Reward-predictive Representation (from Chapter 6) }
\nomenclature[S]{MINE}{Mutual Information Neural Estimation}
\nomenclature[S]{CAE}{Variational Autoencoder }
\nomenclature[S]{VAE}{Convolutional Autoencoder }
\nomenclature[S]{LEM}{Laplacian Eigenmaps }
\nomenclature[S]{Conv.}{Convolutional }
\nomenclature[S]{ReLU}{Rectified Linear Unit }
\nomenclature[S]{t-SNE}{t-distributed Stochastic Neighbor Embedding }
\nomenclature[S]{PPO}{Proximal Policy Optimization}
\nomenclature[S]{DQN}{Deep Q-Network}
\nomenclature[S]{ACKTR}{Actor Critic using Kronecker-Factored Trust Region}
\nomenclature[S]{MSE}{Mean Squared Error}
\nomenclature[S]{MBRL}{Model-based Reinforcement Learning}
\nomenclature[S]{MBRL}{Model-free Reinforcement Learning}

\nomenclature[V]{$\phi$}{Representation}
\nomenclature[V]{$\eta$}{Learning rate parameter}

\nomenclature[V]{$\pi$}{Policy function }
\nomenclature[V]{$\pi^*$}{Optimal Policy }
\nomenclature[V]{$\theta$}{The parameters of a differentiable function}

\nomenclature[C]{$\mathcal{P}$}{State Transition }
\nomenclature[C]{$P$}{Probability }

\nomenclature[C]{$\mathcal{R}$}{Reward function}

\nomenclature[C]{$ V_\pi$}{state-value function of $\pi$}
\nomenclature[C]{$ q_\pi$}{action-value function of $\pi$}

\nomenclature[C]{$\mathcal{D}$}{Data set}
\nomenclature[V]{$\Omega$}{Observation space}
\nomenclature[C]{$\mathcal{O}$}{Observation function}

\nomenclature[C]{$t$}{Time step index}
\nomenclature[C]{$a$}{Action}
\nomenclature[C]{$s$}{State}
\nomenclature[C]{$o$}{Observation}
\nomenclature[C]{$r$}{Reward}
\nomenclature[V]{$\gamma$}{Discount factor}
\nomenclature[C]{$\mathcal{L}$}{Loss function}

\printnomenclature

\newpage

\end{document}